\documentclass[10pt,letterpaper]{article}
\ifdefined\pdfsuppressptexinfo \pdfsuppressptexinfo=15 \fi
\ifdefined\pdfvariable \pdfvariable suppressoptionalinfo 15 \fi
\PassOptionsToPackage{expansion=false}{microtype}
\usepackage{arxiv}
\usepackage[T1]{fontenc}
\usepackage[utf8]{inputenc}
\usepackage{amsmath,amssymb}
\usepackage{booktabs,array,multirow,tabularx}
\usepackage{graphicx}
\usepackage{placeins}
\usepackage[table]{xcolor}
\usepackage{microtype}
\usepackage{url}
\usepackage[font=small,labelfont=bf]{caption}
\usepackage[colorlinks=true,citecolor=blue!55!black,urlcolor=blue!55!black,linkcolor=blue!55!black]{hyperref}
\extrafloats{100}
\renewcommand{\headeright}{} 
\renewcommand{\shorttitle}{Can Multimodal Fusion \textit{MakeSense} for Embodied Vision in Degraded Sensing?}

\makeatletter
\newenvironment{teaserfigure}{%
  \par\noindent
  \begin{minipage}{\textwidth}%
    \centering
    \def\@captype{figure}%
}{%
  \end{minipage}\par
}
\makeatother

\makeatletter\@ifundefined{captionof}{\newcommand{\captionof}[1]{\def\@captype{#1}\caption}}{}\makeatother

\newcommand{\dn}{$\downarrow$}
\newcommand{\up}{$\uparrow$}

\definecolor{sdgray}{gray}{0.45}
\newcommand{\pmsd}[1]{{\tiny\textcolor{sdgray}{$\pm$#1}}}   
\newcommand{\pmsdt}[1]{{\tiny\textcolor{sdgray}{$\pm$#1}}}
\newcommand{\ctx}[1]{\textcolor{black!55}{#1}}            
\newcommand{\hd}[1]{{\tiny #1}}                           
\newcommand{\ours}{MakeSense}   
\newif\ifsupp\suppfalse 
\newcommand{\supref}[1]{\ifsupp\ref{#1}\else\csname supfb@#1\endcsname\fi}   
\definecolor{gaingreen}{RGB}{0,128,60}
\definecolor{sevbest}{RGB}{198,239,206}
\definecolor{sevok}{RGB}{255,242,204}
\definecolor{sevmid}{RGB}{252,213,180}
\definecolor{sevbad}{RGB}{248,196,196}
\definecolor{figred}{HTML}{D62728}     
\definecolor{figgreen}{HTML}{009E73}
\definecolor{figamber}{HTML}{F0A202}
\definecolor{figblue}{HTML}{1F4E9C}
\newcommand{\gain}[1]{\textcolor{gaingreen}{#1}}
\newcommand{\grp}[1]{\multicolumn{1}{@{}l}{\textbf{#1}}}
\newcommand{\mh}[1]{{\scriptsize #1}}
\providecommand{\gain}[1]{\textcolor{gaingreen}{#1}}
\providecommand{\sev}[2]{\cellcolor{#1}#2}
\providecommand{\sevkey}[2]{{\setlength{\fboxsep}{0.8pt}\colorbox{#1}{#2}}}

\expandafter\def\csname supfb@fig:dose\endcsname{S2}
\expandafter\def\csname supfb@fig:grid\endcsname{S6}
\expandafter\def\csname supfb@fig:inputs\endcsname{S5}
\expandafter\def\csname supfb@fig:qual\endcsname{S4}
\expandafter\def\csname supfb@fig:reli\endcsname{S1}
\expandafter\def\csname supfb@tab:baselines\endcsname{S2}
\expandafter\def\csname supfb@tab:gap\endcsname{S1}
\expandafter\def\csname supfb@tab:he_ref\endcsname{S8}
\expandafter\def\csname supfb@tab:mmaud\endcsname{S5}
\expandafter\def\csname supfb@tab:mmfi_detail\endcsname{S3}
\expandafter\def\csname supfb@tab:reli\endcsname{S7}
\expandafter\def\csname supfb@tab:rgb\endcsname{S4}
\expandafter\def\csname supfb@tab:v2x_val\endcsname{S6}
\expandafter\def\csname supfb@fig:deg_final\endcsname{S8}
\expandafter\def\csname supfb@fig:forecast_transfer\endcsname{S10}
\expandafter\def\csname supfb@fig:task_results\endcsname{S7}
\expandafter\def\csname supfb@fig:wm_ablation\endcsname{S9}
\expandafter\def\csname supfb@tab:forecast\endcsname{S10}
\newcommand{\mods}{\mathcal M}
\newcommand{\sg}{\operatorname{sg}}
\renewcommand{\arraystretch}{1.05}

\title{Can Multimodal Fusion \textit{MakeSense} for Embodied Vision in Degraded Sensing?}
\author{%
\begin{minipage}{\textwidth}
\centering
{\small\bfseries
Cheng-Ru Chou\textsuperscript{*,1},
Shih-Yu Lai\textsuperscript{*,2,3,4},
Pei-Lin Chiang\textsuperscript{1},
Shou-Po Tsai\textsuperscript{1},
Jing-Zhe Huang\textsuperscript{1}\\[3pt]
Fukuharu Tanaka\textsuperscript{4,5},
Hirozumi Yamaguchi\textsuperscript{\dag,4,5},
Bing-Yu Chen\textsuperscript{\dag,2},
Yu-Lun Liu\textsuperscript{\dag,6},
and Hsun-Ping Hsieh\textsuperscript{\dag,1}\\[5pt]}
{\footnotesize
\begin{tabular}{@{}c@{\hspace{10pt}}c@{}}
\textsuperscript{1}National Cheng Kung University, Tainan, Taiwan &
\textsuperscript{2}National Taiwan University, Taipei, Taiwan \\[1pt]
\textsuperscript{3}MoonShine Animation Studio, Taipei, Taiwan &
\textsuperscript{4}RIKEN Center for Computational Science, Kobe, Japan \\[1pt]
\textsuperscript{5}The University of Osaka, Osaka, Japan &
\textsuperscript{6}National Yang Ming Chiao Tung University, Hsinchu, Taiwan
\end{tabular}}
\end{minipage}%
}
\date{}
\hypersetup{pdftitle={Can Multimodal Fusion MakeSense for Embodied Vision in Degraded Sensing?},
  pdfauthor={Cheng-Ru Chou, Shih-Yu Lai, Pei-Lin Chiang, Shou-Po Tsai, Jing-Zhe Huang, Fukuharu Tanaka, Hirozumi Yamaguchi, Bing-Yu Chen, Yu-Lun Liu, Hsun-Ping Hsieh}}

\begin{document}

\twocolumn[{%
  \begingroup
    \renewcommand{\twocolumn}[1][]{#1}%
    \maketitle
  \endgroup
  \begin{teaserfigure}
    \includegraphics[width=\linewidth]{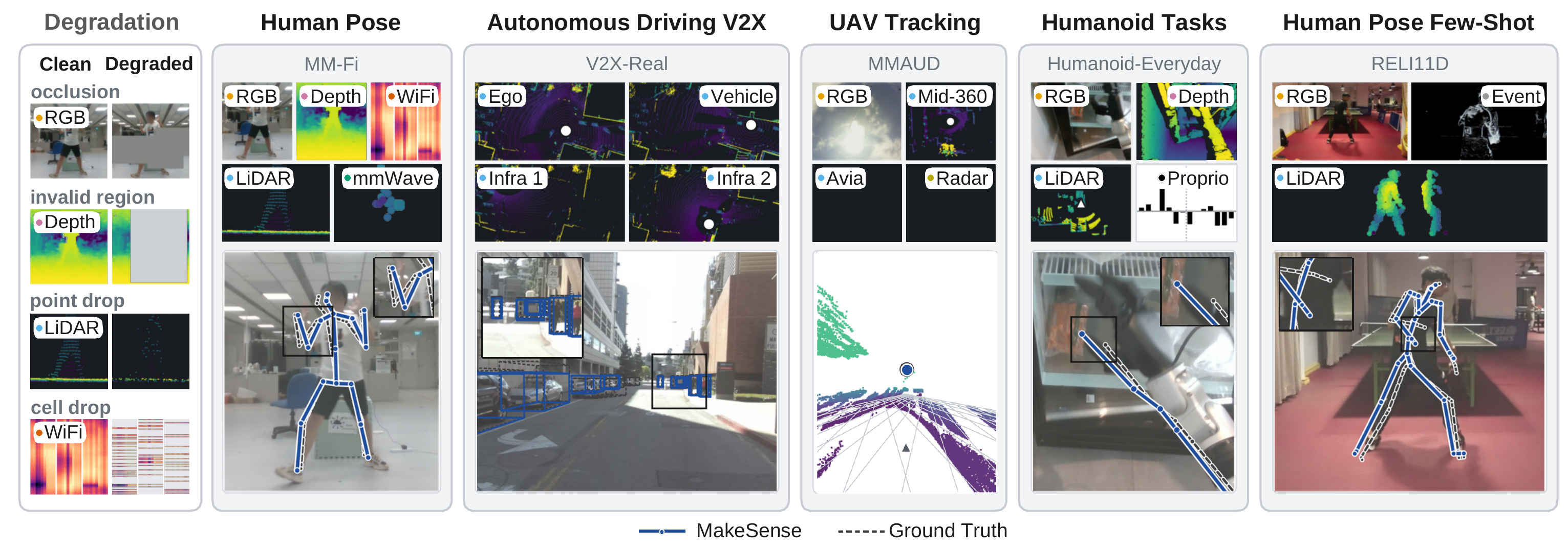}\par
    \caption{\textbf{One fusion design for missing and degraded sensors, on five embodied sensing datasets.} \ours{} takes whichever sensors are present and predicts 3D human pose (MM-Fi, RELI11D), cooperative 3D boxes (V2X-Real), UAV position (MMAUD) and humanoid arm commands (Humanoid-Everyday). Left: clean and degraded inputs illustrating training and test corruptions. 
    }
    \label{fig:teaser}
    \par\vspace*{10pt}
  \end{teaserfigure}
}]

\begin{abstract} 
Embodied vision systems rely on complementary sensors to perceive and act as conditions change. Even models that accept arbitrary sensor subsets can exhibit negative fusion: adding a sensor increases prediction error. We propose \ours{}, a multimodal fusion framework that preserves sensor-specific evidence in private streams and combines information through shared bottleneck tokens. Training pairs full and subset views of the same degraded observations, penalizing higher error with more sensors. Where geometry permits, the distance readout uses range measurements alone, while the rest of the prediction draws on all available inputs. Paired tests across five datasets reveal helpful and harmful sensor additions. \ours{} reduces negative fusion on MM-Fi. With lightweight adapters and per-dataset training, the same design improves UAV localization on MMAUD, cooperative 3D detection on V2X-Real, few-shot pose transfer to RELI11D, and command prediction on Humanoid-Everyday, where a world model trained on the frozen core's outputs forecasts future fused states more accurately than a linear baseline at longer tested horizons. \ours{} supports embodied sensing across human activity, aerial systems, robotics, and autonomous driving.
\end{abstract}
\begin{NoHyper}
\let\thefootnote\relax%
\makeatletter\long\def\@makefntext#1{\noindent#1}\makeatother%
\def\ml#1{{\scriptsize\urlstyle{tt}\nolinkurl{#1}}}%
\footnotetext{%
\textsuperscript{*}These authors contributed equally in this work.
\textsuperscript{\dag}Corresponding authors and advisors.
Contact: \ml{N26140870@gs.ncku.edu.tw} (Chou); \ml{akinesia112@gmail.com} (Lai); \ml{N26150728@gs.ncku.edu.tw} (Chiang); \ml{N26140642@gs.ncku.edu.tw} (Tsai); \ml{N26140634@gs.ncku.edu.tw} (Huang); \ml{fukuharu.tanaka@riken.jp} (Tanaka); \ml{hirozumi.yamaguchi@riken.jp} (Yamaguchi); \ml{robin@ntu.edu.tw} (Chen); \ml{yulunliu@cs.nycu.edu.tw} (Liu); \ml{hphsieh@mail.ncku.edu.tw} (Hsieh).\\
The code, model, webpage and video will be released after acceptance.
}%
\end{NoHyper}

\section{Introduction} 
Embodied vision systems combine sensors that measure different quantities in different formats, and the value of each sensor varies with the scene and the sensor's state. For instance, in poor light, RGB cues degrade while LiDAR and mmWave radar still
provide geometric measurements~\cite{yang2023mmfi}. Sensor availability can also fluctuate due to changes in cooperating agents~\cite{xiang2024v2xreal}
and delayed or missing proprioceptive measurements~\cite{zhao2025humanoideveryday}.

Many recent models run on whichever sensors are present. X-Fi fuses features from the available sensors~\cite{chen2025xfi}, COMPASS completes missing sensors with a generator~\cite{compass}, and HEAL fuses a varying set of agents~\cite{heal}. FlexPose selects LiDAR, mmWave, or both per sample~\cite{flexpose}, while AWC balances learning across modalities~\cite{awc}.

However, training on sensor subsets does not ensure that added sensors improve performance. A model that accepts arbitrary sensor subsets can still exhibit
\emph{negative fusion}: adding inputs increases prediction error. This raises a question: \emph{\textbf{When does adding a sensor hurt perception, and how can fusion reduce that harm as sensing conditions change?}}

\begin{figure}[t]
\centering
\includegraphics[width=\linewidth]{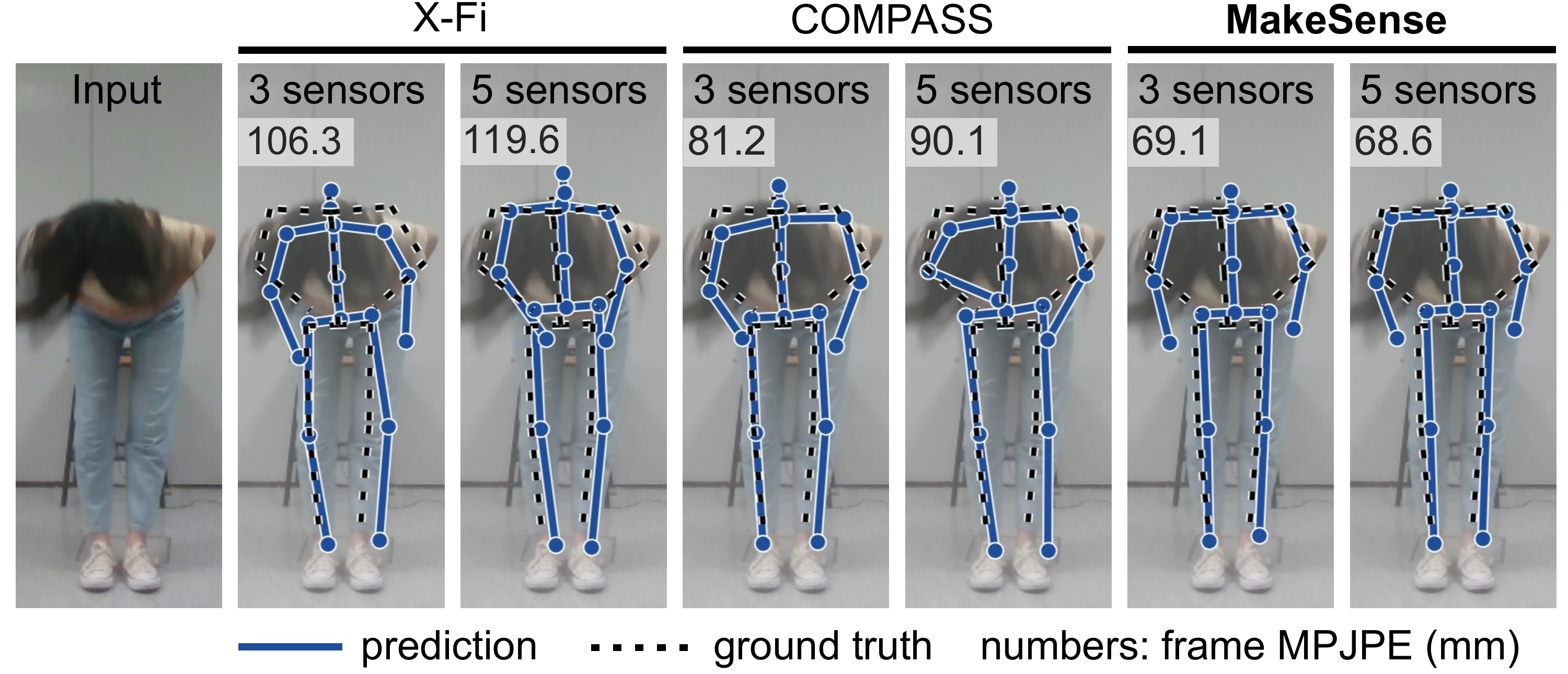}
\vspace{-0.5cm}
\caption{\textbf{Adding sensors can make a fusion model worse.}
On an MM-Fi in-distribution test frame, adding RGB \& mmWave to depth, LiDAR and WiFi CSI raises MPJPE for X-Fi ($106.3$ to $119.6$) and COMPASS ($81.2$ to $90.1$), but not ours ($69.1$ to $68.6$).
}
\label{fig:why}
\vspace{-0.3cm}
\end{figure}

On the MM-Fi test frame in Fig.~\ref{fig:why}, adding RGB and mmWave
to depth, LiDAR, and WiFi channel state information (CSI) raises pose
error for X-Fi and COMPASS, while slightly lowering it for \ours{}.
Scores averaged across sensor subsets can hide these effects.

\begin{figure*}[t]
\centering
\vspace{-0.7cm}
\includegraphics[width=\textwidth]{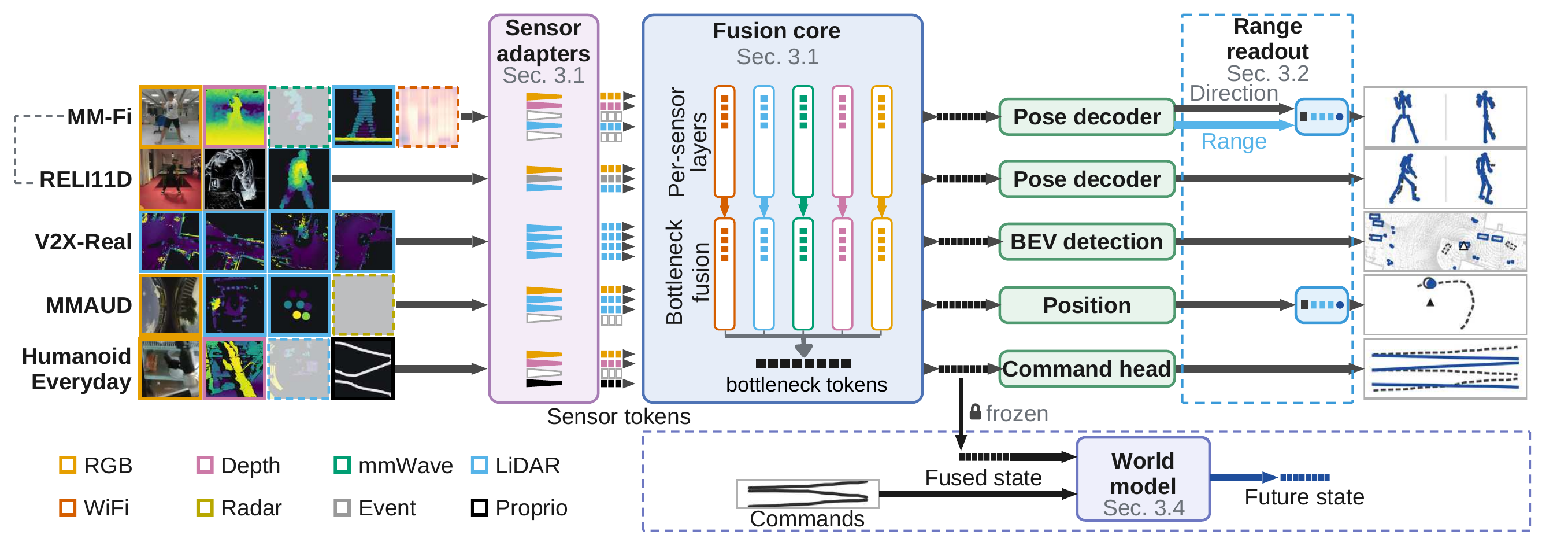}
\vspace{-0.5cm}
\caption{\textbf{\ours{} overview.}
Each present sensor passes through its own adapter, while a missing sensor (faded) contributes no token. The fusion core
first processes every sensor separately, then mixes them through shared bottleneck tokens. Dataset heads read the
fused tokens, and range-measuring sensors set the distance. A world model on the frozen fused state forecasts its
future from robot commands. 
}
\label{fig:overview}
\vspace{-0.3cm}
\end{figure*}

\ours{} (Fig.~\ref{fig:overview}) encodes only the sensors present.
Sensors keep private token streams, which interact only through eight shared bottleneck tokens, without a completion network.
In training, we apply sensor-specific degradations and fuse two views of each sample: one with all available sensors and one with a random subset. In MM-Fi, RGB and WiFi CSI can provide pose cues without directly measuring target range. On Humanoid-Everyday, a world model uses the frozen fusion representation to forecast future fused states.
Our contributions are:
\begin{itemize} 
\item \textbf{Paired protocol for negative fusion and degraded sensors.}
We quantify the harm of adding a sensor while holding the model, test
samples, and existing inputs fixed. We measure how often and
by how much error rises across all 75 single-sensor additions over 31
nonempty subsets in familiar and unseen environments. We also compare errors when retaining or dropping degraded sensors.
\item \textbf{Measurement-grounded multimodal fusion.}
\ours{} combines cues from the available sensors while restricting the
distance readout to range measurements where geometry permits. To
generalize the design across tasks and sensor suites, we train models with the same fusion core and dataset-specific adapters and heads.
\item \textbf{Thorough evaluation across tasks, models, and sensors.}
We measure negative fusion across five datasets and multiple fusion
models. Training with the range readout
reduces the penalty of adding RGB and WiFi CSI on both MM-Fi splits.
On MMAUD and Humanoid-Everyday, keeping degraded LiDAR and biased
proprioception yields lower error than dropping them. \ours{} is more accurate than every baseline. On V2X-Real and RELI11D,
any performance loss from a tested sensor addition stays within only 2\%.
\end{itemize} 

\section{Related Work}
\paragraph{Multimodal fusion and negative fusion.}
Fusion architectures control how complementary cues interact through bottleneck tokens, masked streams, or fixed-size latents~\cite{nagrani2021mbt,recasens2023zorro,jaegle2022perceiverio}. Applications span pose estimation~\cite{chen2025xfi,fan2024mmdiff}, detection in a shared bird's-eye view (BEV)~\cite{attfuse,xu2022v2xvit,heal,bevfusion},
UAV tracking~\cite{mmudct,unlidar}, and LiDAR motion capture~\cite{li2022lidarcap,Yan_2024_CVPR}. Yet joint training can underuse or overfit modalities and underperform the best unimodal model~\cite{wang2020makes,peng2022ogmge,huang2022modality}.
With fixed training sensor sets, AWC balances learning and FlexPose selects inputs~\cite{awc,flexpose}. SimMLM penalizes regressions from added inputs and reports their pairwise frequency as the \emph{counterintuitive rate}~\cite{li2025simmlm}. We retain signed error
changes for each single-sensor addition and condition.

\paragraph{Missing and degraded inputs.}
Missing inputs motivate subset training~\cite{neverova2016moddrop,bachmann2022multimae},
reconstruction or distillation~\cite{ma2021smil,wei2023mmanet},
feature separation~\cite{wang2023shaspec}, and adaptive Transformers
or prompts~\cite{ma2022robust,lee2023prompting}. COMPASS aligns masked
views to detached full-view targets and generates missing inputs with
frozen encoders~\cite{compass}, while \ours{} trains end to end on available inputs alone. Absence masks do not describe degradation. PTA distills from a noisy teacher~\cite{pta}, and driving methods address sensor failures~\cite{ge2023metabev,wang2024unibev,park2025mome}.
Corruption benchmarks test detectors trained on clean data~\cite{dong2023benchmarking},
while we train and test with sensor-specific degradations.
Even clean monocular RGB leaves absolute distance ambiguous~\cite{moon2019camera}.
Camera--radar detectors position image features using radar range~\cite{kim2023crn},
and \ours{} estimates target distance from available range measurements alone.

\paragraph{Shared cores across sensors and tasks.}
HPT shares a Transformer trunk across robot embodiments with embodiment-specific stems and heads~\cite{wang2024hpt}, and AnySat shares one model across Earth-observation sensors and resolutions~\cite{astruc2025anysat}. Unlike these methods, which share one set of weights, \ours{} shares the core design and training recipe, and a separate model with its own adapters and heads is trained on each dataset. Humanoid-Everyday evaluates action-generating policies such as Action Chunking with Transformers (ACT), Diffusion Policy and 3D Diffusion Policy~\cite{act,chi2023diffusionpolicy,dp3}, trained per task for closed-loop control. In contrast, we score offline command prediction.

\paragraph{Predictive world models.}
World models can forecast future states in a latent space. Methods such
as JEPA and V-JEPA~2~\cite{ijepa,assran2025vjepa2}, DreamerV3~\cite{hafner2025dreamerv3},
and MUVO~\cite{bogdoll2025muvo} form their latent states with their own
encoders. MRSSM and AquaJEPA study latent prediction when sensor
observations are missing~\cite{tremblay2021multimodal,gazzaev2026aquajepa}.
We study whether an existing fusion representation supports forecasting
by training predictors inspired by JEPA, DreamerV3, and DINO-WM~\cite{zhou2025dinowm}
on the fused states produced by a frozen fusion model.

\section{Methodology}\label{sec:method}
A dataset declares a set of sources $\mods$. At time $t$, the present set $\Omega_t\subseteq\mods$ holds the sources that are available and valid. For every nonempty $\Omega\subseteq\Omega_t$, one network with weights $\theta$ predicts $\hat y^{\Omega}=f_\theta(\{x_t^m\}_{m\in\Omega})$ without imputing absent inputs. Each dataset has its own head (17-joint 3D pose, UAV position, cooperative 3D boxes, or the next 15 commands of 28 dimensions), so the datasets share the same architecture with a task-specific objective.

\subsection{\ours{} fusion core}\label{sec:core}
    Each present source $m$ is encoded by a trainable encoder $E_m$ and resampled to 16 tokens $H_m^{(0)}=R_m(A_m(E_m(x^m)))$ of width 256, where $A_m$ adds a modality embedding and a position code and $R_m$ is a cross-attention from 16 learned queries. An absent modality has no tokens: it is neither a query nor a key anywhere. The core has $L=6$ pre-LN Transformer layers $\Phi_\ell$ that share weights across modalities and keep one LayerNorm per modality. $\Phi_\ell(Q;K)$ lets queries $Q$ read keys and values $K$. In the first $L_p=3$ layers each sensor's tokens attend only to themselves. Eight learned bottleneck tokens $Z$ then join, and for $\ell>L_p$
\begin{equation}
\begin{aligned}
H_m^{(\ell)}&=\Phi_\ell\big(H_m^{(\ell-1)};H_m^{(\ell-1)}\big),\quad m\in\Omega,\\
Z^{(\ell)}&=\Phi_\ell\big(Z^{(\ell-1)};[Z^{(\ell-1)};H_\Omega^{(\ell-1)}]\big),
\end{aligned}
\label{eq:core}
\end{equation}
with $H_\Omega=[H_m]_{m\in\Omega}$. Modality tokens never read $Z$ or another modality. A two-layer query decoder with task queries $Q_y$ reads the last layer, $\hat y^{\Omega}=D(Q_y,[Z;H_\Omega])$. A linear head $\hat y_Z=g_Z(Z)$ and one auxiliary head $\hat y_m=g_m(E_m(x^m))$ per modality supervise the core and the encoders and are unused at inference. Other datasets change only the adapters and the head (Supplementary, Sec.~A.1).

\subsection{Range readout}\label{sec:readout}
Depth, mmWave and LiDAR ($\mods_r$) measure range; RGB and WiFi CSI do not. For a fused set $\Omega$ with range sensors $\Omega_r=\Omega\cap\mods_r\neq\emptyset$, the core runs once more on the same tokens restricted to $\Omega_r$, and the target's distance comes from that pass. For the root position $p$ on MM-Fi,
\begin{equation}
\hat p=\big\|\hat p^{\,\Omega_r}\big\|\,\frac{\hat p^{\,\Omega}}{\|\hat p^{\,\Omega}\|},\qquad \hat y=\hat y^{\Omega}-\hat p^{\,\Omega}+\hat p,
\label{eq:readout}
\end{equation}
with norms measured from the origin of the annotation frame. If $\Omega_r=\emptyset$, then we set $\hat y=\hat y^{\Omega}$. On MMAUD the position is read from the point clouds. V2X-Real uses an aligned detection readout trained with the core, and Humanoid-Everyday has no range target.

\begin{figure}[t]
\includegraphics[width=\linewidth]{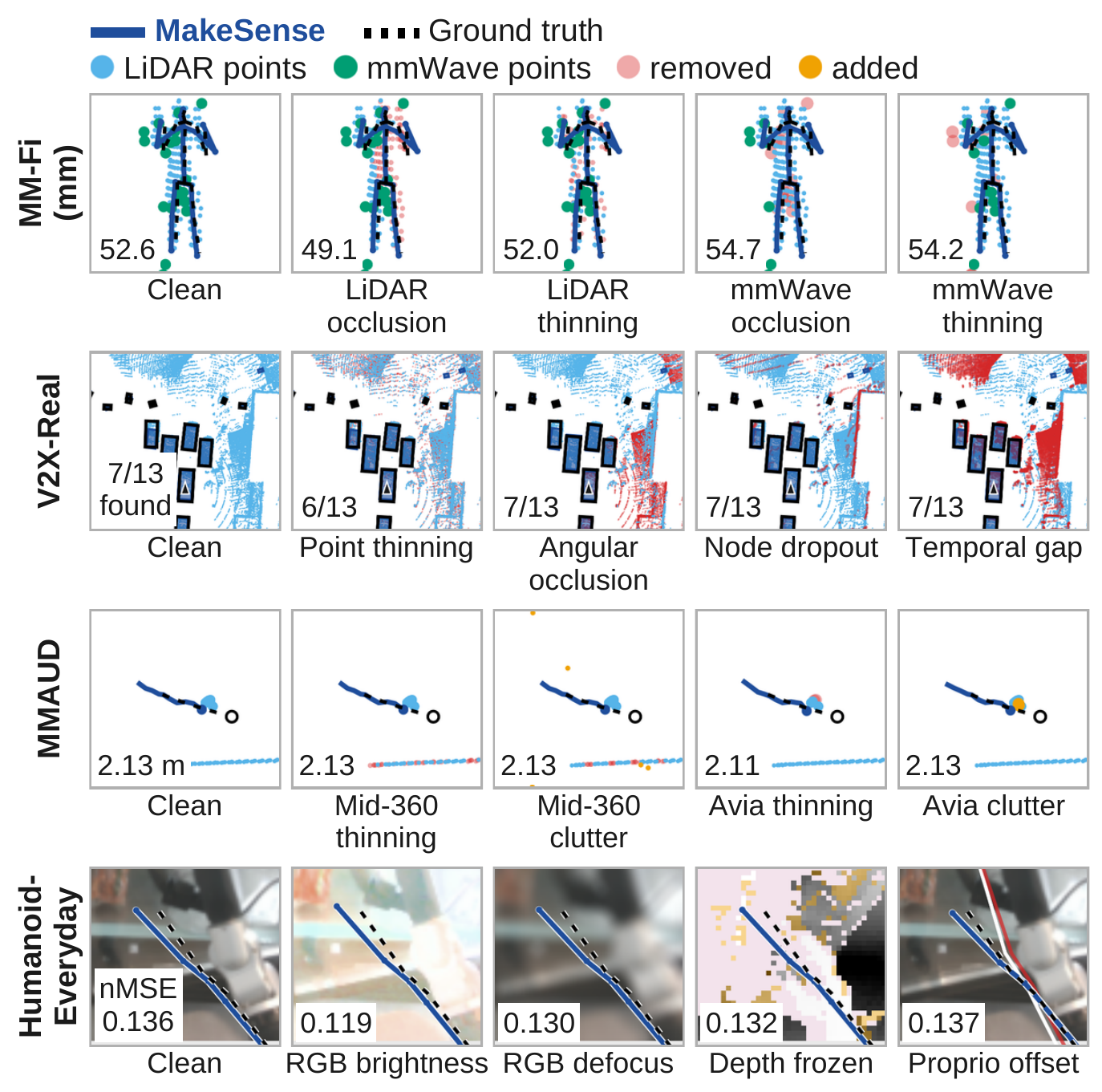}
\vspace{-0.3cm}
\caption{
\textbf{\ours{} under test-time sensor degradation.}
Each degraded panel applies one fault at strength 0.5 (numbers: error of the shown frame; V2X-Real: objects found). In the last panel, \textcolor{figred}{red} is the arm pose reported by the offset encoders and white the actual arm pose.
Table~\ref{tab:robust} gives the means.
}
\label{fig:degradation}
\vspace{-0.3cm}
\end{figure}

\subsection{Training with degradation}\label{sec:degradation}
Each update first applies the range shift $\mathcal T$ (when enabled) and degradation $\mathcal D$ to the raw inputs, yielding $\tilde x=\mathcal D(\mathcal T(x))$. The inputs are encoded once, and the same tokens are fused in two views: view A uses all present sensors $\Omega_A$, while view B uses a subset $\Omega_B\subseteq\Omega_A$. (all of $\Omega_A$ with probability 0.2, one sensor with probability 0.3, otherwise a uniform subset of intermediate size).

The degradation operator $\mathcal D$ is defined by sensor type (Fig.~\ref{fig:teaser}, left; Fig.~\ref{fig:degradation}). It applies \textit{occlusion rectangles and noise} to RGB, \textit{invalid regions and multiplicative noise} for depth, \textit{dropped cells and noise} for CSI, \textit{point thinning, angular occlusion and clutter} for point clouds, and \textit{dropout and stale frames} for streams; temporal gaps and node dropout are introduced only at test time. On MM-Fi half of the samples stay clean. Otherwise each sensor stays clean or draws one family with equal probability, at a strength whose cap rises to 0.8 (Supplementary, Sec.~A.3 lists all parameters). On MM-Fi every RGB image is also rescaled about its center by a factor $s\sim\mathcal U[0.8,1.25]$ and cropped or padded back, with the label unchanged (image-scale augmentation). $\mathcal T$ draws $u\sim\mathcal U[-0.3,0.3]$ for half of the samples and moves the target by $u$ along the camera-distance axis in the label and in the depth image, LiDAR returns and mmWave points, leaving RGB and CSI unchanged, so the shifted distance can only be learned from the range sensors.
On a shifted sample, a view or auxiliary head without a range sensor is not supervised along that axis and is left out of the alignment loss; subsets without a range sensor show no measurable loss of accuracy (Supplementary, Sec.~A.4). $\mathcal{T}$ requires a declared range axis and frame checks. It is used on MM-Fi and, as a radial shift, on MMAUD, but is turned off on V2X-Real, where frame checks fail.

With $\ell_{\rm mse}$ the mean squared error, $\rho(z)=z/\sqrt{\overline{z^2}+\epsilon}$ the root-mean-square normalization of each token and $\sg$ the stop-gradient,
\begin{equation}
\begin{aligned}
\mathcal L={}&\textstyle\sum_{v\in\{A,B\}}\big[\ell_{\rm mse}(\hat y^{v},y)+\tfrac12\ell_{\rm mse}(\hat y_Z^{v},y)\big]\\
&\textstyle+\sum_{m\in\Omega_A}\ell_{\rm mse}(\hat y_m,y)+\lambda\,\overline{\big(\rho(Z^{B})-\sg\rho(Z^{A})\big)^2}\\
&+\lambda\big[\ell_{\rm mse}(\hat y^{A},y)-\sg\ell_{\rm mse}(\hat y^{B},y)\big]_++\lambda\,\mathcal L_{\rm var}.
\end{aligned}\label{eq:loss}
\end{equation}
The three $\lambda$ terms align the subset view to the full view ($\mathcal L_{\rm KD}$), penalize a full-view loss above the subset-view loss ($\mathcal L_{\rm MoFe}$~\cite{li2025simmlm}), and prevent collapse with a hinge on the batch standard deviation of each bottleneck dimension ($\mathcal L_{\rm var}$~\cite{bardes2022vicreg}). We fix $\lambda=0.1$ before training and train MM-Fi models for 30k updates with a batch size of 32. At inference, the present inputs are encoded once without applying $\mathcal D$ or $\mathcal T$.

\subsection{World model on the fused state}\label{sec:wm}
Because an absent sensor contributes no tokens, \ours{} yields a fixed-size fused state $\rho(Z_t)$ at every step. After training, we freeze the core and fit a predictor $P$ that reads the state of a context frame and the logged commands,
\begin{equation}
\hat\rho_{t+k}=P\big(\rho(Z_t),a_{t:t+k}\big),\quad \mathcal L_P=\textstyle\sum_{k}\overline{\big(\hat\rho_{t+k}-\rho(Z_{t+k})\big)^2},
\label{eq:wm}
\end{equation}
for $k=1,\dots,5$ steps of 0.2\,s. The target is the frozen state on the clean frame $t+k$ with every sensor present. With the same information and budget, we fit a block-causal Transformer as in V-JEPA~2-AC~\cite{assran2025vjepa2}, with and without commands, DreamerV3's recurrent state-space model (RSSM)~\cite{hafner2025dreamerv3} and a DINO-WM-style predictor. The last state and a linear autoregressive model with command inputs serve as references (Supplementary, Sec.~A.5).

\section{Experimental Setting}
\begin{table}[t]
\centering\footnotesize
\setlength{\tabcolsep}{2.0pt}
\caption{\textbf{Datasets and splits.} $K$: number of input subsets. R, D, M, L, W: RGB, depth, mmWave, LiDAR, WiFi CSI.}
\label{tab:datasets}
\vspace{-0.3cm}
\begin{tabular}{@{}llll@{}}
\toprule
Dataset & Task / inputs & Train\,/\,valid\,/\,test & Metrics \\
\midrule
MM-Fi~\cite{yang2023mmfi} & 3D pose (17 joints) & 24\,/\,3\,/\,3 subjects & MPJPE\dn \\
$K=31$ & R, D, M, L, W & + E04 (10 subjects) & PA-MPJPE\dn \\
\addlinespace[2pt]
MMAUD~\cite{yuan2024mmaud} & UAV position & 102\,/\,16\,/\,59 sequences & APE\dn \\
$K=15$ & R, 2 L, radar & & Coverage\up \\
\addlinespace[2pt]
HE~\cite{zhao2025humanoideveryday} & next 15 commands & 87\,/\,12\,/\,15 episodes & nMSE\dn \\
$K=15$ & R, D, L, proprio & & MAE\dn \\
\addlinespace[2pt]
V2X-Real~\cite{xiang2024v2xreal} & coop.\ 3D detection & 5,769\,/\,717\,/ & AP50\up \\
$K=8$ & 4 nodes (L) & 1,253 frames & AP70\up \\
\addlinespace[2pt]
RELI11D~\cite{Yan_2024_CVPR} & 3D pose (24 joints) & 1--100\% of 1,131\,/ & MPJPE\dn \\
$K=7$ & R/event features, L & 165\,/\,225 windows & PA-MPJPE\dn \\
\bottomrule
\end{tabular}
\vspace{-0.3cm}
\end{table}

Table~\ref{tab:datasets} lists the datasets. MM-Fi has subject-disjoint splits with an in-distribution (ID) test and an unseen-environment test (E04, out-of-distribution, OOD). V2X-Real follows the QuantV2X~\cite{zhao2026quantv2x} split, and its eight subsets keep the ego node. RELI11D uses 16-frame windows of a same-subject time split. Humanoid-Everyday (HE) holds out the last episodes of each task's recording session for validation and test. Models are selected on the validation split. 

\paragraph{Negative-fusion protocol.} Let $e(\Omega)$ be the test error of one trained model that reads the sensor subset $\Omega$. Adding a sensor $m\notin\Omega$ changes it by \mbox{$\delta(\Omega,m)=e(\Omega\cup\{m\})/e(\Omega)-1$} on the same test samples, and the addition is harmful when $\delta>0$. Neg.\ counts the harmful additions (out of 75 on MM-Fi), and $G$ is the change when RGB and WiFi CSI are both added to depth, mmWave and LiDAR. Both also fall when small subsets become worse, so we read them next to Avg.\,($K$), the mean error over all $K$ subsets. SimMLM's counterintuitive rate averages over samples and pairs~\cite{li2025simmlm}. We keep each addition because the average hides which one hurts or helps.

All baselines are retrained on our splits with their published recipes (budgets and deviations in Supplementary, Sec.~B.2). On MM-Fi, they are X-Fi~\cite{chen2025xfi}, COMPASS~\cite{compass} and a late fusion of single-sensor experts, and on MMAUD they are MMUDCT~\cite{mmudct} and UnLiDAR~\cite{unlidar}. On V2X-Real, late fusion, AttFuse~\cite{attfuse}, HEAL~\cite{heal} and its Pyramid Fusion subnetwork fuse the nodes within QuantV2X's 70\,m range, whereas \ours{} fuses every present node. On RELI11D they are LEIR~\cite{Yan_2024_CVPR} and LiDARCap~\cite{li2022lidarcap}. On HE, they are ACT~\cite{act}, Diffusion Policy~\cite{chi2023diffusionpolicy}, late fusion, MBT~\cite{nagrani2021mbt}, and Perceiver IO~\cite{jaegle2022perceiverio} (reference predictors in Supplementary, Sec.~B.3); ACT and Diffusion Policy read one RGB camera and the joint state, and their missing-input results are zero-shot. 

\begin{figure*}[t]
\vspace{-0.7cm}
\includegraphics[width=\linewidth]{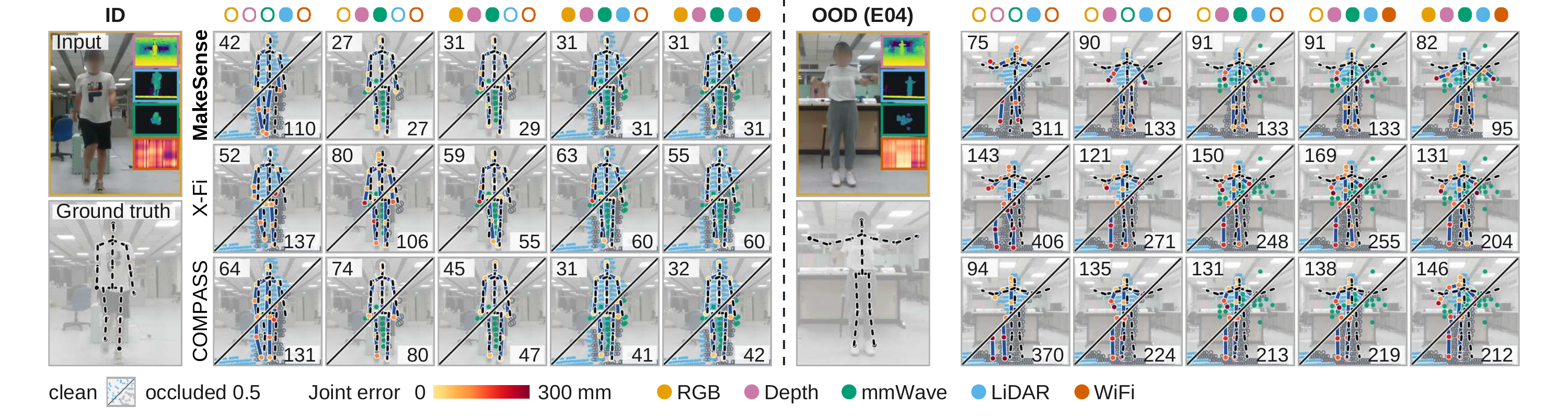}\\[-1pt]
\includegraphics[width=\linewidth]{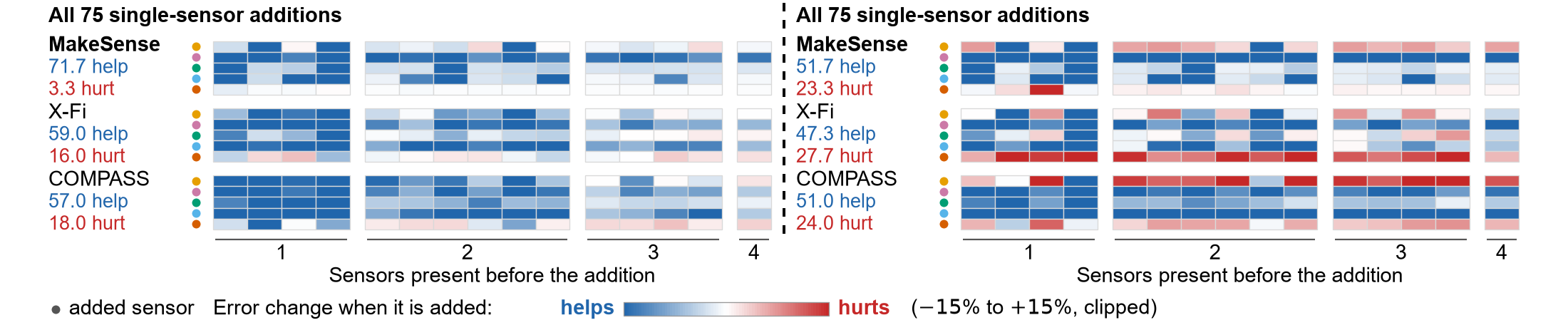}
\vspace{-0.7cm}
\caption{
\textbf{MM-Fi pose estimation (1--5 sensors), on an ID and an OOD (E04 room) test.} Each column uses the best subset of its size (filled dots mark the sensors used).
\textcolor{figblue}{Dark blue}: predictions with clean inputs or with LiDAR occluded at strength 0.5 (mmWave if LiDAR is absent). Numbers: frame MPJPE (mm). The frames are examples on which \ours{} is more accurate
than X-Fi with all 5 sensors. Table~\ref{tab:main} gives the means.
Bottom: error change for each of the 75 single-sensor additions (rows: added sensor; \textcolor{figblue}{blue}: helps, \textcolor{figred}{red}: hurts).
}
\label{fig:mmfi}
\vspace{-0.3cm}
\end{figure*}

\begin{figure*}[t]
\includegraphics[width=\linewidth]{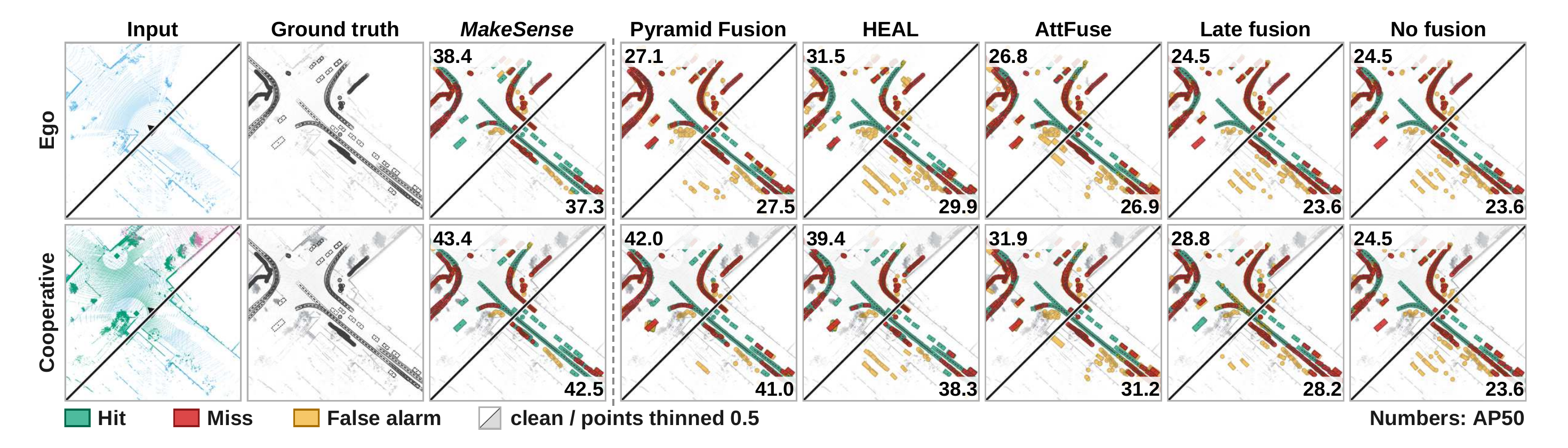}
\vspace{-0.7cm}
\caption{
\textbf{V2X-Real cooperative 3D detection.}
Box trails span one 10.6\,s clip (ego vs.\ 4 nodes). AP50: full test set, one seed (Table~\ref{tab:main}: 3-seed means); lower right: 50\%
of points. Pyramid Fusion is HEAL's fusion subnetwork~\cite{heal},
trained end to end without alignment.
}
\label{fig:v2x}
\vspace{-0.3cm}
\end{figure*}

\section{Results}
\begin{table}[t]
\centering\footnotesize
\setlength{\tabcolsep}{2.0pt}
\caption{
\textbf{Main results on 5 datasets.}
Mean\,$\pm$\,SD over 3 seeds; entries without $\pm$ are from 1 run. Avg.\,($K$) is the mean over all $K$ input subsets. Coverage: share of test steps with an output. $\Delta$ gives the gain of \ours{} on the first metric (\%; AP in points). Colors: gap to the column best (\sevkey{sevbest}{best}, \sevkey{sevok}{$\leq$20\%}, \sevkey{sevmid}{20--50\%}, \sevkey{sevbad}{$>$50\%}).
}
\label{tab:main}
\begin{tabular}{@{}l ccc c@{}}
\toprule
\multicolumn{1}{@{}l}{\textbf{MM-Fi ID {\scriptsize(mm)}}} & {\scriptsize MPJPE\dn} & {\scriptsize PA-MPJPE\dn} & {\scriptsize Avg.\,(31)\dn} & {\scriptsize $\Delta$} \\
\midrule
Late fusion & \sev{sevmid}{86.8} & \sev{sevmid}{52.3} & \sev{sevmid}{97.3} & \gain{+33\%} \\
X-Fi~\cite{chen2025xfi} & \sev{sevmid}{82.1\pmsd{4.8}} & \sev{sevmid}{49.3\pmsd{1.9}} & \sev{sevmid}{97.3\pmsd{2.9}} & \gain{+29\%} \\
COMPASS~\cite{compass} & \sev{sevmid}{80.5\pmsd{10.2}} & \sev{sevmid}{51.9\pmsd{7.3}} & \sev{sevmid}{100.5\pmsd{26.6}} & \gain{+27\%} \\
\ours{} & \sev{sevbest}{\textbf{58.4}\pmsd{1.4}} & \sev{sevbest}{\textbf{39.1}\pmsd{0.2}} & \sev{sevbest}{\textbf{72.7}\pmsd{1.7}} &  \\
\midrule
\multicolumn{1}{@{}l}{\textbf{MM-Fi OOD {\scriptsize(E04, mm)}}} & {\scriptsize MPJPE\dn} & {\scriptsize PA-MPJPE\dn} & {\scriptsize Avg.\,(31)\dn} &  \\
\midrule
Late fusion & \sev{sevbad}{181.0} & \sev{sevmid}{75.6} & \sev{sevbad}{190.4} & \gain{+56\%} \\
X-Fi~\cite{chen2025xfi} & \sev{sevbad}{133.7\pmsd{18.4}} & \sev{sevmid}{69.3\pmsd{2.5}} & \sev{sevmid}{155.3\pmsd{8.8}} & \gain{+40\%} \\
COMPASS~\cite{compass} & \sev{sevbad}{149.2\pmsd{9.4}} & \sev{sevmid}{77.3\pmsd{8.8}} & \sev{sevbad}{180.8\pmsd{30.8}} & \gain{+46\%} \\
\ours{} & \sev{sevbest}{\textbf{80.2}\pmsd{0.7}} & \sev{sevbest}{\textbf{52.5}\pmsd{0.8}} & \sev{sevbest}{\textbf{118.1}\pmsd{3.7}} &  \\
\midrule
\multicolumn{1}{@{}l}{\textbf{MMAUD}} & {\scriptsize APE {\scriptsize(m)}\dn} & {\scriptsize Coverage {\scriptsize(\%)}\up} & {\scriptsize Avg.\,(15)\dn} &  \\
\midrule
UnLiDAR~\cite{unlidar} & \sev{sevbad}{29.35} & -- & -- & \gain{+59\%} \\
MMUDCT~\cite{mmudct} & \sev{sevmid}{17.48\pmsd{0.87}} & \sev{sevmid}{51.0\pmsd{3.7}} & -- & \gain{+31\%} \\
\ours{} & \sev{sevbest}{\textbf{11.99}\pmsd{0.25}} & \sev{sevbest}{\textbf{100}\pmsd{0.0}} & \sev{sevbest}{\textbf{20.39}\pmsd{0.22}} &  \\
\midrule
\multicolumn{1}{@{}l}{\textbf{Humanoid-Everyday}} & {\scriptsize nMSE\dn} & {\scriptsize MAE\dn} & {\scriptsize Avg.\,(15)\dn} &  \\
\midrule
Late fusion & \sev{sevok}{0.180\pmsd{.001}} & \sev{sevok}{0.0605\pmsd{.0002}} & \sev{sevbad}{0.793\pmsd{.003}} & \gain{+8\%} \\
MBT~\cite{nagrani2021mbt} & \sev{sevok}{0.185\pmsd{.004}} & \sev{sevok}{0.0586\pmsd{.0008}} & \sev{sevbad}{2.186\pmsd{.060}} & \gain{+10\%} \\
Perceiver IO~\cite{jaegle2022perceiverio} & \sev{sevmid}{0.210\pmsd{.008}} & \sev{sevok}{0.0639\pmsd{.0015}} & \sev{sevbad}{0.895\pmsd{.062}} & \gain{+21\%} \\
\ours{} & \sev{sevbest}{\textbf{0.166}\pmsd{.005}} & \sev{sevbest}{\textbf{0.0558}\pmsd{.0007}} & \sev{sevbest}{\textbf{0.390}\pmsd{.002}} &  \\
\midrule
\multicolumn{1}{@{}l}{\textbf{V2X-Real}} & {\scriptsize AP50\up} & {\scriptsize AP70\up} & {\scriptsize Avg.\,(8)\up} &  \\
\midrule
Late fusion & \sev{sevmid}{29.3\pmsd{1.2}} & \sev{sevbad}{12.8\pmsd{0.7}} & \sev{sevmid}{27.7\pmsd{1.2}} & \gain{+15.1} \\
AttFuse~\cite{attfuse} & \sev{sevmid}{30.8\pmsd{1.7}} & \sev{sevbad}{14.4\pmsd{2.8}} & \sev{sevmid}{28.6\pmsd{1.2}} & \gain{+13.6} \\
HEAL~\cite{heal} & \sev{sevok}{39.4\pmsd{0.2}} & \sev{sevmid}{23.0\pmsd{1.7}} & \sev{sevok}{35.8\pmsd{0.2}} & \gain{+5.0} \\
Pyramid Fusion~\cite{heal} & \sev{sevok}{40.7\pmsd{1.1}} & \sev{sevok}{24.8\pmsd{1.3}} & \sev{sevok}{35.4\pmsd{1.0}} & \gain{+3.7} \\
\ours{} & \sev{sevbest}{\textbf{44.4}\pmsd{1.2}} & \sev{sevbest}{\textbf{29.3}\pmsd{0.5}} & \sev{sevbest}{\textbf{42.6}\pmsd{1.2}} &  \\
\quad 70\,m range & \sev{sevok}{44.2\pmsd{1.2}} & \sev{sevok}{29.1\pmsd{0.6}} & \sev{sevok}{42.4\pmsd{1.2}} &  \\
\midrule
\multicolumn{1}{@{}l}{\textbf{RELI11D {\scriptsize(mm)}}} & {\scriptsize PA-MPJPE\dn} & {\scriptsize MPJPE\dn} & {\scriptsize Avg.\,(7)\dn} &  \\
\midrule
LEIR~\cite{Yan_2024_CVPR} & \sev{sevbad}{62.4\pmsd{0.1}} & \sev{sevbad}{129.5\pmsd{1.2}} & -- & \gain{+43\%} \\
LiDARCap~\cite{li2022lidarcap} & \sev{sevmid}{52.2\pmsd{1.2}} & -- & -- & \gain{+32\%} \\
\ours{} & \sev{sevbest}{\textbf{35.7}\pmsd{0.4}} & \sev{sevbest}{\textbf{42.7}\pmsd{0.8}} & \sev{sevbest}{\textbf{56.4}\pmsd{0.4}} &  \\
\quad 10\% labels & 68.2\pmsd{0.7} & 91.9\pmsd{0.4} & -- &  \\
\quad\; + MM-Fi init. & 64.4\pmsd{1.1} & 87.0\pmsd{0.9} & -- &  \\
\bottomrule
\end{tabular}
\end{table}

\subsection{Negative fusion on MM-Fi}\label{sec:negfusion}
\ours{} has the least negative fusion. In distribution, 72 of its 75 single-sensor additions lower the error and only 3 raise it, against 16 for X-Fi and 18 for COMPASS (Table~\ref{tab:ablation} and Fig.~\ref{fig:mmfi}, bottom). No addition raises its error in all three seeds, against 4 for X-Fi and 7 for COMPASS (Supplementary, Sec.~C.1). With all five sensors, \ours{} is as accurate as with the three range sensors, while both baselines lose accuracy. In the unseen room (E04), \ours{} has the smallest gap $G$. Its additions that raise the error by more than 2\% are about half as many as those of the baselines (Table~\ref{tab:ablation}, right). Fusion also helps on average. The mean error of all three models falls with each added sensor (Supplementary, Sec.~C.5).

\begin{table}[t]
\centering
\scriptsize
\setlength{\tabcolsep}{1.5pt}\renewcommand{\arraystretch}{1.0}
\caption{\textbf{Robustness to test-time sensor degradation (strength 0.5).} 
$\Delta$: mean worsening over the 4 faults.
Sensor rm: the sensor is dropped.
deg.: degradation training; scale: image-scale augmentation.
V2X-Real baselines use a 70\,m ego range.
Mean\,$\pm$\,SD over 3 seeds; $^\dagger$fewer seeds (no $\pm$: one). Bold: best per column.
Tint: worsening relative to clean (\sevkey{sevbest}{$<$2\%}, \sevkey{sevok}{$\geq$2\%}, \sevkey{sevmid}{$\geq$5\%}, \sevkey{sevbad}{$\geq$15\%}).
}
\label{tab:robust}
\resizebox{\columnwidth}{!}{%
\begin{tabular}{@{}l rrrrr r@{}}
\toprule
\shortstack[l]{\textbf{MM-Fi ID}\\{\tiny(MPJPE, mm\dn)}} & \hd{Clean} & \hd{\shortstack{LiDAR\\occl.}} & \hd{\shortstack{LiDAR\\thin.}} & \hd{\shortstack{mmWave\\occl.}} & \hd{\shortstack{mmWave\\thin.}} & $\Delta$\%\dn \\
\midrule
X-Fi & 82.1\pmsdt{4.8} & \cellcolor{sevbad}104.5\pmsdt{2.3} & \cellcolor{sevok}84.0\pmsdt{5.6} & \cellcolor{sevbest}83.0\pmsdt{5.0} & \cellcolor{sevbest}82.7\pmsdt{5.1} & \cellcolor{sevmid}+7.9\pmsdt{0.9} \\
X-Fi\,+\,deg., scale & 79.6\pmsdt{7.9} & \cellcolor{sevmid}88.2\pmsdt{10.4} & \cellcolor{sevok}82.5\pmsdt{8.3} & \cellcolor{sevbest}80.6\pmsdt{9.0} & \cellcolor{sevbest}80.2\pmsdt{8.7} & \cellcolor{sevok}+4.1\pmsdt{1.4} \\
COMPASS & 80.5\pmsdt{10.2} & \cellcolor{sevbad}94.2\pmsdt{11.8} & \cellcolor{sevok}83.1\pmsdt{12.5} & \cellcolor{sevbest}81.1\pmsdt{9.9} & \cellcolor{sevbest}80.8\pmsdt{10.0} & \cellcolor{sevmid}+5.4\pmsdt{1.2} \\
\textbf{\ours{}} & \textbf{58.4}\pmsdt{1.4} & \cellcolor{sevbest}\textbf{59.1}\pmsdt{1.3} & \cellcolor{sevbest}\textbf{58.7}\pmsdt{1.4} & \cellcolor{sevbest}\textbf{59.1}\pmsdt{1.3} & \cellcolor{sevbest}\textbf{58.8}\pmsdt{1.4} & \cellcolor{sevbest}\textbf{+0.9}\pmsdt{0.1} \\
\quad\ctx{sensor rm} & -- & \cellcolor{sevbest}58.8\pmsdt{1.3} & \cellcolor{sevbest}58.8\pmsdt{1.3} & \cellcolor{sevok}59.9\pmsdt{1.1} & \cellcolor{sevok}59.9\pmsdt{1.1} & -- \\
\quad\ctx{w/o deg., scale} & 63.3\pmsdt{2.3} & \cellcolor{sevmid}69.8\pmsdt{2.6} & \cellcolor{sevok}65.5\pmsdt{2.2} & \cellcolor{sevbest}64.0\pmsdt{2.3} & \cellcolor{sevbest}63.6\pmsdt{2.3} & \cellcolor{sevok}+3.8\pmsdt{0.5} \\
\midrule
\shortstack[l]{\textbf{MM-Fi OOD}\\{\tiny(MPJPE, mm\dn)}} & \hd{Clean} & \hd{\shortstack{LiDAR\\occl.}} & \hd{\shortstack{LiDAR\\thin.}} & \hd{\shortstack{mmWave\\occl.}} & \hd{\shortstack{mmWave\\thin.}} & $\Delta$\%\dn \\
\midrule
X-Fi & 133.7\pmsdt{18.4} & \cellcolor{sevbad}189.4\pmsdt{16.0} & \cellcolor{sevbest}134.2\pmsdt{19.1} & \cellcolor{sevok}137.5\pmsdt{18.6} & \cellcolor{sevok}136.6\pmsdt{18.5} & \cellcolor{sevmid}+11.9\pmsdt{2.4} \\
X-Fi\,+\,deg., scale & 114.4\pmsdt{5.0} & \cellcolor{sevbad}151.8\pmsdt{7.1} & \cellcolor{sevbest}114.2\pmsdt{7.6} & \cellcolor{sevok}118.1\pmsdt{6.8} & \cellcolor{sevok}117.3\pmsdt{6.1} & \cellcolor{sevmid}+9.6\pmsdt{1.7} \\
COMPASS & 149.2\pmsdt{9.4} & \cellcolor{sevbad}179.7\pmsdt{12.0} & \cellcolor{sevbest}144.5\pmsdt{12.8} & \cellcolor{sevbest}150.8\pmsdt{9.6} & \cellcolor{sevbest}150.7\pmsdt{9.5} & \cellcolor{sevok}+4.9\pmsdt{0.9} \\
\textbf{\ours{}} & \textbf{80.2}\pmsdt{0.7} & \cellcolor{sevmid}\textbf{87.3}\pmsdt{2.8} & \cellcolor{sevbest}\textbf{80.5}\pmsdt{0.5} & \cellcolor{sevbest}\textbf{80.8}\pmsdt{0.7} & \cellcolor{sevbest}\textbf{80.3}\pmsdt{0.8} & \cellcolor{sevok}\textbf{+2.5}\pmsdt{1.0} \\
\quad\ctx{sensor rm} & -- & \cellcolor{sevbest}81.8\pmsdt{0.8} & \cellcolor{sevbest}81.8\pmsdt{0.8} & \cellcolor{sevbest}81.5\pmsdt{0.9} & \cellcolor{sevbest}81.5\pmsdt{0.9} & -- \\
\quad\ctx{w/o deg., scale} & 135.9\pmsdt{14.3} & \cellcolor{sevmid}150.5\pmsdt{14.9} & \cellcolor{sevbest}137.5\pmsdt{15.1} & \cellcolor{sevbest}136.6\pmsdt{14.1} & \cellcolor{sevbest}136.1\pmsdt{14.2} & \cellcolor{sevok}+3.2\pmsdt{0.5} \\

\midrule
\shortstack[l]{\textbf{V2X-Real}\\{\tiny(AP50\up)}} & \hd{Clean} & \hd{\shortstack{Point\\thin.}} & \hd{\shortstack{Angular\\occl.}} & \hd{\shortstack{Node\\dropout}} & \hd{\shortstack{Temporal\\gap}} & $\Delta$\%\dn \\
\midrule
HEAL & 39.4\pmsdt{0.2} & \cellcolor{sevbest}38.7\pmsdt{0.4} & \cellcolor{sevbad}26.6\pmsdt{0.6} & \cellcolor{sevok}37.8\pmsdt{0.4} & \cellcolor{sevmid}35.3\pmsdt{0.6} & \cellcolor{sevmid}+12.3\pmsdt{1.0} \\
Pyramid Fusion & 40.7\pmsdt{1.1} & \cellcolor{sevbest}40.2\pmsdt{0.8} & \cellcolor{sevbad}26.1\pmsdt{1.4} & \cellcolor{sevmid}35.9\pmsdt{0.6} & \cellcolor{sevmid}35.2\pmsdt{0.7} & \cellcolor{sevbad}+15.6\pmsdt{0.5} \\
AttFuse & 30.8\pmsdt{1.7} & \cellcolor{sevbest}30.4\pmsdt{1.2} & \cellcolor{sevbad}16.3\pmsdt{0.8} & \cellcolor{sevok}29.9\pmsdt{1.4} & \cellcolor{sevmid}27.9\pmsdt{1.0} & \cellcolor{sevbad}+15.1\pmsdt{2.7} \\
Late fusion & 29.3\pmsdt{1.2} & \cellcolor{sevok}28.7\pmsdt{1.3} & \cellcolor{sevbad}24.1\pmsdt{0.7} & \cellcolor{sevok}27.9\pmsdt{1.5} & \cellcolor{sevmid}27.8\pmsdt{1.2} & \cellcolor{sevmid}+7.5\pmsdt{0.3} \\
\textbf{\ours{}} & 44.4\pmsdt{1.2} & \cellcolor{sevok}\textbf{43.5}\pmsdt{1.2} & \cellcolor{sevbad}\textbf{35.4}\pmsdt{0.7} & \cellcolor{sevok}43.3\pmsdt{1.2} & \cellcolor{sevok}42.3\pmsdt{1.3} & \cellcolor{sevmid}\textbf{+7.3}\pmsdt{0.1} \\
\quad\ctx{w/o deg.$^\dagger$} & \textbf{44.6}\pmsdt{1.1} & \cellcolor{sevok}\textbf{43.5}\pmsdt{1.3} & \cellcolor{sevbad}32.4\pmsdt{1.1} & \cellcolor{sevok}\textbf{43.5}\pmsdt{1.2} & \cellcolor{sevok}\textbf{42.5}\pmsdt{1.0} & \cellcolor{sevmid}+9.3\pmsdt{0.9} \\
\midrule
\shortstack[l]{\textbf{MMAUD}\\{\tiny(APE, m\dn)}} & \hd{Clean} & \hd{\shortstack{Mid-360\\thin.}} & \hd{\shortstack{Mid-360\\clutter}} & \hd{\shortstack{Avia\\thin.}} & \hd{\shortstack{Avia\\clutter}} & $\Delta$\%\dn \\
\midrule
\textbf{\ours{}} & 11.99\pmsdt{0.25} & \cellcolor{sevmid}13.60\pmsdt{0.19} & \cellcolor{sevbad}\textbf{16.41}\pmsdt{0.47} & \cellcolor{sevok}12.31\pmsdt{0.24} & \cellcolor{sevbest}12.21\pmsdt{0.32} & \cellcolor{sevmid}\textbf{+13.7}\pmsdt{2.0} \\
\quad\ctx{sensor rm} & -- & \cellcolor{sevbad}20.28\pmsdt{0.36} & \cellcolor{sevbad}20.28\pmsdt{0.36} & \cellcolor{sevbad}21.03\pmsdt{0.56} & \cellcolor{sevbad}21.03\pmsdt{0.56} & -- \\
\quad\ctx{w/o deg.$^\dagger$} & \textbf{11.12} & \cellcolor{sevbad}\textbf{13.19} & \cellcolor{sevbad}20.17 & \cellcolor{sevok}\textbf{11.50} & \cellcolor{sevmid}\textbf{11.98} & \cellcolor{sevbad}+27.8 \\
\midrule
\grp{HE {\tiny(nMSE\dn)}} & \hd{Clean} & \hd{\shortstack{RGB\\bright.}} & \hd{\shortstack{RGB\\defocus}} & \hd{\shortstack{Depth\\frozen}} & \hd{\shortstack{Proprio\\offset}} & $\Delta$\%\dn \\
\midrule
ACT~\cite{act} & 0.175\pmsdt{.003} & \cellcolor{sevbad}0.214\pmsdt{.009} & \cellcolor{sevmid}0.190\pmsdt{.007} & \cellcolor{sevbest}0.175\pmsdt{.003} & \cellcolor{sevok}0.180\pmsdt{.002} & \cellcolor{sevmid}+8.3\pmsdt{0.9} \\
Diffusion Policy~\cite{chi2023diffusionpolicy} & 0.209\pmsdt{.009} & \cellcolor{sevmid}0.225\pmsdt{.006} & \cellcolor{sevbest}0.209\pmsdt{.009} & \cellcolor{sevbest}0.209\pmsdt{.009} & \cellcolor{sevok}0.214\pmsdt{.010} & \cellcolor{sevok}+2.5\pmsdt{0.7} \\
Late fusion & 0.180\pmsdt{.001} & \cellcolor{sevbest}0.183\pmsdt{.001} & \cellcolor{sevbest}0.180\pmsdt{.001} & \cellcolor{sevbest}0.179\pmsdt{.001} & \cellcolor{sevok}0.186\pmsdt{.001} & \cellcolor{sevbest}\textbf{+1.3}\pmsdt{0.1} \\
MBT~\cite{nagrani2021mbt} & 0.185\pmsdt{.004} & \cellcolor{sevbad}0.244\pmsdt{.013} & \cellcolor{sevok}0.189\pmsdt{.005} & \cellcolor{sevok}0.189\pmsdt{.005} & \cellcolor{sevok}0.192\pmsdt{.005} & \cellcolor{sevmid}+9.8\pmsdt{1.2} \\
Perceiver IO~\cite{jaegle2022perceiverio} & 0.210\pmsdt{.008} & \cellcolor{sevbad}0.405\pmsdt{.022} & \cellcolor{sevok}0.216\pmsdt{.010} & \cellcolor{sevbest}0.210\pmsdt{.008} & \cellcolor{sevmid}0.224\pmsdt{.008} & \cellcolor{sevbad}+25.5\pmsdt{4.4} \\
\textbf{\ours{}} & 0.166\pmsdt{.005} & \cellcolor{sevok}0.170\pmsdt{.008} & \cellcolor{sevbest}0.165\pmsdt{.005} & \cellcolor{sevbest}0.166\pmsdt{.005} & \cellcolor{sevok}0.172\pmsdt{.004} & \cellcolor{sevbest}\textbf{+1.3}\pmsdt{0.2} \\
\quad\ctx{sensor rm} & -- & \cellcolor{sevbest}0.166\pmsdt{.006} & \cellcolor{sevbest}0.166\pmsdt{.006} & \cellcolor{sevok}0.170\pmsdt{.006} & \cellcolor{sevbad}0.620\pmsdt{.015} & -- \\
\quad\ctx{w/o deg.} & \textbf{0.162}\pmsdt{.004} & \cellcolor{sevok}\textbf{0.167}\pmsdt{.005} & \cellcolor{sevbest}\textbf{0.162}\pmsdt{.004} & \cellcolor{sevbest}\textbf{0.162}\pmsdt{.005} & \cellcolor{sevok}\textbf{0.169}\pmsdt{.005} & \cellcolor{sevbest}+1.6\pmsdt{0.4} \\
\bottomrule
\end{tabular}}
\end{table}

\subsection{Why \ours{} reduces negative fusion}\label{sec:recipes}
Adding WiFi CSI to the three range sensors changes the error of \ours{} by less than 1\%, but raises the error of X-Fi and COMPASS (Fig.~\ref{fig:mmfi}, bottom). The remaining gap of \ours{} comes from RGB.

We split the error into the pose (PA-MPJPE) and the placement of the skeleton. In distribution, adding RGB lowers the pose error of our model. Without the readout or the range shift, RGB still raises the full error, so the error it adds is a placement error. The range readout (Eq.~\eqref{eq:readout}) and the range shift remove it (Table~\ref{tab:ablation}). In every seed, the model trained with the readout has a lower error and fewer harmful additions than the model trained without it (Supplementary, Sec.~C.2).

The readout helps in the unseen room only when the model is trained with it. Switching the readout off at test time raises the error with all sensors. A readout added after training lowers the error of a model trained without it only in distribution. For X-Fi and COMPASS, such a readout gives no consistent gain (Supplementary, Sec.~C.3). In the unseen room, RGB also harms the pose. Adding RGB raises the pose error of every model there, and degradation and image-scale training reduce this increase (Supplementary, Sec.~C.5). On the pose alone, the gap of \ours{} is 8\%, against 15\% for X-Fi and 21\% for COMPASS (Table~\ref{tab:ablation}, right).

\subsection{Ablations, margins and robustness}\label{sec:degradation_test}\label{sec:ablation}
Subset training alone does not prevent negative fusion. Without the readout, the range shift and the two augmentations, the model is less accurate with all five sensors than with depth, mmWave and LiDAR (Table~\ref{tab:ablation}, w/o all four). Degradation and image-scale augmentation fix the pose part. Together they cut $G$ in the unseen room from 73\% to 7\% and the E04 error by 41\% (Table~\ref{tab:ablation}, Fig.~\ref{fig:ablation_render}). Each of the two contributes. The two augmentations also lower the E04 error of X-Fi (Table~\ref{tab:robust}).

\paragraph{Subset view, core and losses.} The subset view keeps small subsets accurate. Without it, the mean error over the 31 subsets is 42\% higher in distribution (Table~\ref{tab:ablation}). The bottleneck core, $\mathcal L_{\rm KD}$ and $\mathcal L_{\rm MoFe}$ each lower $G$ in the unseen room.

\paragraph{Margins and robustness.} \ours{} has a lower error than every retrained baseline on both test sets. With all five sensors, its error is 27\% lower in distribution and 40\% lower in the unseen room. Its mean error over the 31 subsets is 25\% lower in distribution and 24\% lower in the unseen room (Table~\ref{tab:main}, Fig.~\ref{fig:mmfi}). The proposed parts produce most of this margin. Without them, the same network keeps a small lead in distribution and none in the unseen room (Table~\ref{tab:ablation}, w/o all four). Training the encoders of X-Fi end to end does not close the gap. \ours{} stays 21\% lower in distribution, and X-Fi then has more harmful additions on both test sets (Supplementary, Sec.~C.4). Under test-time degradation, the error of \ours{} grows the least, also against X-Fi trained with our degradations (Table~\ref{tab:robust}, Fig.~\ref{fig:degradation}).

\subsection{Generalization to four more datasets and tasks}\label{sec:he}\label{sec:reli}
We train the same design separately on four more datasets. It outperforms the retrained baselines on each of them (Table~\ref{tab:main}), and harmful additions are rare (Table~\ref{tab:ablation}, right).

\begin{figure}[t]
\includegraphics[width=\linewidth]{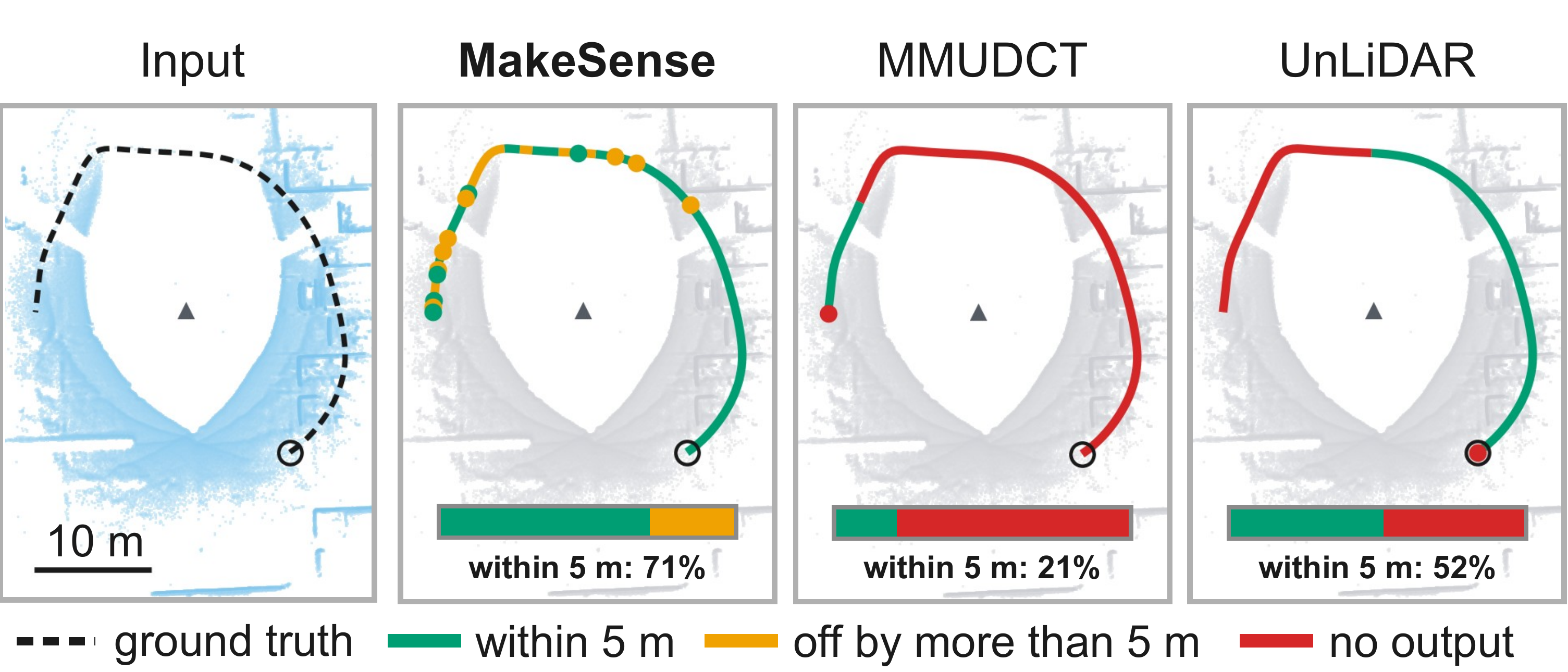}
\vspace{-0.5cm}
\caption{
\textbf{MMAUD example: MMUDCT is accurate but mostly silent; \ours{} reports at every step.} The track is colored in the 20\,s top view (\textcolor{figgreen}{green}: error $\leq 5$\,m; \textcolor{figamber}{amber}: $>5$\,m; \textcolor{figred}{red}: no output). Bars: shares over the steps of this clip.
}
\label{fig:mmaud}
\end{figure}

\begin{figure}[t]
\centering
\includegraphics[width=\linewidth]{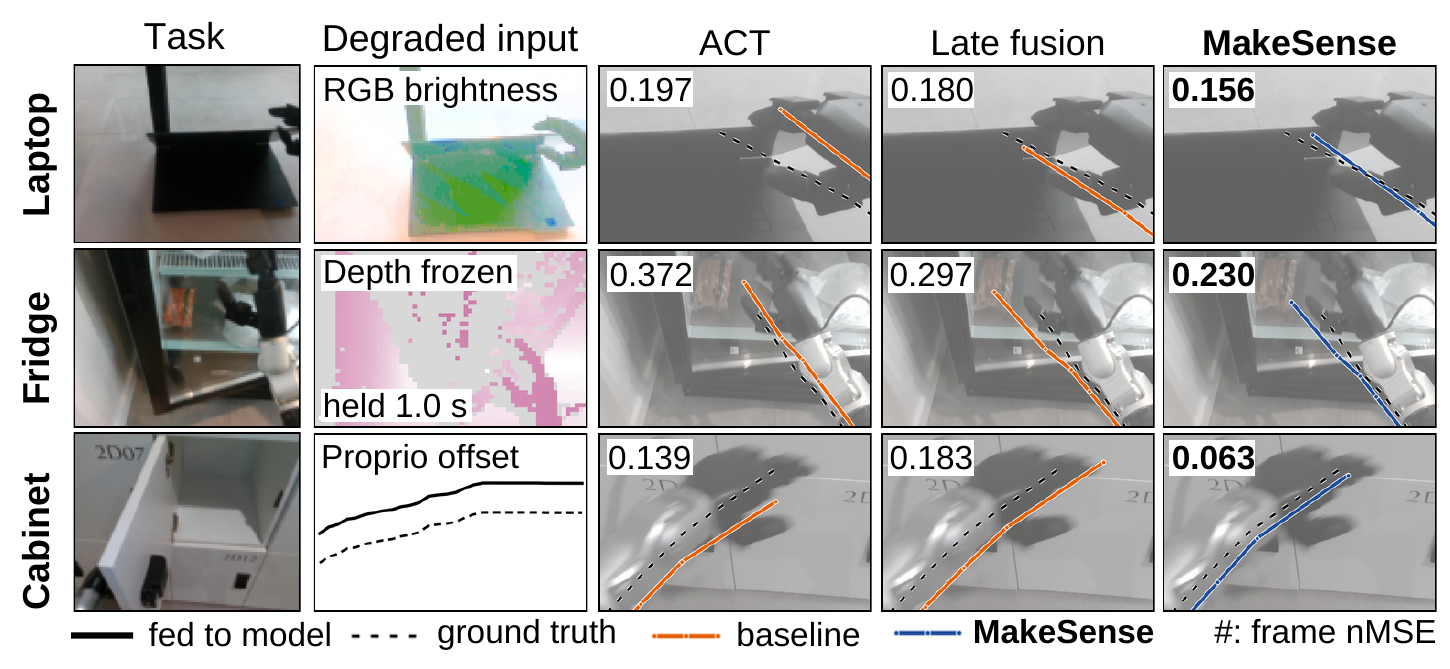}
\vspace{-0.7cm}
\caption{
\textbf{Humanoid-Everyday under single-sensor faults.}
Each task shows one fault at strength 0.5 (Table~\ref{tab:robust}).
Arm poses are overlaid on head-camera frames 0.3\,s after input.
The frames are examples on which \ours{} has the lowest nMSE.
}
\label{fig:he_faults}
\end{figure}

\begin{table}[t]
\centering \footnotesize
\setlength{\tabcolsep}{2.6pt}
\caption{\textbf{Forecasting the frozen fused state (HE, test).} MSE$\downarrow$ of the standardized state. Mean\,$\pm$\,SD over 3 seeds. Tint: gap to the row best. All horizons in Supplementary, Sec.~F.1.}
\vspace{-0.3cm}
\label{tab:wm}
\begin{tabular}{@{}lcccccc@{}}
\toprule
 & & & & \multicolumn{2}{c}{JEPA} & \\
\cmidrule(lr){5-6}
Ahead & Last State & Linear & DINO-WM & w/ cmd. & w/o cmd. & RSSM \\
\midrule
0.2\,s & \cellcolor{sevbad}0.21\pmsdt{.00} & \cellcolor{sevok}0.15\pmsdt{.01} & \cellcolor{sevbest}\textbf{0.14}\pmsdt{.00} & \cellcolor{sevbest}\textbf{0.14}\pmsdt{.01} & \cellcolor{sevmid}0.18\pmsdt{.02} & \cellcolor{sevmid}0.19\pmsdt{.01} \\
0.6\,s & \cellcolor{sevbad}0.90\pmsdt{.05} & \cellcolor{sevmid}0.37\pmsdt{.03} & \cellcolor{sevmid}0.35\pmsdt{.04} & \cellcolor{sevok}0.31\pmsdt{.01} & \cellcolor{sevbad}0.43\pmsdt{.01} & \cellcolor{sevbest}\textbf{0.28}\pmsdt{.02} \\
1.0\,s & \cellcolor{sevbad}1.44\pmsdt{.12} & \cellcolor{sevmid}0.45\pmsdt{.04} & \cellcolor{sevmid}0.45\pmsdt{.05} & \cellcolor{sevmid}0.40\pmsdt{.03} & \cellcolor{sevbad}0.55\pmsdt{.05} & \cellcolor{sevbest}\textbf{0.30}\pmsdt{.02} \\
\bottomrule
\end{tabular}
\vspace{-0.3cm}
\end{table}

\begin{figure}[t]
\includegraphics[width=\linewidth]{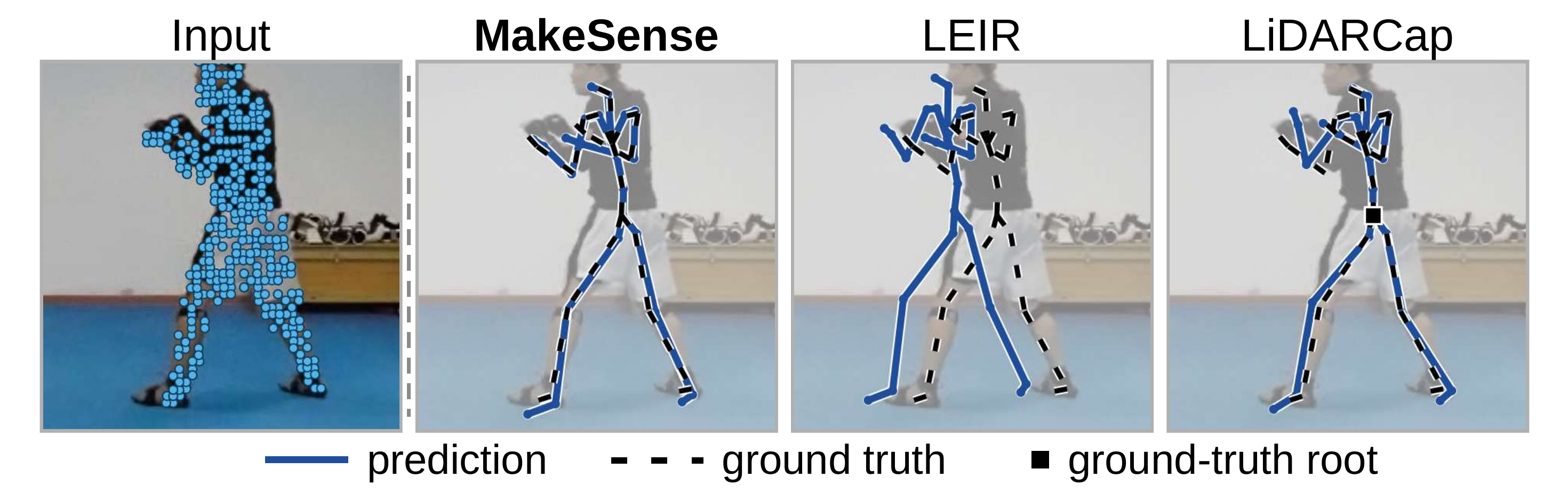}
\vspace{-0.7cm}
\caption{
\textbf{RELI11D example: \ours{} matches the ground truth best.}
LiDARCap predicts no translation, so its pose is drawn at the true root (square). Table~\ref{tab:main} gives the test-set means.
}
\label{fig:reli}
\vspace{-0.3cm}
\end{figure}

\begin{table*}[t]
\centering\footnotesize
\vspace{-0.5cm}
\setlength{\tabcolsep}{2pt}
\caption{\textbf{Negative fusion and ablations on MM-Fi.}
Mean\,$\pm$\,SD over 3 seeds ($^\dagger$2 seeds). MPJPE in mm. Avg.\,(31): mean over all subsets.
$G$: \% error change when RGB and WiFi CSI are added to depth, mmWave and LiDAR. Neg.: single-sensor additions (of 75) that raise the error. Scale: image-scale augmentation. W/o all four: no readout, range shift, deg.\ or scale. Concat.\ core: full attention over all tokens. Tint: $G$ and Neg.\ from \sevkey{sevbest}{no harm} to \sevkey{sevbad}{most}; errors \sevkey{sevmid}{$>$20\%} or \sevkey{sevbad}{$>$50\%} above the best. Right: Neg.\ at margins of 2\% and 5\%, the pose-only gap, and all five sensors against the best subset. Below: the 2\% count on the other datasets (out of the number after the slash).
}
\label{tab:ablation}
\vspace{-0.3cm}
\begin{minipage}[t]{0.60\textwidth}
\vspace{0pt}
\centering
\resizebox{\linewidth}{!}{%
\begin{tabular}{@{}l rrrr rrrr@{}}
\toprule
 & \multicolumn{4}{c}{ID test} & \multicolumn{4}{c}{OOD (E04)} \\
\cmidrule(lr){2-5}\cmidrule(lr){6-9}
\grp{MM-Fi model} & \mh{MPJPE\dn} & \mh{Avg.\,(31)\dn} & \mh{$G$\dn} & \mh{Neg.\dn} & \mh{MPJPE\dn} & \mh{Avg.\,(31)\dn} & \mh{$G$\dn} & \mh{Neg.\dn} \\
\midrule
X-Fi~\cite{chen2025xfi} & \cellcolor{sevmid}82.1\pmsdt{4.8} & \cellcolor{sevmid}97.3\pmsdt{2.9} & \cellcolor{sevok}+1.9\pmsdt{4.9} & \cellcolor{sevmid}16.0\pmsdt{2.6} & \cellcolor{sevbad}133.7\pmsdt{18.4} & \cellcolor{sevmid}155.3\pmsdt{8.8} & \cellcolor{sevmid}+12.8\pmsdt{17.4} & \cellcolor{sevbad}27.7\pmsdt{9.3} \\
COMPASS~\cite{compass} & \cellcolor{sevmid}80.5\pmsdt{10.2} & \cellcolor{sevmid}100.5\pmsdt{26.6} & \cellcolor{sevok}+3.2\pmsdt{7.5} & \cellcolor{sevmid}18.0\pmsdt{9.6} & \cellcolor{sevbad}149.2\pmsdt{9.4} & \cellcolor{sevbad}180.8\pmsdt{30.8} & \cellcolor{sevmid}+19.7\pmsdt{7.4} & \cellcolor{sevmid}24.0\pmsdt{7.8} \\
\midrule
\textbf{\ours} & 58.4\pmsdt{1.4} & 72.7\pmsdt{1.7} & \cellcolor{sevbest}$-$1.1\pmsdt{1.6} & \cellcolor{sevbest}3.3\pmsdt{4.9} & 80.2\pmsdt{0.7} & 118.1\pmsdt{3.7} & \cellcolor{sevmid}+7.3\pmsdt{3.8} & \cellcolor{sevmid}23.3\pmsdt{2.5} \\
\midrule
w/o readout & 60.3\pmsdt{1.8} & 74.0\pmsdt{2.2} & \cellcolor{sevok}+2.3\pmsdt{4.2} & \cellcolor{sevok}10.7\pmsdt{5.9} & 88.4\pmsdt{6.1} & 119.7\pmsdt{5.2} & \cellcolor{sevmid}+15.9\pmsdt{6.5} & \cellcolor{sevbad}27.7\pmsdt{3.1} \\
readout off at test & 66.4\pmsdt{5.0} & 77.4\pmsdt{3.8} & \cellcolor{sevmid}+12.5\pmsdt{7.5} & \cellcolor{sevmid}15.3\pmsdt{1.5} & 93.5\pmsdt{7.8} & 130.7\pmsdt{10.1} & \cellcolor{sevbad}+24.9\pmsdt{8.2} & \cellcolor{sevmid}24.3\pmsdt{2.5} \\
w/o range shift & 64.7\pmsdt{3.7} & 76.5\pmsdt{2.7} & \cellcolor{sevok}+4.9\pmsdt{6.2} & \cellcolor{sevok}11.3\pmsdt{3.8} & 87.3\pmsdt{7.0} & 123.9\pmsdt{4.9} & \cellcolor{sevmid}+12.8\pmsdt{10.8} & \cellcolor{sevbad}26.0\pmsdt{1.7} \\
w/o deg. & 59.0\pmsdt{2.2} & 72.9\pmsdt{1.3} & \cellcolor{sevbest}$-$0.8\pmsdt{4.2} & \cellcolor{sevok}7.0\pmsdt{5.6} & \cellcolor{sevmid}106.0\pmsdt{6.3} & 134.1\pmsdt{8.6} & \cellcolor{sevbad}+35.5\pmsdt{7.0} & \cellcolor{sevbad}27.7\pmsdt{2.1} \\
w/o scale & 61.4\pmsdt{0.6} & 72.9\pmsdt{0.3} & \cellcolor{sevmid}+5.0\pmsdt{1.3} & \cellcolor{sevmid}18.7\pmsdt{2.5} & \cellcolor{sevmid}98.1\pmsdt{7.5} & 123.1\pmsdt{7.4} & \cellcolor{sevbad}+30.8\pmsdt{7.3} & \cellcolor{sevbad}27.7\pmsdt{1.2} \\
w/o deg., scale & 63.3\pmsdt{2.3} & 74.0\pmsdt{1.7} & \cellcolor{sevmid}+6.6\pmsdt{4.4} & \cellcolor{sevmid}17.3\pmsdt{6.4} & \cellcolor{sevbad}135.9\pmsdt{14.3} & \cellcolor{sevmid}150.8\pmsdt{10.6} & \cellcolor{sevbad}+73.1\pmsdt{25.5} & \cellcolor{sevbad}30.3\pmsdt{2.9} \\
w/o all four & \cellcolor{sevmid}73.3\pmsdt{4.9} & 79.5\pmsdt{3.4} & \cellcolor{sevbad}+20.5\pmsdt{8.1} & \cellcolor{sevok}12.7\pmsdt{0.6} & \cellcolor{sevbad}152.1\pmsdt{17.6} & \cellcolor{sevmid}159.8\pmsdt{20.8} & \cellcolor{sevbad}+94.0\pmsdt{25.9} & \cellcolor{sevbad}29.3\pmsdt{3.1} \\
w/o subset view & 58.4\pmsdt{1.0} & \cellcolor{sevmid}103.3\pmsdt{0.7} & \cellcolor{sevbest}$-$31.0\pmsdt{5.0} & \cellcolor{sevbest}0.3\pmsdt{0.6} & 93.8\pmsdt{5.1} & \cellcolor{sevmid}151.4\pmsdt{11.3} & \cellcolor{sevbest}$-$14.4\pmsdt{11.9} & \cellcolor{sevok}10.7\pmsdt{3.2} \\
Concat.\ core$^\dagger$ & 58.7\pmsdt{0.9} & 73.0\pmsdt{1.2} & \cellcolor{sevbest}+0.0\pmsdt{0.8} & \cellcolor{sevok}9.0\pmsdt{0.0} & 85.6\pmsdt{6.0} & 123.4\pmsdt{1.8} & \cellcolor{sevmid}+14.2\pmsdt{6.8} & \cellcolor{sevbad}25.5\pmsdt{0.7} \\
w/o $\mathcal L_{\rm KD}$$^\dagger$ & 57.9\pmsdt{1.2} & 72.6\pmsdt{0.7} & \cellcolor{sevbest}$-$2.8\pmsdt{1.4} & \cellcolor{sevbest}2.5\pmsdt{2.1} & 82.5\pmsdt{1.2} & 119.6\pmsdt{1.1} & \cellcolor{sevmid}+10.1\pmsdt{1.9} & \cellcolor{sevbad}27.0\pmsdt{1.4} \\
w/o $\mathcal L_{\rm MoFe}$$^\dagger$ & 59.4\pmsdt{0.4} & 74.0\pmsdt{1.0} & \cellcolor{sevok}+0.9\pmsdt{1.2} & \cellcolor{sevok}10.0\pmsdt{1.4} & 84.9\pmsdt{8.4} & 121.8\pmsdt{4.5} & \cellcolor{sevmid}+13.7\pmsdt{10.6} & \cellcolor{sevbad}25.0\pmsdt{1.4} \\
\bottomrule
\end{tabular}}
\end{minipage}\hfill
\begin{minipage}[t]{0.37\textwidth}
\vspace{0pt}
\centering
\resizebox{\linewidth}{!}{%
\begin{tabular}{@{}l l rrr@{}}
\toprule
\grp{MM-Fi} & & \mh{\ours{}} & \mh{X-Fi} & \mh{COMPASS} \\
\midrule
Neg.\ $>$2\%\dn & ID & \cellcolor{sevbest}1.3\pmsdt{2.3} & 7.0\pmsdt{1.7} & 10.3\pmsdt{7.5} \\
 & E04 & \cellcolor{sevbest}13.0\pmsdt{1.7} & 24.3\pmsdt{8.7} & 22.7\pmsdt{7.1} \\
Neg.\ $>$5\%\dn & ID & \cellcolor{sevbest}1.3\pmsdt{2.3} & 2.3\pmsdt{2.5} & 2.0\pmsdt{1.7} \\
 & E04 & \cellcolor{sevbest}7.3\pmsdt{2.1} & 19.0\pmsdt{8.2} & 14.0\pmsdt{2.6} \\
 $G$, pose only\dn & ID & \cellcolor{sevbest}$-$2.8\pmsdt{1.0} & $-$0.5\pmsdt{1.4} & +1.3\pmsdt{4.4} \\
{\tiny(PA-MPJPE)} & E04 & \cellcolor{sevbest}+8.2\pmsdt{1.8} & +14.9\pmsdt{6.1} & +21.0\pmsdt{5.4} \\
vs.\ best subset\dn & ID & \cellcolor{sevbest}+0.2\pmsdt{0.8} & +3.6\pmsdt{3.2} & +6.1\pmsdt{3.6} \\
{\tiny(\% error)} & E04 & \cellcolor{sevbest}+7.4\pmsdt{3.7} & +18.6\pmsdt{15.1} & +19.7\pmsdt{7.4} \\
\midrule
\grp{Other datasets} & & \mh{\ours{}} & \multicolumn{2}{r}{\mh{Best baseline}} \\
\midrule
Neg.\ $>$2\%\dn & HE {\tiny/\,28} & \cellcolor{sevbest}2.0\pmsdt{0.0} & \multicolumn{2}{r}{2.3\pmsdt{0.6} {\tiny(Late fusion)}} \\
 & V2X-Real {\tiny/\,12} & \cellcolor{sevbest}0.0\pmsdt{0.0} & \multicolumn{2}{r}{\cellcolor{sevbest}0.0\pmsdt{0.0} {\tiny(all four baselines)}} \\
 & MMAUD {\tiny/\,28} & \cellcolor{sevbest}0.7\pmsdt{0.6} & \multicolumn{2}{r}{--} \\
 & RELI11D {\tiny/\,9} & \cellcolor{sevbest}0.0\pmsdt{0.0} & \multicolumn{2}{r}{--} \\
\bottomrule
\end{tabular}}
\end{minipage}
\vspace{-0.3cm}
\end{table*}

\begin{figure*}[t!]
\centering
\includegraphics[width=0.93\linewidth]{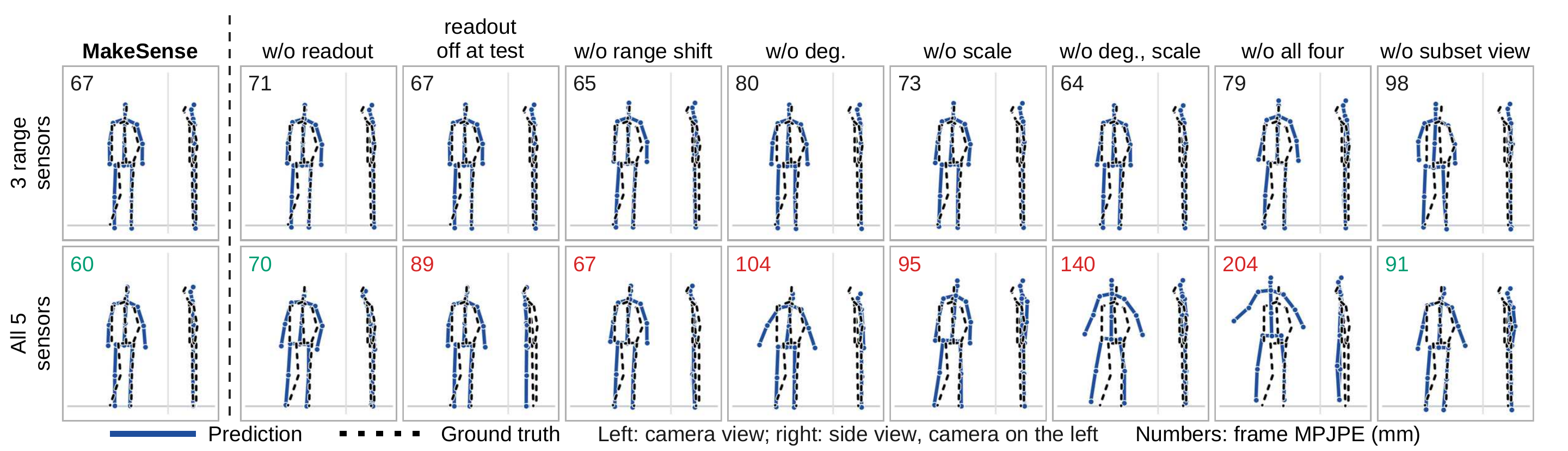}
\vspace{-0.2cm}
\caption{
\textbf{On this E04 test frame, adding RGB and WiFi helps \ours{} but hurts most of the ablations.} Without the two augmentations, the added sensors lift the body and spread its arms (\textcolor{figgreen}{green}: 5 sensors better; \textcolor{figred}{red}: worse). Table~\ref{tab:ablation} reports means.
}
\label{fig:ablation_render}
\vspace{-0.3cm}
\end{figure*}

\paragraph{UAV localization.} On MMAUD, \ours{} reports a position at every test step and has the lowest error over all steps (Table~\ref{tab:main}, Fig.~\ref{fig:mmaud}). MMUDCT reports at about half of the steps. We score its silent steps at the training-mean position, and \ours{} also leads when they take the last or the nearest output of MMUDCT (Supplementary, Sec.~E.1). Keeping a thinned or cluttered LiDAR gives a lower error than dropping it (Table~\ref{tab:robust}).

\paragraph{Cooperative detection.} On V2X-Real, \ours{} has the highest AP50 and AP70, also within the baselines' 70\,m range (Table~\ref{tab:main}). It keeps the highest AP50 under each fault and loses the smallest share of its AP (Table~\ref{tab:robust}, Fig.~\ref{fig:v2x}).

\paragraph{Humanoid command prediction.} On Humanoid-Everyday, \ours{} has the lowest next-command error among fusion models and learned policies. Its error is 8\% below the best fusion baseline with the same four inputs (Table~\ref{tab:main}) and 5\% below ACT, which reads RGB and joint state (Table~\ref{tab:robust}). \ours{} stays ahead of ACT for up to five steps (Supplementary, Sec.~E.3). Keeping a biased proprioceptive stream is better than dropping it (Table~\ref{tab:robust}, Fig.~\ref{fig:he_faults}). On the frozen fused state, an RSSM and a JEPA predictor forecast better than a linear model from 0.6\,s on, and commands help as well (Table~\ref{tab:wm}).

\paragraph{LiDAR motion capture.} On RELI11D, \ours{} trained from scratch is more accurate than LEIR and LiDARCap (Table~\ref{tab:main}, Fig.~\ref{fig:reli}). With 10\% of the labels, initialization from MM-Fi lowers the error by 5\% (Table~\ref{tab:main}).

\section{Limitations}
The design of \ours{} primarily focuses on robust multimodal perception under varying sensor availability and degraded sensing conditions. The current framework adopts per-dataset training and offline evaluation of the world model, while joint multi-dataset training and online evaluation remain unexplored. Joint training across datasets, online evaluation of the world model, and adaptive sensor configuration at inference time are left for future work.

\section{Conclusion}

We introduced \ours{}, a multimodal framework for embodied sensing with changing sensor availability and quality. The framework pairs one core design with task-specific heads for human pose estimation, aerial localization, humanoid command prediction, and cooperative 3D detection. Training on full and subset views of the same observations under sensor-specific degradations helps the model combine complementary cues when inputs are missing or unreliable. On MM-Fi, \ours{} mitigates negative fusion: it closes the gap between full-sensor and range-only fusion in distribution and narrows it in an unseen room. Trained separately on each dataset, the same architecture improves UAV localization on MMAUD and cooperative 3D detection on V2X-Real. The design core also supports few-shot pose transfer from MM-Fi to RELI11D. On Humanoid-Everyday, a world model uses the frozen core’s representations to forecast future fused states. In the unseen MM-Fi room, dropping an occluded LiDAR stream can still yield lower pose error than retaining it. Reliable fusion requires learning whether a prediction would be more accurate with or without a sensor, so the model can decide at inference time when a degraded sensor is still worth using.

\section*{LLM Usage Statement}
We used Claude Opus 5.5 as a coding assistant to implement training, evaluation, and visualization code, and as a writing assistant to correct grammar and wording and to find and fix inaccurate or inconsistent statements in the manuscript. All methods, experiments, analyses, and claims were designed, conducted, and verified by the authors, and we take full responsibility for the content.

\label{page:references_start}
{\small
\bibliographystyle{plain}
\bibliography{references}
}
\label{page:references_end}

\setcounter{figure}{0}\renewcommand{\thefigure}{S\arabic{figure}}
\twocolumn[{\csname nolinenumbers\endcsname\begin{center}{\large Supplementary Material}\end{center}\vspace{2pt}

\begin{teaserfigure}
\includegraphics{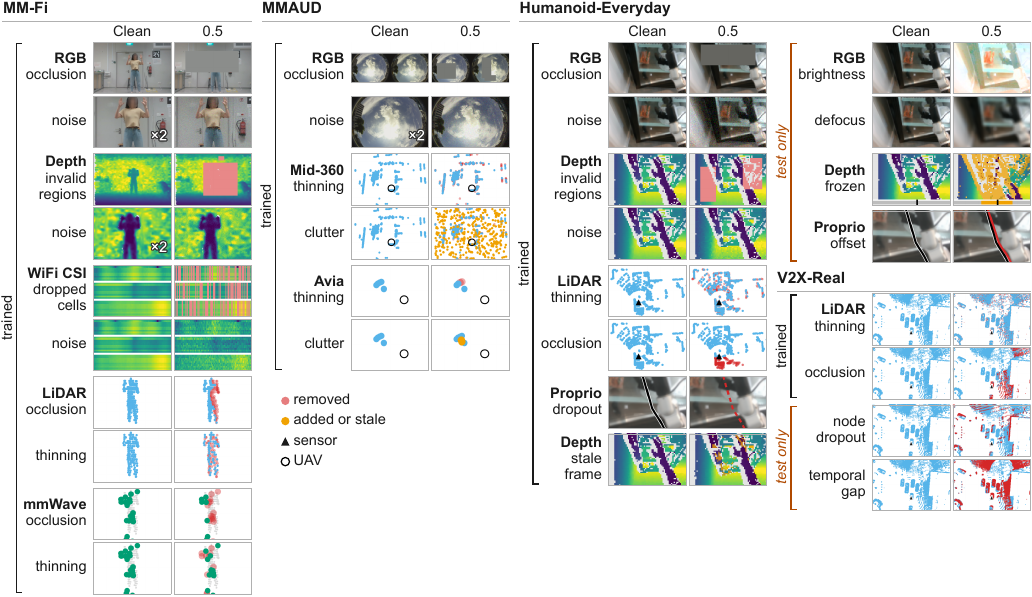}
\caption{\textbf{Training includes every degradation family shown except those marked ``test only.''} Model inputs for one sample per dataset, clean and at strength 0.5 (the test strength of Table~3 of the main paper); red: removed, amber: added or stale, $\times$2: twice the scale.}
\label{fig:supp_deg_gallery}
\end{teaserfigure}\vspace{8pt}
\noindent This supplement provides full details for each pointer in the main paper. It also presents two additional comparisons against X-Fi: an end-to-end trained variant of X-Fi (Sec.~\ref{supp:xfi_e2e}), and an evaluation of \ours{} on X-Fi's original split, where \ours{} achieves lower error than the published results of both X-Fi and COMPASS (Sec.~\ref{supp:xfi_split}). Sec.~\ref{supp:more} lists every input subset and shows median wins and losses on each dataset.\par\vspace{10pt}}]
\csname linenumbers\endcsname
\appendix
\setcounter{table}{0}\renewcommand{\thetable}{S\arabic{table}}
\section{\texorpdfstring{Method details \textnormal{(main paper, Sec.~3)}}{Method details (main paper, Sec. 3)}}
\subsection{Encoders, adapters and heads per dataset}\label{supp:arch}
\noindent All datasets share the same fusion core, while the sensor encoders, adapters, and task heads are tailored to each benchmark (Table~\ref{tab:arch}\label{fref:tab:arch}).

\begin{table}[t]
\centering\scriptsize
\setlength{\tabcolsep}{2.5pt}
\caption{\textbf{Dataset-specific encoders, heads, trainable parameters, and FLOPs.} Parameters in millions (trainable weights only, all encoders initialized from random weights); FLOPs of one forward pass on one test frame with all sensors present.}
\label{tab:arch}
\begin{tabular}{@{}>{\raggedright\arraybackslash}p{0.33\columnwidth}>{\raggedright\arraybackslash}p{0.49\columnwidth}r@{}}
\toprule
Sensor and input & Encoder & \mh{Param.~(M)} \\
\midrule
\multicolumn{3}{@{}p{\columnwidth}@{}}{\textbf{MM-Fi} (58.63 in total; 59.7\,GFLOPs). Head: 17 joint queries in a two-layer decoder over the bottleneck and sensor tokens, predicting a 3D position per joint (2.11).} \\
\addlinespace[1pt]
RGB, $480{\times}640$ & ResNet-18 of X-Fi~\cite{chen2025xfi}, group-normalized stem & 11.23 \\
Depth, $480{\times}640$, metres & same network as RGB, single input channel & 11.22 \\
mmWave points & Point Transformer of X-Fi (mmWave) & 1.28 \\
LiDAR, 1{,}024 points & Point Transformer of X-Fi (LiDAR), first stage & 3.26 \\
WiFi CSI, $3{\times}114{\times}10$ & ResNet-34 trunk of X-Fi's CSI network & 21.28 \\
\midrule
\multicolumn{3}{@{}p{\columnwidth}@{}}{\textbf{MMAUD} (37.64 in total; 54.0--84.6\,GFLOPs). Head: one query in a two-layer decoder for UAV position and type (four classes), plus a second query for the readout of Sec.~\ref{supp:readout} (4.25).} \\
\addlinespace[1pt]
RGB, two views, $240{\times}640$ & same ResNet-18 as MM-Fi & 11.23 \\
LiDAR (Mid-360), $\le$1{,}024 points & Point Transformer (LiDAR), second stage & 4.99 \\
LiDAR (Avia), $\le$1{,}024 points & same architecture & 4.99 \\
Radar points & same architecture & 4.99 \\
\midrule
\multicolumn{3}{@{}p{\columnwidth}@{}}{\textbf{Humanoid-Everyday} (12.04 in total; 30.2\,GFLOPs). Head: 15 command queries with one cross-attention layer over the bottleneck tokens and a feed-forward block, outputting 28 values per command (0.54).} \\
\addlinespace[1pt]
RGB, $96{\times}128$ & four stride-2 convolutions with group normalization & 0.39 \\
Depth, $48{\times}64$, metres and validity & three identical stride-2 convolutions & 0.09 \\
LiDAR, $\le$1{,}024 points & Point Transformer (LiDAR), first stage & 3.26 \\
Proprioception, 69 values, 9 frames & per-block MLP over the state, followed by temporal attention across frames & 0.61 \\
\midrule
\multicolumn{3}{@{}p{\columnwidth}@{}}{\textbf{V2X-Real} (16.29 in total; about 1.7\,TFLOPs). Head: per window, 64 cell queries in a two-layer decoder over the bottleneck and node tokens predict a residual on the ego map (2.17), read by an anchor-based detection head (0.02).} \\
\addlinespace[1pt]
LiDAR of each of the four nodes, in the ego frame & single PointPillars encoder shared across nodes; BEV map of $100{\times}256$ cells at 0.8\,m resolution & 8.05 \\
\midrule
\multicolumn{3}{@{}p{\columnwidth}@{}}{\textbf{RELI11D} (17.43 in total; 21.4\,GFLOPs). Head: the 17-query decoder of MM-Fi (2.11) followed by a mapping from 17 to the 24 dataset joints (0.15).} \\
\addlinespace[1pt]
RGB feature, 2{,}048 values & two-layer MLP on the released feature & 2.63 \\
Event feature, 2{,}048 values & two-layer MLP on the released feature & 2.63 \\
LiDAR, $\le$512 points, centred & Point Transformer (LiDAR), first stage & 3.26 \\
\bottomrule
\end{tabular}
\end{table}

Before encoding, the MMAUD adapter of \ours{} filters out Mid-360 returns with $z\le1$\,m in the sensor frame and radar returns beyond 300\,m. MMUDCT and UnLiDAR retain their original preprocessing of the released point clouds.

On MMAUD, any empty Avia or radar frame is substituted with the most recent non-empty frame within the preceding ten annotation steps.

\subsection{Range readout per dataset}\label{supp:readout}
\noindent On MM-Fi, the range readout is a parameter-free second pass through the fusion core; on MMAUD and \mbox{V2X-Real}, it is a lightweight learned module operating solely on the point clouds. We ablate the readout on MM-Fi in Secs.~\ref{supp:readout_seeds}~and~\ref{supp:posthoc}.

\subsection{Degradation families and parameters}\label{supp:degradation}
\noindent Point thinning, angular occlusion, and clutter belong to the training degradation families, whereas node dropout, temporal gaps, and the four Humanoid-Everyday faults are evaluated exclusively at test time (Fig.~\ref{fig:supp_deg_gallery}\label{fref:fig:supp_deg_gallery}, Table~\ref{tab:deg}\label{fref:tab:deg}).

Each family is parameterized by a severity strength $d\in[0,1]$, where $d=0$ leaves the input unchanged. The test faults in Table~3 of the main paper use $d=0.5$.

\begin{table*}[!t]
\centering\scriptsize
\setlength{\tabcolsep}{3pt}
\caption{\textbf{Degradation families and test-time faults (main paper, Table~3).} $d$: degradation strength; $n$: number of points in the cloud; HE: Humanoid-Everyday; the right columns list each test fault's parameters at strengths 0.3, 0.5 (used in the main paper) and 0.7.}
\label{tab:deg}
\begin{tabular}{@{}ll>{\raggedright\arraybackslash}p{0.27\textwidth}>{\raggedright\arraybackslash}p{0.15\textwidth}lccc@{}}
\toprule
 & & & & \multicolumn{4}{c}{Test fault of Table~3} \\
\cmidrule(l){5-8}
Sensor & Family & Effect at strength $d$ & In training on & Datasets & 0.3 & 0.5 & 0.7 \\
\midrule
RGB image & Occlusion rectangles & 1--4 rectangles (aspect ratio in $[1/3,3]$, total area $0.5\,d$) filled with the mean colour & MM-Fi, MMAUD, HE & -- &  &  &  \\
 & Noise & Gaussian noise with SD $0.1\,d$ of the value range, clipped & MM-Fi, MMAUD, HE & -- &  &  &  \\
\addlinespace[1pt]
Depth image & Invalid regions & same rectangles as RGB, marked invalid & MM-Fi, HE & -- &  &  &  \\
 & Multiplicative noise & valid pixels multiplied by $1+e$, $e\sim\mathcal N(0,(0.05\,d)^2)$ & MM-Fi, HE & -- &  &  &  \\
\addlinespace[1pt]
WiFi CSI & Dropped cells & fraction $d$ of the 342 antenna-subcarrier cells dropped and refilled with the mean of the remaining cells in the packet & MM-Fi & -- &  &  &  \\
 & Noise & Gaussian noise with SD $d$ times the frame SD & MM-Fi & -- &  &  &  \\
\addlinespace[1pt]
Point cloud & Point thinning & $d\,n$ points removed at random & MM-Fi (LiDAR, mmWave), MMAUD (two LiDARs), HE, V2X-Real & MM-Fi, MMAUD, V2X-Real & 30\% & 50\% & 70\% \\
 & Angular occlusion & points in a random $360^\circ d$ azimuth sector removed; centred on the sensor (MM-Fi: on the median $x$-$y$ of the cloud excluding its lowest 0.2\,m) & MM-Fi (LiDAR, mmWave), HE, V2X-Real & MM-Fi, V2X-Real & $108^\circ$ & $180^\circ$ & $252^\circ$ \\
 & Clutter & $d\,n$ points added uniformly within the bounding box & MMAUD (two LiDARs) & MMAUD & 30\% & 50\% & 70\% \\
\addlinespace[1pt]
Stream & Dropout & sensor dropped with probability $d$ & HE (proprioception) & -- &  &  &  \\
 & Stale frame & replaced by a frame from 1--3 steps earlier (30\,Hz) with probability 0.1 per sensor, independent of $d$ & HE (all four sensors) & -- &  &  &  \\
 & Temporal gap & blocks of 10 consecutive frames dropped, each with probability $d$ & none & V2X-Real & 0.3 & 0.5 & 0.7 \\
\addlinespace[1pt]
Node & Node dropout & each non-ego node dropped for an entire sequence with probability $d$ & none & V2X-Real & 0.3 & 0.5 & 0.7 \\
\addlinespace[2pt]
\multicolumn{8}{@{}l}{\textit{Test faults of Humanoid-Everyday, never drawn in training}} \\
RGB image & Brightness & HSV value channel (in $[0,1]$) raised by & none & HE & 0.5 & 0.6 & 0.7 \\
 & Defocus & disk blur with radius specified at 224\,px height, scaled to image size & none & HE & 4\,px & 8\,px & 10\,px \\
\addlinespace[1pt]
Depth image & Frozen frame & a single depth frame repeated over a fraction of the episode & none & HE & 20\% & 40\% & 60\% \\
\addlinespace[1pt]
Stream & Proprio.\ offset & constant offset per arm joint (14 joints), sampled uniformly within $\pm$ & none & HE & 0.01\,rad & 0.05\,rad & 0.10\,rad \\
\bottomrule
\end{tabular}
\end{table*}

\begin{table}[!t]
\centering\scriptsize
\setlength{\tabcolsep}{2.5pt}
\caption{\textbf{World-model predictors evaluated in Table~4 of the main paper.} Parameters in millions (with\,/\,without commands); all learned predictors share one training recipe (AdamW, peak learning rate $3\times10^{-4}$, 5{,}000 updates of 32 windows).}
\label{tab:wm_predictors}
\begin{tabularx}{\columnwidth}{@{}l>{\raggedright\arraybackslash}Xc>{\raggedright\arraybackslash}p{0.24\columnwidth}@{}}
\toprule
Predictor & Reads at each step & Param. & Training loss \\
\midrule
Last State & repeats the context state & 0 & none \\
\addlinespace[1pt]
Linear & previous state (shared map across the 8 tokens) and mean commands & 0.07 & ridge, one step ahead \\
\addlinespace[1pt]
DINO-WM~\cite{zhou2025dinowm} & previous state with 10-D commands appended to each token (6-block ViT) & 4.92\,/\,4.74 & squared error, one step from true states \\
\addlinespace[1pt]
JEPA~\cite{assran2025vjepa2} & all previous states (block-causal); mean commands modulate LayerNorm (4 blocks, width 256) & 5.07\,/\,5.00 & same as DINO-WM, plus a 2-step rollout \\
\addlinespace[1pt]
RSSM~\cite{hafner2025dreamerv3} & 2{,}048-unit recurrent state updated with mean commands & 8.42\,/\,7.75 & reconstruction and two KL terms \\
\bottomrule
\end{tabularx}
\end{table}

\textit{Schedule.} During training, the degradation strength cap remains at 0 until a warm-up milestone, then ramps linearly to 0.8 at a second milestone and stays fixed thereafter. These start/end milestones are updates 1{,}500/3{,}000 on MM-Fi, 500/1{,}000 on MMAUD and HE, and 1{,}570/3{,}140 on V2X-Real. For each sample and sensor, the strength is drawn as $d=u\cdot{\rm cap}$ with $u\sim\mathcal U(0,1)$. Each sample remains entirely clean with probability 0.5; otherwise, each sensor independently stays clean with probability 0.5 or is corrupted by one uniformly sampled family of its type. RELI11D is trained without degradation.

\begin{figure}[t]
\centering
\includegraphics{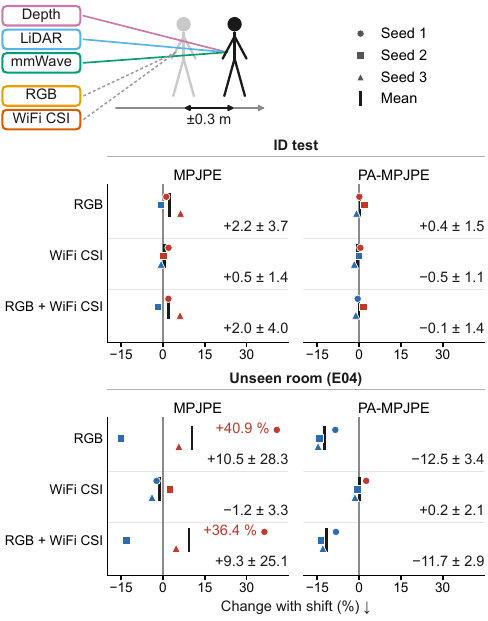}
\caption{\textbf{Range shifting lowers unseen-room RGB PA-MPJPE in every seed without consistently degrading subsets without a range sensor.} Top: the shift displaces the target only within depth, LiDAR and mmWave inputs; bottom: per-seed error change on MM-Fi (red: higher error) with mean\,$\pm$\,SD. In the unseen room, both RGB subsets improve in PA-MPJPE in every seed, while a single seed drives their mean MPJPE rise (RGB: +40.9\,\%; RGB\,+\,WiFi CSI: +36.4\,\%).}
\label{fig:supp_range_shift}
\end{figure}

\subsection{Range shift and objective}\label{supp:objective}
\noindent Under the per-seed consistency criterion of Sec.~\ref{supp:perseed}, range shifting causes no consistent degradation on subsets without a range sensor: none worsens across all three seeds in either MPJPE or PA-MPJPE (Fig.~\ref{fig:supp_range_shift}\label{fref:fig:supp_range_shift}).

\looseness=-1 In the training objective, the auxiliary heads carry weight 1 and the two bottleneck heads carry weight $\tfrac12$. The weight of $\mathcal L_{\rm KD}$ ramps linearly to $\lambda$ over the first 3{,}000 updates, while $\mathcal L_{\rm MoFe}$ begins at update 3{,}000 on batch means. $\mathcal L_{\rm var}$ is active from the first update, applying a hinge at 1 to the batch standard deviation of each of the $8\times256$ view-A bottleneck features prior to their final LayerNorm, and $\rho$ uses $\epsilon=10^{-6}$.

\subsection{World-model predictors}\label{supp:wm_predictors}
\noindent All predictors forecast the frozen fused state from the same context state and logged commands. They share a single untuned training recipe and are evaluated offline via open-loop rollouts starting from clean four-sensor frames (Table~\ref{tab:wm_predictors}).

\section{\texorpdfstring{Protocol details \textnormal{(main paper, Sec.~4)}}{Protocol details (main paper, Sec. 4)}}
\subsection{Splits, checkpoints and measures}\label{supp:splits}
\noindent Every test set is strictly separated from training along its split unit, and no test data are used for checkpoint selection~(Table~\ref{tab:splits}).

No E04 data are used during training or validation. When an mmWave point cloud is empty, every model defaults to the training-mean pose. On V2X-Real, a fixed rule in the QuantV2X loader excludes five test scenes from one recording day (919 frames) uniformly for all models. On RELI11D, training and validation splits are separated by 64 frames. MPJPE is measured in the depth-camera frame and thus includes global skeleton placement; Humanoid-Everyday nMSE normalizes each command dimension's squared error by its training variance, with MAE reported in radians; AP50 and AP70 denote QuantV2X average precision at IoU 0.5 and 0.7.

\begin{table}[t]
\centering\scriptsize\setlength{\tabcolsep}{2.5pt}
\caption{\textbf{Dataset splits, test sets, sensor configurations, and checkpoint selection for \ours{}.} Sample counts match Table~1 of the main paper, with $K=2^{|\mathcal M|}-1$ non-empty input subsets ($K=8$ on V2X-Real, where the ego node is always present).}\label{tab:splits}
\begin{tabular}{@{}>{\raggedright\arraybackslash}p{0.17\columnwidth}>{\raggedright\arraybackslash}p{0.80\columnwidth}@{}}
\toprule
Dataset & Split unit, test set, sensors and checkpoint \\
\midrule
MM-Fi~\cite{yang2023mmfi} & \textit{Split:} Subject; 24\,/\,3\,/\,3 subjects across three environments. E04: the 10 subjects of a fourth environment. \textit{Test:} ID: 24,057 frames. E04: 80,190 frames. \textit{Sensors ($K$):} RGB, depth, mmWave, LiDAR, WiFi CSI (31). \textit{Checkpoint:} Last update. \\
\addlinespace[2pt]
MMAUD~\cite{yuan2024mmaud} & \textit{Split:} Sequence, official split; 102\,/\,16\,/\,59 sequences. \textit{Test:} 5,900 steps, 100 per sequence. \textit{Sensors ($K$):} RGB, two LiDARs (Mid-360, Avia), radar (15). \textit{Checkpoint:} Lowest validation position error with all sensors, evaluated on every fourth annotated step; one snapshot per 1,000 updates. \\
\addlinespace[2pt]
HE~\cite{zhao2025humanoideveryday} & \textit{Split:} Episode, in recording order; 29\,/\,4\,/\,5 episodes for each of three tasks. \textit{Test:} 2,544 frames across 15 episodes. \textit{Sensors ($K$):} RGB, depth, LiDAR, proprioception (15). \textit{Checkpoint:} Lowest validation nMSE with all sensors; snapshots at 2,000, 3,000, and 4,000 updates. \\
\addlinespace[2pt]
V2X-Real~\cite{xiang2024v2xreal} & \textit{Split:} Scene, QuantV2X split~\cite{zhao2026quantv2x}; 5,769\,/\,717\,/\,1,253 frames. \textit{Test:} 1,253 frames across 9 scenes; 75,839 boxes. \textit{Sensors ($K$):} LiDAR of four nodes (8). \textit{Checkpoint:} Highest validation AP50 with all nodes; one snapshot per epoch. \\
\addlinespace[2pt]
RELI11D~\cite{Yan_2024_CVPR} & \textit{Split:} Time, within each recording; 1,131\,/\,165\,/\,225 windows of 16 frames. \textit{Test:} 3,600 frames. \textit{Sensors ($K$):} RGB features, event features, LiDAR (7). \textit{Checkpoint:} Last update; total updates chosen by three-sensor validation PA-MPJPE. \\
\bottomrule
\end{tabular}
\end{table}

\paragraph{How the settings were chosen.} Core capacity, loss weights (with $\lambda$ adopted from SimMLM~\cite{li2025simmlm}), subset-sampling and clean-sample probabilities, the degradation strength cap, and the MM-Fi training length were fixed in advance without tuning. On the MM-Fi validation subjects, adding RGB increased pose error---primarily in distance to the camera---so we introduced per-sensor degradations, image-scale augmentation, and range shifting; their parameters were specified before the first training run and kept unchanged thereafter, and their benefit was verified against the recipe without them on the validation subjects. Checkpoints are selected by best validation score (Table~\ref{tab:splits}), except on MM-Fi where the last update is kept (as it lies close to the validation optimum) and on RELI11D where the total update count is chosen on validation.

\subsection{Baselines and training budgets}\label{supp:baselines}
\noindent Every baseline is evaluated on the exact same test frames as \ours{}, and every trainable baseline is retrained on our splits (Table~\ref{tab:baselines}\label{fref:tab:baselines}).

\begin{table}[!t]
\centering\scriptsize
\setlength{\tabcolsep}{2pt}
\caption{\textbf{Baselines and training budgets.} Param.: trainable parameters in millions (``of'': total parameters when pre-trained components stay frozen or unused; $^{\ast}$: total parameter count; --: training-free for UnLiDAR or 15-step variant); grey: codebase modifications and input modalities; training: updates or epochs $\times$ batch size; $n$: number of seeds.}
\label{tab:baselines}
\begin{tabular}{@{}>{\raggedright\arraybackslash}p{0.53\columnwidth}>{\raggedright\arraybackslash}p{0.15\columnwidth}>{\raggedright\arraybackslash}p{0.22\columnwidth}c@{}}
\toprule
Model & Param.\,(M) & Training & $n$ \\
\midrule
\grp{MM-Fi} & & & \\
X-Fi~\cite{chen2025xfi} {\color{sdgray}Released code, ported; pre-trained encoders, as in the paper; 5 sensors} & 19.30 of 67.57 & 100k updates & 3 \\
\quad end to end {\color{sdgray}All encoders trained (Sec.~\ref{supp:xfi_e2e}); 5 sensors} & 67.57 & 100k updates & 3 \\
\quad + deg., scale {\color{sdgray}Our degradations and image scale; 5 sensors} & as X-Fi & 100k updates & 3 \\
COMPASS~\cite{compass} {\color{sdgray}Released code; pre-trained encoders, as in the paper; 5 sensors} & 2.84 of 51.11 & 60k updates & 3 \\
Late fusion {\color{sdgray}Mean of five single-sensor models; 5 sensors} & 48.41 & 12 epochs per model & 3 \\
\ours{} {\color{sdgray}5 sensors} & 58.63 & 30k $\times$ 32 & 3 \\
\midrule
\grp{MMAUD} & & & \\
MMUDCT~\cite{mmudct} {\color{sdgray}Released code; tracker LSTM retrained; two LiDARs} & 0.02 & 20 epochs (LSTM) & 3 \\
UnLiDAR~\cite{unlidar} {\color{sdgray}Released code, defaults, no training; LiDAR} & -- & -- & 1 \\
\ours{} {\color{sdgray}RGB, two LiDARs, radar} & 37.64 of 63.58 & 10k $\times$ 32 & 3 \\
\midrule
\grp{V2X-Real} & & & \\
Late fusion {\color{sdgray}QuantV2X code; detections merged; nodes within 70\,m} & 13.14$^{\ast}$ & 20 epochs $\times$ 4 & 3 \\
AttFuse~\cite{attfuse} {\color{sdgray}QuantV2X code; nodes within 70\,m} & 16.13$^{\ast}$ & 20 epochs $\times$ 2 & 3 \\
HEAL~\cite{heal} {\color{sdgray}QuantV2X code, three stages; nodes within 70\,m} & 6.38$^{\ast}$ & three stages & 3 \\
Pyramid Fusion~\cite{heal} {\color{sdgray}QuantV2X code; HEAL's fusion, one stage; nodes within 70\,m} & 5.71$^{\ast}$ & 18 epochs $\times$ 4 & 3 \\
\ours{} {\color{sdgray}every present node} & 16.29 & 28,840 $\times$ 4 (20 epochs) & 3 \\
\midrule
\grp{RELI11D} & & & \\
LEIR~\cite{Yan_2024_CVPR} {\color{sdgray}Released code and settings, all inputs; RGB and event features, LiDAR} & 566.84 & 200 epochs $\times$ 6 & 3 \\
LiDARCap~\cite{li2022lidarcap} {\color{sdgray}Released code and settings; LiDAR} & 34.93 & 200 epochs $\times$ 8 & 3 \\
\ours{} {\color{sdgray}RGB and event features, LiDAR} & 17.43 of 17.62 & 18k $\times$ 32 & 3 \\
\midrule
\grp{Humanoid-Everyday} & & & \\
ACT~\cite{act} {\color{sdgray}Released code; offline command target; RGB, joint state} & 51.66 & 22k $\times$ 8 (2,000 epochs) & 3 \\
Diffusion Policy~\cite{chi2023diffusionpolicy} {\color{sdgray}Released code; offline command target; RGB, joint state} & 77.29 & 246k $\times$ 64 (1,000 epochs) & 3 \\
Late fusion (one-step) {\color{sdgray}Our code; learned mean of per-sensor predictions; 4 sensors} & 1.20 & 10k $\times$ 32 & 3 \\
Late fusion (15-step) {\color{sdgray}Separate model, 15-command head (Sec.~\ref{supp:he_horizon}); 4 sensors} & -- & -- & 3 \\
MBT~\cite{nagrani2021mbt} {\color{sdgray}Released architecture and recipe; 4 sensors} & 342.12 & 12.3k $\times$ 64 (50 epochs) & 3 \\
Perceiver IO~\cite{jaegle2022perceiverio} {\color{sdgray}Released architecture and recipe; 4 sensors} & 47.87 & 1,650 $\times$ 1,024 (110 epochs) & 3 \\
\ours{} {\color{sdgray}4 sensors} & 12.04 of 14.49 & 2k--4k $\times$ 32 & 3 \\
\bottomrule
\end{tabular}
\end{table}

Published baselines follow their original protocols, with any modifications noted in grey in Table~\ref{tab:baselines}. As in their respective papers, X-Fi and COMPASS use single-sensor encoders pre-trained on our training split (shared across all three seeds), while ACT and MBT initialize their image backbones from ImageNet. In our port of X-Fi, the RGB encoder processes scaled pixels through a GroupNorm stem and the depth encoder takes metric depth. We also retrain X-Fi with our sensor degradations and image-scale augmentation while keeping its pre-trained encoders and training schedule (+\,deg., scale; Table~3 of the main paper). On Humanoid-Everyday, baselines are trained on complete inputs and evaluated on sensor subsets without retraining. ``No fusion'' in Fig.~6 of the main paper refers to the late-fusion detector evaluated on the ego node alone. Table~2 of the main paper lists the first of the three MM-Fi late-fusion runs; the three give 87.0\pmsdt{1.4}\,mm in distribution and 178.4\pmsdt{9.5}\,mm in the unseen room. All runs use NVIDIA GeForce RTX~5090 GPUs.

Table~\ref{tab:recipe} gives the training recipe of \ours{}, and Table~\ref{tab:he_zeroshot} the missing-input results of ACT and~Diffusion~Policy.

\begin{table}[t]
\centering\scriptsize\renewcommand{\arraystretch}{1.0}\setlength{\tabcolsep}{3pt}
\caption{\textbf{Training recipe of \ours{}.} Every run uses AdamW ($\beta_1{=}0.9$, $\beta_2{=}0.95$), weight decay 0.05, gradient clipping at 1.0, fp32 and peak learning rates of $10^{-3}$ for the encoders and $3\times10^{-4}$ for the fusion parameters, with a linear warm-up, a cosine to 1\,\% of the peak and a restart from one third of the peak; hours: wall clock of one run on one GPU, range over three seeds (Humanoid-Everyday keeps the snapshot chosen on validation, Table~\ref{tab:splits}).}\label{tab:recipe}
\begin{tabular}{@{}lrrrr@{}}
\toprule
Dataset & Batch\,$\times$\,updates & Warm-up & Restart & Hours \\
\midrule
MM-Fi & 32\,$\times$\,30{,}000 & 1{,}500 & 26{,}000 & 11.3--11.5 \\
MMAUD & 32\,$\times$\,10{,}000 & 500 & 8{,}000 & 3.4--3.7 \\
Humanoid-Everyday & 32\,$\times$\,5{,}000 & 500 & -- & 0.85--0.86 \\
V2X-Real & 4\,$\times$\,28{,}840 & 1{,}442 & 23{,}072 & 23.3--23.4 \\
RELI11D & 32\,$\times$\,18{,}000 & 900 & -- & 6.1--6.5 \\
\bottomrule
\end{tabular}
\end{table}

\begin{table}[t]
\centering\scriptsize\renewcommand{\arraystretch}{1.0}
\setlength{\tabcolsep}{3.5pt}
\caption{\textbf{ACT and Diffusion Policy are evaluated zero-shot on missing inputs.} Humanoid-Everyday test nMSE\dn, mean\,$\pm$\,SD over three seeds, by which of RGB and the joint state remain (both models are trained with all inputs present and read no depth or LiDAR); \ours{}: mean over the~row's~subsets.}
\label{tab:he_zeroshot}
\begin{tabular}{@{}l c ccc@{}}
\toprule
\mh{RGB / joint state} & \mh{Subsets} & \mh{ACT} & \mh{\shortstack{Diffusion\\Policy}} & \mh{\ours{}} \\
\midrule
both & 4 & 0.175\pmsdt{.003} & 0.209\pmsdt{.009} & 0.168\pmsdt{.006} \\
RGB only & 4 & 0.992\pmsdt{.037} & 1.130\pmsdt{.035} & 0.660\pmsdt{.010} \\
joint state only & 4 & 0.553\pmsdt{.046} & 0.298\pmsdt{.010} & 0.171\pmsdt{.008} \\
neither & 3 & 1.090\pmsdt{.000} & 1.090\pmsdt{.000} & 0.621\pmsdt{.017} \\
\addlinespace[2pt]
Avg.\,(15) & 15 & 0.677\pmsdt{.002} & 0.655\pmsdt{.014} & 0.390\pmsdt{.002} \\
\bottomrule
\end{tabular}
\end{table}

\begin{figure}[t]
\centering
\includegraphics{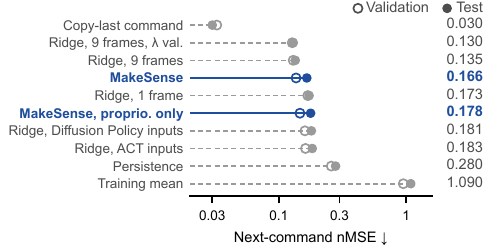}
\caption{\textbf{Reference predictors for next-command estimation.} Humanoid-Everyday next-command nMSE on test (filled markers, values on the right) and validation (open markers) on a log scale; dashed grey: closed-form references, of which only copy-last reads past commands; \ours{}: mean\,$\pm$\,SD over three seeds, and ``proprio.\ only'' the same weights with proprioception alone.}
\label{fig:supp_he_ref}
\end{figure}

\begin{figure*}[t]
\centering
\includegraphics{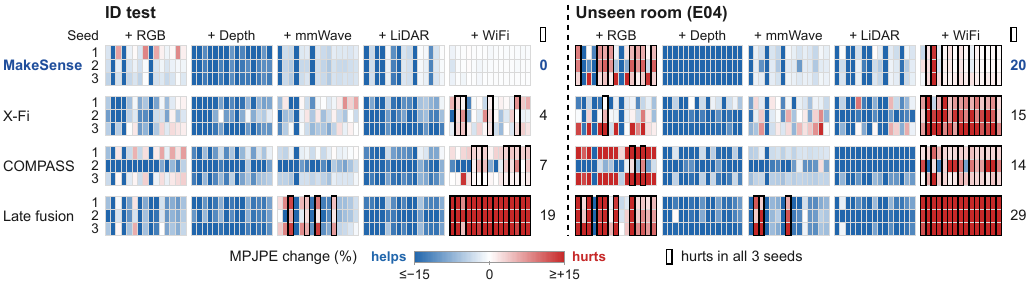}
\caption{\textbf{In distribution, no single-sensor addition degrades \ours{} across all three seeds.} All 75 single-sensor additions on MM-Fi, one row per seed, coloured by relative MPJPE change (clipped to $\pm$15\,\%), with outlined columns (tallied on the right) for additions that raise error in all three seeds; in the unseen room, \ours{} still has the fewest harmful additions at every margin (Table~\ref{tab:counts}), despite tighter cross-seed consistency (20 vs.\ 15 for X-Fi and 14 for COMPASS).}
\label{fig:supp_strips}
\end{figure*}

\subsection{Reference predictors on Humanoid-Everyday}\label{supp:he_ref}
\noindent Following ACT and Diffusion Policy, none of the learned models in this paper take past commands as input; Fig.~\ref{fig:supp_he_ref}\label{fref:fig:supp_he_ref} provides closed-form references for comparison, including a copy-last baseline that does. The measured arm joints follow the commanded pose by about 8 to 10 frames (about 0.3\,s), so their history carries the recent commands.

\section{\texorpdfstring{Negative fusion on MM-Fi\\\textnormal{(main paper, Secs.~5.1--5.3)}}{Negative fusion on MM-Fi (main paper, Secs. 5.1\textendash 5.3)}}
\subsection{Harmful additions per seed}\label{supp:perseed}
\noindent In distribution, no single-sensor addition degrades \ours{} across all three seeds, compared with 4 for X-Fi and 7 for COMPASS (Fig.~\ref{fig:supp_strips}\label{fref:fig:supp_strips}). Table~\ref{tab:counts} counts the harmful additions at margins of 0, 2 and 5\,\% and with 95\,\% intervals.

\begin{table}[t]
\centering\scriptsize\renewcommand{\arraystretch}{1.0}\setlength{\tabcolsep}{1.8pt}
\caption{\textbf{On seed means, \ours{} has the fewest harmful additions at every margin.} Neg.\,($\tau$): additions that raise the error by more than $\tau$ (mean\,$\pm$\,SD over three seeds); all seeds: worse in every seed; Neg.\,(CI): 95\,\% interval above 0, from 2{,}000 resamplings of the test recordings (81 ID, 270 E04), not of subjects.}\label{tab:counts}
\begin{tabular}{@{}l rrrrr@{}}
\toprule
 & \mh{Neg.\,(0)} & \mh{Neg.\,(2\%)} & \mh{Neg.\,(5\%)} & \mh{All seeds} & \mh{Neg.\,(CI)} \\
\midrule
\multicolumn{6}{@{}l}{\textit{MPJPE, ID test (75 additions)}} \\
\ours{} & 3.3\pmsdt{4.9} & 1.3\pmsdt{2.3} & 1.3\pmsdt{2.3} & 0 & 1.3\pmsdt{2.3} \\
X-Fi & 16.0\pmsdt{2.6} & 7.0\pmsdt{1.7} & 2.3\pmsdt{2.5} & 4 & 7.0\pmsdt{2.0} \\
COMPASS & 18.0\pmsdt{9.6} & 10.3\pmsdt{7.5} & 2.0\pmsdt{1.7} & 7 & 9.7\pmsdt{6.4} \\
Late fusion & 22.0\pmsdt{2.6} & 20.7\pmsdt{3.2} & 18.3\pmsdt{2.5} & 19 & 20.3\pmsdt{2.9} \\
\addlinespace[2pt]
\multicolumn{6}{@{}l}{\textit{MPJPE, unseen room E04 (75)}} \\
\ours{} & 23.3\pmsdt{2.5} & 13.0\pmsdt{1.7} & 7.3\pmsdt{2.1} & 20 & 21.0\pmsdt{2.6} \\
X-Fi & 27.7\pmsdt{9.3} & 24.3\pmsdt{8.7} & 19.0\pmsdt{8.2} & 15 & 26.0\pmsdt{9.2} \\
COMPASS & 24.0\pmsdt{7.8} & 22.7\pmsdt{7.1} & 14.0\pmsdt{2.6} & 14 & 23.7\pmsdt{7.5} \\
Late fusion & 30.7\pmsdt{1.5} & 30.0\pmsdt{1.0} & 26.3\pmsdt{3.5} & 29 & 30.0\pmsdt{1.0} \\
\midrule
\multicolumn{6}{@{}l}{\textit{PA-MPJPE, ID test (75 additions)}} \\
\ours{} & 8.0\pmsdt{1.7} & 0.0\pmsdt{0.0} & 0.0\pmsdt{0.0} & 3 & 2.3\pmsdt{1.5} \\
X-Fi & 12.3\pmsdt{4.5} & 4.3\pmsdt{2.5} & 0.7\pmsdt{0.6} & 3 & 7.3\pmsdt{3.1} \\
COMPASS & 18.7\pmsdt{7.1} & 6.7\pmsdt{2.1} & 1.7\pmsdt{1.5} & 2 & 9.7\pmsdt{3.2} \\
Late fusion & 23.3\pmsdt{0.6} & 19.7\pmsdt{2.1} & 16.3\pmsdt{1.5} & 22 & 20.0\pmsdt{1.7} \\
\addlinespace[2pt]
\multicolumn{6}{@{}l}{\textit{PA-MPJPE, unseen room E04 (75)}} \\
\ours{} & 24.7\pmsdt{3.8} & 13.3\pmsdt{2.9} & 11.7\pmsdt{3.2} & 16 & 22.3\pmsdt{5.1} \\
X-Fi & 29.7\pmsdt{3.8} & 21.3\pmsdt{6.0} & 15.0\pmsdt{3.5} & 23 & 28.3\pmsdt{3.5} \\
COMPASS & 27.3\pmsdt{2.1} & 18.3\pmsdt{1.2} & 12.0\pmsdt{4.4} & 11 & 24.7\pmsdt{1.5} \\
Late fusion & 30.0\pmsdt{1.0} & 28.7\pmsdt{1.5} & 24.7\pmsdt{2.5} & 28 & 29.3\pmsdt{1.2} \\
\bottomrule
\end{tabular}
\end{table}

\begin{figure*}[t]
\centering
\includegraphics{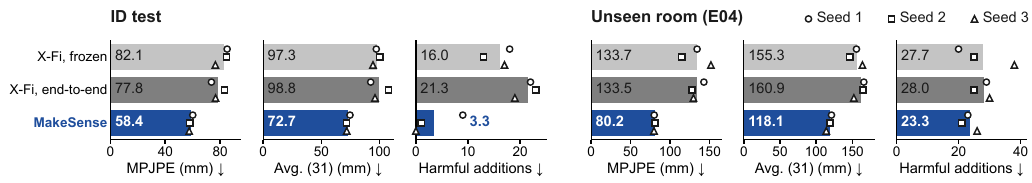}
\caption{\textbf{End-to-end training of X-Fi's encoders does not close the performance gap to \ours{}.} MM-Fi ID test (left) and unseen room E04 (right), three-seed means (bars and values) and individual seeds (markers); \ours{} achieves the lowest mean in all six panels.}
\label{fig:supp_xfi_e2e}
\end{figure*}

\begin{figure}[t]
\centering
\includegraphics{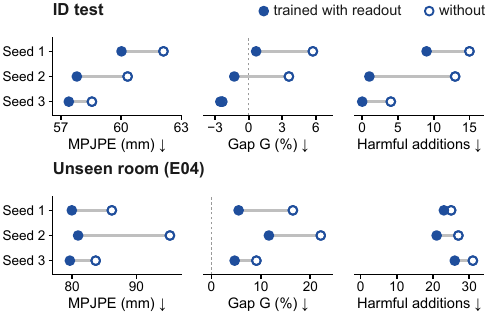}
\caption{\textbf{Training with the range readout lowers both five-sensor error and harmful additions in every seed.} MM-Fi ID test (top) and unseen room E04 (bottom), $G$ comparing all five sensors against depth, mmWave and LiDAR; filled: \ours{}, open: the same model trained without the readout (in Seed~3 the two ID gaps nearly coincide).}
\label{fig:supp_readout_seeds}
\end{figure}

\subsection{\texorpdfstring{Training with and without\\ the readout, per seed}{Training with and without the readout, per seed}}\label{supp:readout_seeds}
\noindent Across every seed, training with the range readout reduces both five-sensor MPJPE and the number of harmful single-sensor additions (Fig.~\ref{fig:supp_readout_seeds}\label{fref:fig:supp_readout_seeds}).

\begin{figure}[t]
\centering
\includegraphics{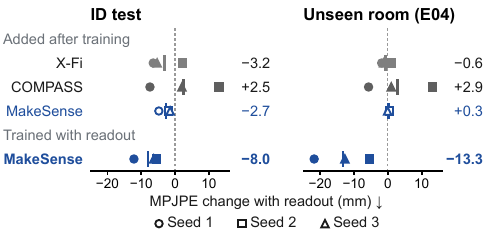}
\caption{\textbf{Applying the readout post hoc yields no consistent gain for X-Fi or COMPASS.} Change in five-sensor MPJPE on MM-Fi when the range readout is applied after training to stored predictions of models trained without it (markers: seeds; ticks and numbers: means in mm), and for \ours{} trained with the readout, enabled against~disabled~at~test~time.}
\label{fig:supp_posthoc}
\end{figure}

\subsection{Readout switched off or added after training}\label{supp:posthoc}
\noindent The range readout improves accuracy on both test sets when incorporated during training, whereas applying it post hoc gives X-Fi and COMPASS no consistent benefit (Fig.~\ref{fig:supp_posthoc}\label{fref:fig:supp_posthoc}).

\begin{figure*}[t]
\centering
\includegraphics{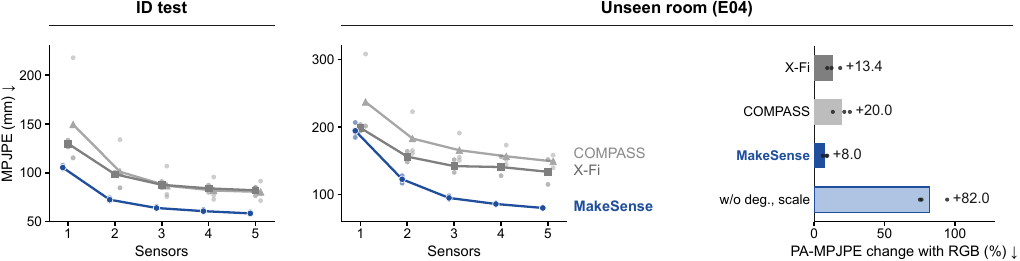}
\caption{\textbf{Mean error falls for all models as subsets grow, while adding RGB in the unseen room raises pose error---least severely for \ours{}.} MM-Fi test, three-seed means (lines, bars) and individual seeds (dots); right: PA-MPJPE change when RGB joins depth, mmWave and LiDAR (``w/o deg., scale'' omits degradation and image-scale training).}
\label{fig:supp_decomp}
\end{figure*}

\subsection{End-to-end X-Fi}\label{supp:xfi_e2e}
\noindent Unfreezing and training X-Fi's encoders end to end does not close the gap to \ours{}.

With all five sensors, \ours{} achieves a 25\,\% lower error than end-to-end X-Fi in distribution and a 40\,\% lower error in the unseen room on three-seed means, outperforming it in every seed~(Fig.~\ref{fig:supp_xfi_e2e}\label{fref:fig:supp_xfi_e2e}). The 21\,\% of Sec.~5.3 of the main paper compares the first X-Fi seed (73.6\,mm with all five sensors in distribution) with the three-seed mean of \ours{}. Relative to frozen encoders, end-to-end training lowers X-Fi's average in-distribution five-sensor error but increases its harmful additions in every seed, while leaving both metrics largely unchanged on average in~the~\mbox{unseen}~room.

\subsection{Which sensor, which part of the error}\label{supp:decomp}
\noindent While mean error decreases for all three models as subset size increases, adding RGB in the unseen room degrades pose accuracy across every model; \ours{} suffers the smallest rise, largely mitigated by degradation and image-scale training.

Fig.~\ref{fig:supp_decomp}\label{fref:fig:supp_decomp} shows each model's error across sensor counts alongside the impact of adding RGB to depth, mmWave, and LiDAR.

\subsection{MM-Fi on X-Fi's own split}\label{supp:xfi_split}
\noindent On X-Fi's original split, \ours{} outperforms the published errors of X-Fi across all 31 subsets on both metrics and surpasses COMPASS at every sensor count (Table~\ref{tab:xfi_split}).

\begin{table}[t]
\centering\scriptsize\setlength{\tabcolsep}{2.2pt}
\caption{\textbf{MM-Fi evaluation on X-Fi's original split.} Mean errors (mm) by sensor count and over all 31 subsets; \ours{}: mean\,$\pm$\,SD over three seeds; X-Fi and COMPASS: their published tables (X-Fi Table~1; COMPASS Tables~3 and~4) under their original training recipes.}\label{tab:xfi_split}
\begin{tabular}{@{}lrrrrrr@{}}
\toprule
 & \multicolumn{3}{c}{MPJPE\dn} & \multicolumn{3}{c}{PA-MPJPE\dn} \\
\cmidrule(lr){2-4}\cmidrule(l){5-7}
 & \ours{} & X-Fi & COMPASS & \ours{} & X-Fi & COMPASS \\
\midrule
1 sensor & \textbf{98.7\pmsdt{1.0}} & 143.2 & 122.1 & \textbf{58.2\pmsdt{0.2}} & 77.4 & 73.1 \\
2 sensors & \textbf{65.7\pmsdt{0.7}} & 105.0 & 86.4 & \textbf{42.7\pmsdt{0.2}} & 59.7 & 56.6 \\
3 sensors & \textbf{57.3\pmsdt{0.3}} & 92.6 & 74.7 & \textbf{39.1\pmsdt{0.2}} & 52.3 & 49.9 \\
4 sensors & \textbf{53.8\pmsdt{0.3}} & 87.9 & 70.5 & \textbf{37.7\pmsdt{0.1}} & 49.6 & 48.0 \\
5 sensors & \textbf{51.8\pmsdt{0.3}} & 83.7 & 69.2 & \textbf{36.8\pmsdt{0.1}} & 47.6 & 47.4 \\
Avg.\,(31) & \textbf{66.0\pmsdt{0.5}} & 103.7 & 85.3 & \textbf{43.1\pmsdt{0.2}} & 58.2 & 55.4 \\
\bottomrule
\end{tabular}
\end{table}

In the paired-addition test (main paper, Sec.~4), 11 of 75 single-sensor additions raise X-Fi's published error (up to +1.7\,\%), versus only 6 for \ours{}'s seed mean at 0.1\,mm precision (up to +0.3\,\%; per seed: 5.0\pmsdt{1.0}, up to +0.6\,\%). Adding RGB and WiFi CSI to depth, mmWave, and LiDAR lowers error for both models (\ours{}: $-$3.2\pmsdt{0.6}\,\% to 51.8\,mm; X-Fi: $-$12.8\,\% to 83.7\,mm).

\ours{} uses the main paper's MM-Fi recipe on X-Fi's random split (MM-Fi toolbox protocol~3: 32 training and 8 test subjects per action) and X-Fi's evaluation code. Unlike X-Fi (which reads 8-bit depth images), \ours{} takes metric depth and predicts the training-mean pose on the 404 frames with empty mmWave clouds. These weights are used solely in this section.

\subsection{Core and losses over three seeds}\label{supp:core_seeds}
\noindent In the unseen room, the full design has a 6$\times$ ($\pm$0.7 vs.\ $\pm$4.3--6.6\,mm) smaller seed SD; the bottleneck core lowers $G$ in every seed, and $\mathcal L_{\rm KD}$, $\mathcal L_{\rm MoFe}$ lower it on average~(Table~\ref{tab:core_seeds}, completing Table~5 to three seeds).

\begin{table}[t]
\centering\scriptsize\setlength{\tabcolsep}{3pt}
\caption{\textbf{Three-seed ablations of the fusion core and loss terms.} MM-Fi test, mean\,$\pm$\,SD over three seeds (MPJPE and Avg.\,(31) in mm); $G$: error change (\%) when RGB and WiFi CSI join depth, mmWave and LiDAR; Neg.: harmful single-sensor additions out of 75.}\label{tab:core_seeds}
\begin{tabular}{@{}lrrrr@{}}
\toprule
 & MPJPE\dn & Avg.\,(31)\dn & $G$\dn & Neg.\dn \\
\midrule
\multicolumn{5}{@{}l}{\textit{ID test}} \\
\ours{} & 58.4\pmsdt{1.4} & 72.7\pmsdt{1.7} & $-$1.1\pmsdt{1.6} & 3.3\pmsdt{4.9} \\
Concat.\ core & 58.1\pmsdt{1.2} & 72.6\pmsdt{1.1} & $-$1.0\pmsdt{1.8} & 6.3\pmsdt{4.6} \\
w/o $\mathcal L_{\rm KD}$ & 57.8\pmsdt{0.9} & 72.2\pmsdt{0.8} & $-$2.2\pmsdt{1.4} & 2.7\pmsdt{1.5} \\
w/o $\mathcal L_{\rm MoFe}$ & 58.8\pmsdt{1.1} & 73.5\pmsdt{1.2} & $-$0.3\pmsdt{2.2} & 7.3\pmsdt{4.7} \\
\multicolumn{5}{@{}l}{\textit{Unseen room (E04)}} \\
\ours{} & 80.2\pmsdt{0.7} & 118.1\pmsdt{3.7} & +7.3\pmsdt{3.8} & 23.3\pmsdt{2.5} \\
Concat.\ core & 84.0\pmsdt{5.1} & 120.8\pmsdt{4.7} & +11.4\pmsdt{6.7} & 25.7\pmsdt{0.6} \\
w/o $\mathcal L_{\rm KD}$ & 84.9\pmsdt{4.3} & 122.6\pmsdt{5.3} & +9.3\pmsdt{1.9} & 25.0\pmsdt{3.6} \\
w/o $\mathcal L_{\rm MoFe}$ & 83.2\pmsdt{6.6} & 119.6\pmsdt{5.0} & +10.0\pmsdt{9.9} & 24.7\pmsdt{1.2} \\
\bottomrule
\end{tabular}
\end{table}

\subsection{Placement and pose of the skeleton}\label{supp:placement}
\noindent The readout changes the output only when RGB or WiFi CSI joins a range sensor, and it lowers the root error while the root-aligned and PA-MPJPE errors stay the same (Table~\ref{tab:placement}).

\begin{table*}[t]
\centering\scriptsize\renewcommand{\arraystretch}{1.0}
\setlength{\tabcolsep}{4.5pt}
\caption{\textbf{The readout corrects the placement of the skeleton, not its pose.} MM-Fi test, the same weights with the readout off and on (mean over the subsets of a group, then over three seeds; change: mean\,$\pm$\,SD over seeds); root: error of the root joint; root-aligned: MPJPE after both root joints coincide.}
\label{tab:placement}
\begin{tabular}{@{}l rrr rr rr@{}}
\toprule
 & \multicolumn{3}{c}{MPJPE\dn} & \multicolumn{2}{c}{Root\dn} & \multicolumn{2}{c}{Pose, on = off} \\
\cmidrule(lr){2-4}\cmidrule(lr){5-6}\cmidrule(l){7-8}
Subsets (number) & off & on & change (\%) & off & on & root-aligned\dn & PA-MPJPE\dn \\
\midrule
\multicolumn{8}{@{}l}{\textit{ID test}} \\
RGB and a range sensor (14) & 75.0 & 64.6 & $-$13.6\pmsdt{4.8} & 64.2 & 51.1 & 56.9 & 41.1 \\
WiFi CSI and a range sensor, no RGB (7) & 68.9 & 68.5 & $-$0.6\pmsdt{0.3} & 58.8 & 58.5 & 60.1 & 44.5 \\
Range sensors only (7) & 68.7 & 68.7 & 0.0 & 58.6 & 58.6 & 59.9 & 44.4 \\
No range sensor (3) & 129.5 & 129.5 & 0.0 & 106.7 & 106.7 & 84.7 & 65.2 \\
\addlinespace[2pt]
\multicolumn{8}{@{}l}{\textit{Unseen room (E04)}} \\
RGB and a range sensor (14) & 126.6 & 98.8 & $-$21.0\pmsdt{11.6} & 99.7 & 76.1 & 81.0 & 62.0 \\
WiFi CSI and a range sensor, no RGB (7) & 104.8 & 104.1 & $-$0.6\pmsdt{0.4} & 82.1 & 81.6 & 75.9 & 58.4 \\
Range sensors only (7) & 99.8 & 99.8 & 0.0 & 78.4 & 78.4 & 74.6 & 57.8 \\
No range sensor (3) & 283.2 & 283.2 & 0.0 & 241.7 & 241.7 & 119.4 & 94.2 \\
\bottomrule
\end{tabular}
\end{table*}

\begin{figure*}[t]
\centering
\includegraphics{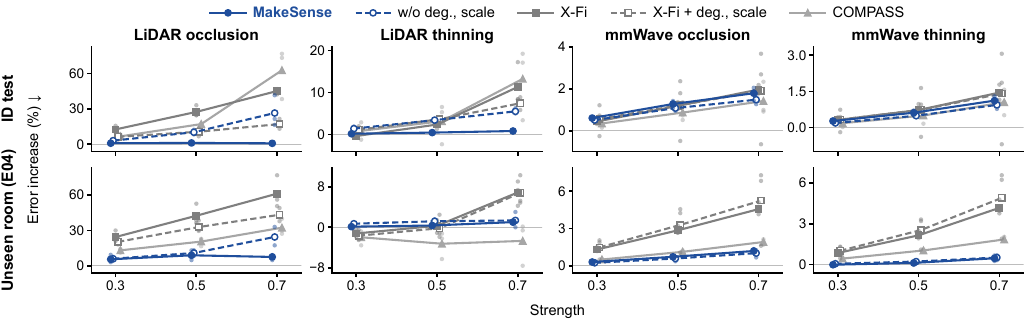}
\caption{\textbf{Under LiDAR occlusion, \ours{} exhibits the smallest accuracy drop at every severity across both test splits.} Relative error increase over each model's clean five-sensor baseline on MM-Fi (lines: three-seed means; dots: individual seeds).}
\label{fig:supp_dose}
\end{figure*}

\begin{figure*}[t]
\centering
\includegraphics{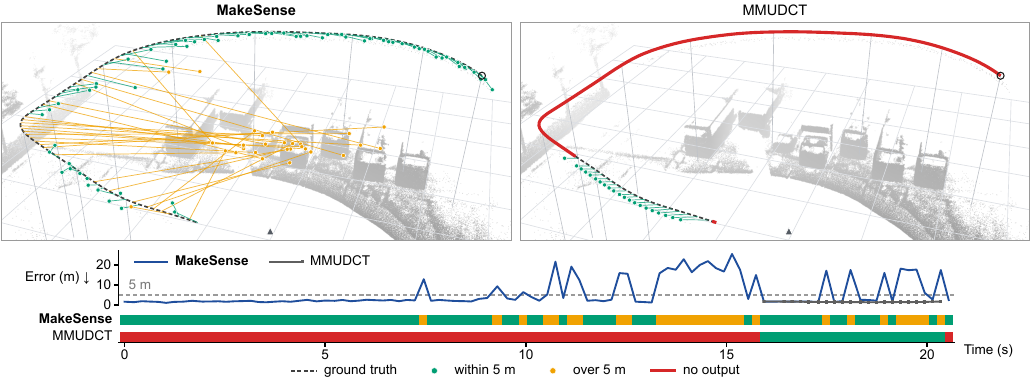}
\caption{\textbf{3D trajectory comparison on MMAUD: \ours{} maintains 100\,\% coverage, whereas MMUDCT misses 79 of 100 steps.} Single-seed visualization of main-paper Fig.~7, joining predictions to ground truth; across the test set (three-seed mean), MMUDCT leaves 49.0\,\% of steps silent, attaining 1.50\,m error vs.\ 8.13\,m for \ours{} only on its 51.0\,\% covered subset.}
\label{fig:supp_mmaud_3d}
\end{figure*}

\section{\texorpdfstring{Robustness to degradation\\\textnormal{(main paper, Sec.~5.3)}}{Robustness to degradation (main paper, Sec. 5.3)}}
\noindent Under test-time faults, \ours{} exhibits the smallest average accuracy drop on both MM-Fi test splits.

\subsection{Three strengths on MM-Fi}\label{supp:dose}
\noindent Under LiDAR occlusion, \ours{} suffers the smallest accuracy drop across all three severity levels on both MM-Fi test splits; in distribution, its error grows by less than 2\,\% across every fault type and strength (Fig.~\ref{fig:supp_dose}\label{fref:fig:supp_dose}).

On Humanoid-Everyday, \ours{} achieves the lowest error among all six models under each of the four faults at strengths 0.5 and 0.7.

\subsection{Seeds per row and the cost on clean inputs}\label{supp:robust_sd}
\noindent\looseness=-1 Degradation training (with image-scale augmentation on MM-Fi) substantially curbs fault-induced error on MM-Fi, MMAUD, and V2X-Real, while any clean-input cost on MMAUD, Humanoid-Everyday, and V2X-Real stays within cross-seed variance.

\begin{figure*}[t]
\centering
\includegraphics{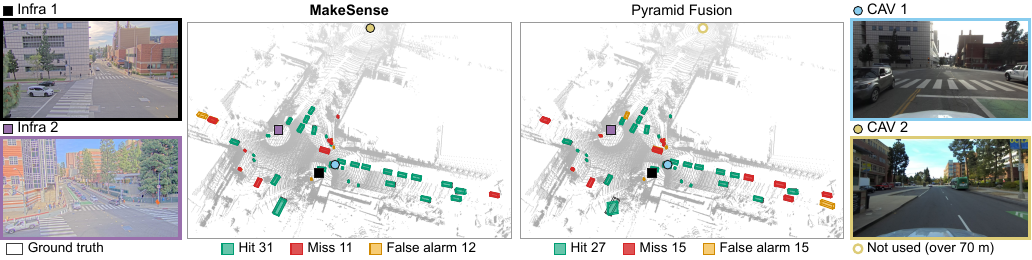}
\caption{\textbf{3D detection comparison on V2X-Real: \ours{} misses 11 of 42 objects versus 15 for Pyramid Fusion.} Single-seed visualization of the Fig.~6 clip of the main paper at 4.9\,s, the median-margin frame among the 91 of 106 frames where \ours{} achieves higher F1; grey: LiDAR from all four nodes (Pyramid Fusion uses only nodes within 70\,m), with camera views for context.}
\label{fig:supp_v2x_3d}
\end{figure*}

Extending the no-degradation controls of main-paper Table~3 (one seed on MMAUD, two on V2X-Real) to three seeds confirms the same advantage: the MMAUD control reaches 11.50\pmsdt{1.64}\,m on clean inputs and degrades by +25.8\pmsdt{7.7}\,\% across the four faults, versus +13.7\pmsdt{2.0}\,\% for \ours{}. Similarly, the three-seed V2X-Real control scores 44.5\pmsdt{0.8} AP50 on clean inputs (versus 44.4\pmsdt{1.2} for \ours{}) and loses 9.1\pmsdt{0.8}\,\% under the four faults, versus 7.3\pmsdt{0.1}\,\% for \ours{}. All remaining clean-input figures match Table~3 of the main paper, and full per-fault breakdowns are released with~the~code.

\subsection{Keeping or dropping a degraded sensor}\label{supp:keep_drop}
\begin{figure*}[t]
\centering
\includegraphics{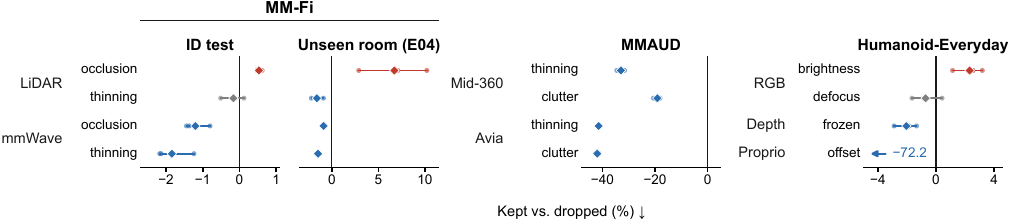}
\caption{\textbf{For most faults, keeping a degraded sensor gives a lower error than dropping it.} \ours{}, strength 0.5: error with the degraded sensor kept over the error of the same model on clean inputs without it, minus 1 (blue: keeping lower in all three seeds; red: dropping lower in all~three;~grey:~mixed).}
\label{fig:supp_keep_drop}
\end{figure*}

\noindent Keeping a degraded sensor lowers error in all three seeds on 11 of 16 faults, showing useful signal extraction despite corruption; only MM-Fi LiDAR occlusion and HE RGB brightness favour dropping it (Fig.~\ref{fig:supp_keep_drop}).

\section{\texorpdfstring{Other datasets \textnormal{(main paper, Sec.~5.4)}}{Other datasets (main paper, Sec. 5.4)}}
\noindent Across the four remaining benchmarks, the same design outperforms the retrained baselines; below, we break down the source of each gain.

\begin{figure*}[t]
\centering
\includegraphics{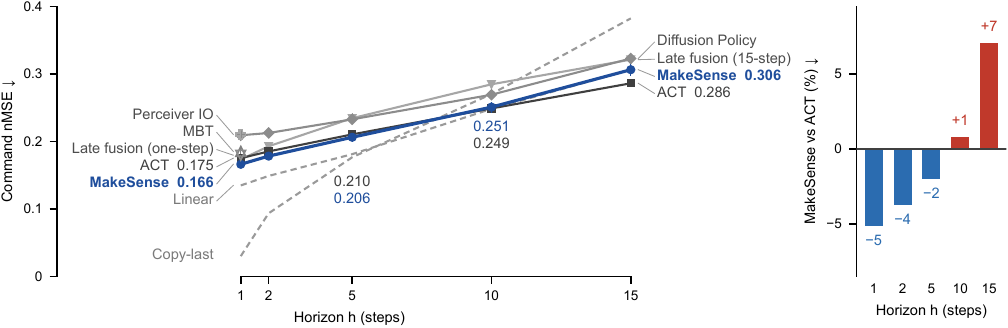}
\caption{\textbf{Among learned models, \ours{} achieves the lowest command error up to a five-step horizon.} Three-seed mean command nMSE on Humanoid-Everyday $h$ steps ahead (values printed for \ours{} and ACT; dashed: references of Sec.~\ref{supp:he_ref}; open: one-step models); right: relative error difference between \ours{} and ACT.}
\label{fig:supp_he_horizon}
\end{figure*}

\begin{figure*}[t]
\centering
\includegraphics[width=0.98\textwidth]{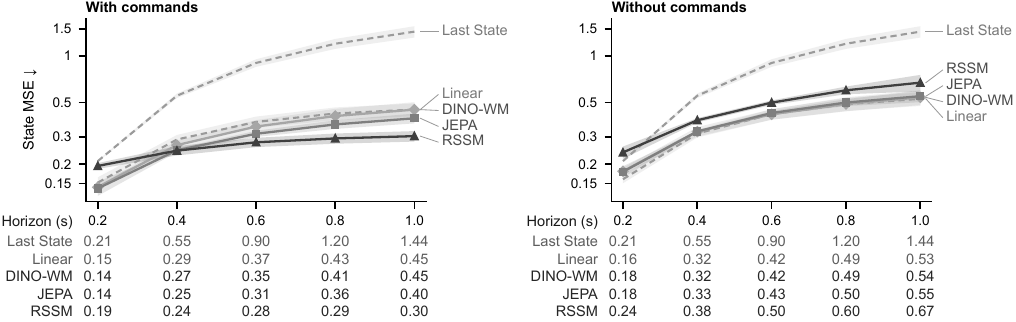}
\caption{\textbf{Action commands drive the advantage of learned world-model predictors.} Standardized fused-state MSE on Humanoid-Everyday with commands (left) and without (right) on a log scale; bands: $\pm$SD over three seeds; dashed: reference baselines; rows below: mean values.}
\label{fig:supp_wm_full}
\end{figure*}

\begin{figure}[t]
\centering
\includegraphics{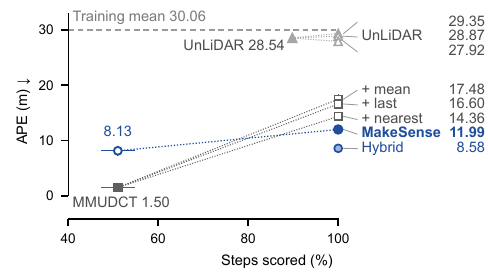}
\caption{\textbf{Across all test steps, \ours{} achieves lower error than MMUDCT under every fill rule.} Position error (m) against coverage, three-seed means (UnLiDAR: one run); open grey: MMUDCT with silent steps filled by the training-mean position (main paper), its last valid output or its nearest output (the training mean in fully silent sequences); open blue: \ours{} on MMUDCT's covered steps; hybrid: MMUDCT when available, else \ours{}.}
\label{fig:supp_mmaud}
\end{figure}

\begin{figure}[t]
\centering
\includegraphics{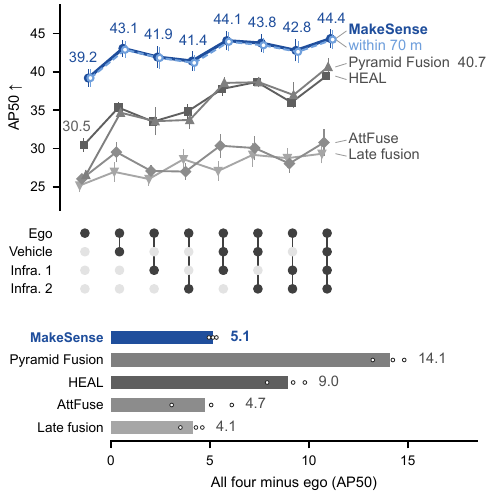}
\caption{\textbf{\ours{} achieves the highest AP50 on every node subset, combining a strong ego detector with consistent multi-node gains.} Top: 3-seed mean V2X-Real test AP50 per ego-inclusive subset (dots: active nodes; grey: strongest baseline; dashed: within 70\,m, differing by at most 0.5 points); bottom: cooperative gain from ego-only to all four nodes.}
\label{fig:supp_v2x_nodes}
\end{figure}

\subsection{MMAUD: scoring silent steps}\label{supp:mmaud}
\noindent \ours{} predicts continuously and outperforms MMUDCT across all steps under every fill rule (11.99\,m vs.\ 14.36--17.48\,m); pairing \ours{}'s full coverage with MMUDCT's sparse high-precision detections further lowers error to 8.58\,m (Fig.~\ref{fig:supp_mmaud}\label{fref:fig:supp_mmaud}).

Fig.~\ref{fig:supp_mmaud_3d}\label{fref:fig:supp_mmaud_3d} provides a 3D visualization of the test sequence from Fig.~7 of the main paper.

\subsection{V2X-Real: ego detector and cooperation}\label{supp:v2x}
\noindent\looseness=-1 \ours{} achieves the highest AP50 and AP70 across every node subset---even within the baselines' 70\,m range---leading by 8.7 AP50 ego-only (nearly matching four-node HEAL) and still gaining +5.1 AP50 from cooperation to lead all-node fusion by 3.7 points (Fig.~\ref{fig:supp_v2x_nodes}\label{fref:fig:supp_v2x_nodes}, Table~\ref{tab:sub_v2x}).

Fig.~\ref{fig:supp_v2x_3d}\label{fref:fig:supp_v2x_3d} renders a 3D frame from the Fig.~6 clip of the main paper alongside each node's camera view.

\subsection{Humanoid-Everyday: longer horizons}\label{supp:he_horizon}
\noindent \ours{} outperforms ACT on short horizons (1--5 steps; 5\,\% lower one-step error) and matches it at 10 steps (0.251 vs.\ 0.249), despite 4$\times$ fewer trainable parameters and no ImageNet pre-training~(Fig.~\ref{fig:supp_he_horizon}\label{fref:fig:supp_he_horizon}).

At one step ahead, \ours{} (0.166) also reduces error by 8\,\% over one-step late fusion (0.180; main-paper Tables~2--3) and by 4\,\% over a separate 15-step late-fusion model (0.173).

\subsection{RELI11D: label budgets}\label{supp:reli}
\begin{figure*}[t]
\centering
\includegraphics{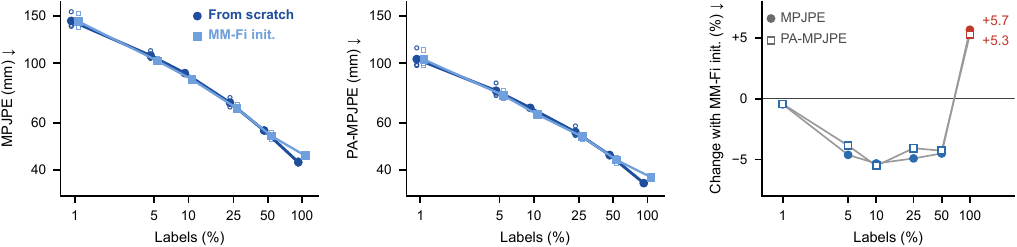}
\caption{\textbf{MM-Fi initialization consistently boosts low-label RELI11D accuracy (5--50\,\% labels), but full-data training converges best from scratch.} markers: runs; lines: means; right: change with MM-Fi initialization (positive is worse).}
\label{fig:supp_reli}
\end{figure*}

\noindent With 5 to 50\,\% of the labels, MM-Fi initialization lowers the RELI11D error in every draw; the main paper reports 10\,\% (Fig.~\ref{fig:supp_reli}). With all labels, \ours{} trained from scratch has the lowest error on every measure (Table~\ref{tab:reli_full}).

\section{\texorpdfstring{World model \textnormal{(main paper, Sec.~5.4)}}{World model (main paper, Sec. 5.4)}}
\subsection{All predictors and horizons}\label{supp:wm_full}
\noindent When conditioned on action commands, both JEPA and RSSM outperform the linear baseline in forecasting the fused state from 0.4\,s onward in every seed (Fig.~\ref{fig:supp_wm_full}\label{fref:fig:supp_wm_full}).

\looseness=-1 Extending Table~4 of the main paper (which reports 0.2, 0.6, and 1.0\,s), Fig.~\ref{fig:supp_wm_full} adds the intermediate 0.4 and 0.8\,s horizons alongside command-free ablations; removing commands raises forecasting error for every predictor at every horizon.

\section{\texorpdfstring{More results \textnormal{(main paper, Sec.~5)}}{More results (main paper, Sec. 5)}}\label{supp:more}

\subsection{All input subsets}\label{supp:subsets}
\noindent On MM-Fi, \ours{} has the lowest MPJPE on 29 of the 31 input subsets in distribution and on 28 in the unseen room, and each exception lacks a range sensor (Tables~\ref{tab:sub_mmfi_id} and~\ref{tab:sub_mmfi_ood}). It leads with every subset on Humanoid-Everyday and V2X-Real (Tables~\ref{tab:sub_he} and~\ref{tab:sub_v2x}); on MMAUD and RELI11D the baselines read fixed inputs (Tables~\ref{tab:sub_mmaud} and~\ref{tab:sub_reli}).

\begin{table}[t]
\centering\scriptsize\renewcommand{\arraystretch}{1.0}
\setlength{\tabcolsep}{2.2pt}
\caption{\textbf{RELI11D test, all labels: \ours{} has the lowest error on every measure.} Error in mm, mean\,$\pm$\,SD over three runs; root-al.: MPJPE after subtracting the root joint; root: error of the root joint (LiDARCap predicts no translation).}
\label{tab:reli_full}
\begin{tabular}{@{}l cccc@{}}
\toprule
 & \mh{MPJPE\dn} & \mh{Root-al.\dn} & \mh{PA-MPJPE\dn} & \mh{Root\dn} \\
\midrule
LEIR & 129.5\pmsdt{1.2} & 75.0\pmsdt{0.4} & 62.4\pmsdt{0.1} & 104.6\pmsdt{2.0} \\
LiDARCap & -- & 75.1\pmsdt{2.1} & 52.2\pmsdt{1.2} & -- \\
\ours{}, scratch & \textbf{42.7}\pmsdt{0.8} & \textbf{40.5}\pmsdt{0.6} & \textbf{35.7}\pmsdt{0.4} & \textbf{20.2}\pmsdt{0.7} \\
\ours{}, MM-Fi init. & 45.2\pmsdt{0.4} & 43.3\pmsdt{0.3} & 37.6\pmsdt{0.3} & 23.5\pmsdt{1.3} \\
\bottomrule
\end{tabular}
\end{table}

\begin{table}[!t]
\centering\scriptsize\renewcommand{\arraystretch}{1.0}
\setlength{\tabcolsep}{4pt}
\caption{\textbf{MMAUD test, all 15 input subsets of \ours{}.} APE in m and coverage in \%, mean\,$\pm$\,SD over three seeds; a step without input is scored at the training-mean position, as the silent steps of MMUDCT in Table~2 of the main paper.}
\label{tab:sub_mmaud}
\begin{tabular}{@{}l r r@{}}
\toprule
 & \multicolumn{2}{c@{}}{\ours{}} \\
\cmidrule(l){2-3}
Input & APE\dn & Coverage\up \\
\midrule
R & 30.76\pmsdt{1.25} & 100.0\pmsdt{0.0} \\
Mid-360 & 21.85\pmsdt{0.62} & 97.9\pmsdt{0.0} \\
Avia & 21.63\pmsdt{0.36} & 55.9\pmsdt{0.0} \\
radar & 28.47\pmsdt{0.00} & 7.4\pmsdt{0.0} \\
\addlinespace[2pt]
R + Mid-360 & 21.94\pmsdt{0.57} & 100.0\pmsdt{0.0} \\
R + Avia & 20.37\pmsdt{0.35} & 100.0\pmsdt{0.0} \\
R + radar & 29.15\pmsdt{1.04} & 100.0\pmsdt{0.0} \\
Mid-360 + Avia & 11.97\pmsdt{0.33} & 98.3\pmsdt{0.0} \\
Mid-360 + radar & 20.93\pmsdt{0.59} & 97.9\pmsdt{0.0} \\
Avia + radar & 21.53\pmsdt{0.35} & 56.1\pmsdt{0.0} \\
\addlinespace[2pt]
R + Mid-360 + Avia & 12.03\pmsdt{0.24} & 100.0\pmsdt{0.0} \\
R + Mid-360 + radar & 21.03\pmsdt{0.56} & 100.0\pmsdt{0.0} \\
R + Avia + radar & 20.28\pmsdt{0.36} & 100.0\pmsdt{0.0} \\
Mid-360 + Avia + radar & 11.94\pmsdt{0.34} & 98.3\pmsdt{0.0} \\
\addlinespace[2pt]
R + Mid-360 + Avia + radar & 11.99\pmsdt{0.25} & 100.0\pmsdt{0.0} \\
\midrule
Avg.\,(15) & 20.39\pmsdt{0.22} & 87.5\pmsdt{0.0} \\
\bottomrule
\end{tabular}
\end{table}

\begin{table}[!t]
\centering\scriptsize\renewcommand{\arraystretch}{1.0}
\setlength{\tabcolsep}{4pt}
\caption{\textbf{RELI11D test, all 7 input subsets of \ours{}.} Error in mm, mean\,$\pm$\,SD over three seeds with all labels; without LiDAR the skeleton is placed at the mean position of the training windows, so~MPJPE~\mbox{exceeds}~1\,m.}
\label{tab:sub_reli}
\begin{tabular}{@{}l r r@{}}
\toprule
 & \multicolumn{2}{c@{}}{\ours{}} \\
\cmidrule(l){2-3}
Input & PA-MPJPE\dn & MPJPE\dn \\
\midrule
R & 55.5\pmsdt{0.2} & 1074.6\pmsdt{0.4} \\
event & 139.5\pmsdt{0.9} & 1083.7\pmsdt{0.2} \\
L & 36.4\pmsdt{0.9} & 43.8\pmsdt{1.4} \\
\addlinespace[2pt]
R + event & 55.4\pmsdt{0.1} & 1074.1\pmsdt{0.4} \\
R + L & 35.7\pmsdt{0.4} & 42.8\pmsdt{0.8} \\
event + L & 36.3\pmsdt{0.9} & 43.6\pmsdt{1.4} \\
\addlinespace[2pt]
R + event + L & 35.7\pmsdt{0.4} & 42.7\pmsdt{0.8} \\
\midrule
Avg.\,(7) & 56.4\pmsdt{0.4} & 486.5\pmsdt{0.7} \\
\bottomrule
\end{tabular}
\end{table}

\begin{figure*}[t]
\centering
\includegraphics{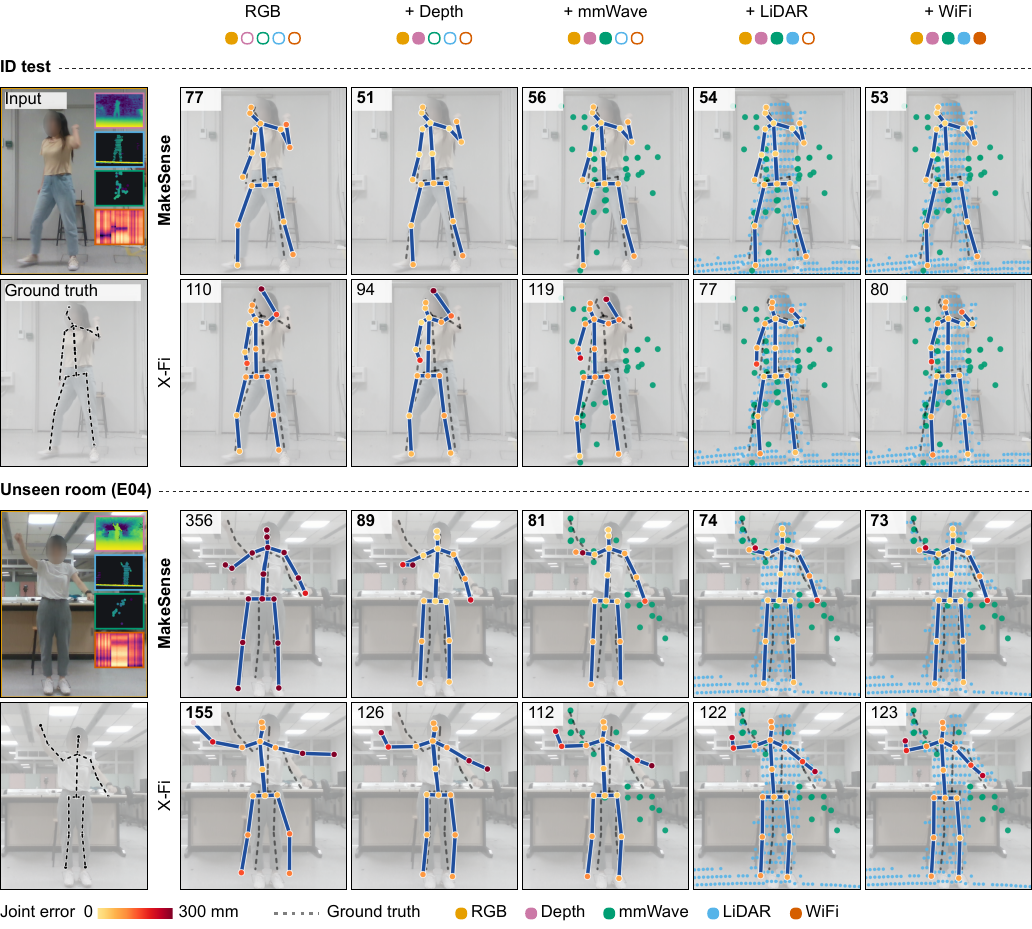}
\caption{\textbf{Once any second sensor joins RGB, \ours{} outperforms X-Fi on 87--96\,\% of frames across both splits.} Sensor-addition progression on median MM-Fi frames (one seed; MPJPE in mm); with RGB alone, X-Fi is lower on the unseen-room example, after which \ours{} leads at every subsequent step.}
\label{fig:supp_sensor_chain}
\end{figure*}

\begin{table*}[!t]
\centering\scriptsize\renewcommand{\arraystretch}{1.0}
\setlength{\tabcolsep}{3pt}
\caption{\textbf{MM-Fi ID test, all 31 input subsets: \ours{} has the lowest MPJPE on 29 and the lowest PA-MPJPE on all 31.} Error in mm, mean\,$\pm$\,SD over three seeds (late fusion: three runs, of which Table~2 of the main paper lists the first); R, D, M, L, W: RGB, depth, mmWave, LiDAR, WiFi CSI; bold:~lowest~mean.}
\label{tab:sub_mmfi_id}
\begin{tabular}{@{}l rrrr rrrr@{}}
\toprule
 & \multicolumn{4}{c}{MPJPE\dn} & \multicolumn{4}{c@{}}{PA-MPJPE\dn} \\
\cmidrule(lr){2-5}\cmidrule(l){6-9}
Input & Late fusion & X-Fi & COMPASS & \ours{} & Late fusion & X-Fi & COMPASS & \ours{} \\
\midrule
R & \textbf{92.3}\pmsdt{4.5} & 114.8\pmsdt{8.1} & 111.6\pmsdt{38.7} & 97.4\pmsdt{10.1} & 53.8\pmsdt{2.8} & 62.1\pmsdt{4.9} & 56.8\pmsdt{8.0} & \textbf{46.0}\pmsdt{0.9} \\
D & 89.1\pmsdt{15.1} & 95.1\pmsdt{7.3} & 153.2\pmsdt{122.9} & \textbf{62.9}\pmsdt{0.5} & 47.3\pmsdt{1.2} & 57.5\pmsdt{2.9} & 75.2\pmsdt{46.5} & \textbf{42.5}\pmsdt{0.3} \\
M & 124.7\pmsdt{7.0} & 136.3\pmsdt{4.2} & 149.4\pmsdt{51.1} & \textbf{102.2}\pmsdt{1.2} & 68.0\pmsdt{2.1} & 76.9\pmsdt{1.9} & 76.6\pmsdt{20.3} & \textbf{57.5}\pmsdt{0.9} \\
L & 79.8\pmsdt{2.4} & 100.6\pmsdt{9.0} & 115.8\pmsdt{67.9} & \textbf{69.3}\pmsdt{0.5} & 49.0\pmsdt{1.7} & 55.0\pmsdt{2.7} & 56.5\pmsdt{15.3} & \textbf{45.0}\pmsdt{0.2} \\
W & \textbf{194.8}\pmsdt{2.4} & 203.2\pmsdt{4.0} & 218.1\pmsdt{16.1} & 195.3\pmsdt{3.7} & 109.7\pmsdt{1.1} & 105.9\pmsdt{2.2} & 110.9\pmsdt{2.0} & \textbf{103.4}\pmsdt{0.2} \\
\addlinespace[2pt]
R+D & 77.6\pmsdt{3.4} & 88.8\pmsdt{4.3} & 87.4\pmsdt{16.0} & \textbf{60.8}\pmsdt{0.9} & 47.1\pmsdt{1.7} & 53.1\pmsdt{2.6} & 54.0\pmsdt{10.2} & \textbf{40.7}\pmsdt{0.2} \\
R+M & 94.1\pmsdt{5.9} & 100.5\pmsdt{0.2} & 94.4\pmsdt{15.2} & \textbf{77.1}\pmsdt{3.2} & 53.6\pmsdt{0.7} & 57.4\pmsdt{1.7} & 57.7\pmsdt{11.3} & \textbf{43.7}\pmsdt{0.7} \\
R+L & 79.1\pmsdt{3.4} & 86.9\pmsdt{2.6} & 87.0\pmsdt{16.5} & \textbf{69.7}\pmsdt{3.2} & 47.2\pmsdt{1.6} & 51.2\pmsdt{1.5} & 53.3\pmsdt{8.1} & \textbf{43.2}\pmsdt{0.4} \\
R+W & 127.9\pmsdt{1.9} & 109.5\pmsdt{5.0} & 106.2\pmsdt{27.6} & \textbf{95.7}\pmsdt{10.4} & 74.2\pmsdt{2.5} & 59.0\pmsdt{1.8} & 59.3\pmsdt{12.3} & \textbf{46.0}\pmsdt{0.9} \\
D+M & 88.0\pmsdt{1.4} & 91.8\pmsdt{4.6} & 103.4\pmsdt{43.0} & \textbf{60.5}\pmsdt{0.7} & 50.8\pmsdt{0.5} & 55.0\pmsdt{1.6} & 59.3\pmsdt{20.2} & \textbf{41.1}\pmsdt{0.3} \\
D+L & 70.3\pmsdt{2.7} & 81.9\pmsdt{4.7} & 95.8\pmsdt{42.5} & \textbf{60.8}\pmsdt{0.3} & 44.5\pmsdt{0.8} & 49.8\pmsdt{1.6} & 54.1\pmsdt{13.7} & \textbf{41.4}\pmsdt{0.3} \\
D+W & 122.1\pmsdt{3.3} & 97.6\pmsdt{7.2} & 107.9\pmsdt{44.0} & \textbf{62.6}\pmsdt{0.4} & 69.5\pmsdt{1.1} & 57.4\pmsdt{4.6} & 58.4\pmsdt{17.5} & \textbf{42.6}\pmsdt{0.2} \\
M+L & 88.4\pmsdt{4.9} & 93.0\pmsdt{3.6} & 88.0\pmsdt{24.7} & \textbf{66.0}\pmsdt{0.3} & 51.3\pmsdt{0.4} & 52.6\pmsdt{1.9} & 56.9\pmsdt{17.9} & \textbf{43.0}\pmsdt{0.3} \\
M+W & 135.8\pmsdt{2.0} & 142.2\pmsdt{0.8} & 143.7\pmsdt{24.2} & \textbf{101.6}\pmsdt{0.7} & 78.3\pmsdt{0.7} & 80.6\pmsdt{2.6} & 79.4\pmsdt{22.0} & \textbf{58.3}\pmsdt{1.0} \\
L+W & 118.6\pmsdt{2.0} & 93.7\pmsdt{1.8} & 97.2\pmsdt{32.1} & \textbf{69.4}\pmsdt{0.4} & 70.2\pmsdt{0.3} & 54.4\pmsdt{2.4} & 55.6\pmsdt{14.3} & \textbf{45.1}\pmsdt{0.3} \\
\addlinespace[2pt]
R+D+M & 80.4\pmsdt{1.7} & 88.5\pmsdt{3.6} & 82.7\pmsdt{11.1} & \textbf{58.9}\pmsdt{1.3} & 48.1\pmsdt{0.8} & 52.1\pmsdt{2.3} & 52.1\pmsdt{7.3} & \textbf{39.5}\pmsdt{0.2} \\
R+D+L & 71.4\pmsdt{0.3} & 81.5\pmsdt{3.1} & 79.8\pmsdt{11.2} & \textbf{60.1}\pmsdt{1.2} & 44.9\pmsdt{1.3} & 49.1\pmsdt{1.9} & 52.6\pmsdt{9.0} & \textbf{40.2}\pmsdt{0.2} \\
R+D+W & 102.1\pmsdt{1.8} & 87.9\pmsdt{5.0} & 88.6\pmsdt{15.8} & \textbf{60.6}\pmsdt{0.9} & 60.4\pmsdt{1.9} & 52.4\pmsdt{2.5} & 53.3\pmsdt{8.6} & \textbf{40.7}\pmsdt{0.2} \\
R+M+L & 82.4\pmsdt{4.8} & 85.4\pmsdt{1.8} & 81.5\pmsdt{9.8} & \textbf{67.8}\pmsdt{3.6} & 48.2\pmsdt{0.7} & 50.5\pmsdt{1.3} & 52.5\pmsdt{7.2} & \textbf{41.6}\pmsdt{0.5} \\
R+M+W & 111.6\pmsdt{3.0} & 100.9\pmsdt{2.9} & 97.0\pmsdt{17.7} & \textbf{76.7}\pmsdt{3.5} & 65.2\pmsdt{1.5} & 56.9\pmsdt{1.4} & 57.4\pmsdt{10.5} & \textbf{43.6}\pmsdt{0.7} \\
R+L+W & 102.0\pmsdt{1.1} & 88.2\pmsdt{2.7} & 89.5\pmsdt{18.6} & \textbf{69.5}\pmsdt{3.2} & 60.6\pmsdt{1.2} & 51.5\pmsdt{2.1} & 53.3\pmsdt{7.9} & \textbf{43.2}\pmsdt{0.4} \\
D+M+L & 75.9\pmsdt{1.3} & 80.7\pmsdt{7.4} & 78.6\pmsdt{15.1} & \textbf{59.0}\pmsdt{0.5} & 46.4\pmsdt{0.1} & 49.6\pmsdt{2.1} & 51.5\pmsdt{9.6} & \textbf{40.2}\pmsdt{0.3} \\
D+M+W & 106.5\pmsdt{1.0} & 93.6\pmsdt{5.1} & 99.7\pmsdt{35.5} & \textbf{60.3}\pmsdt{0.6} & 62.4\pmsdt{0.7} & 55.9\pmsdt{3.5} & 56.6\pmsdt{15.4} & \textbf{41.1}\pmsdt{0.3} \\
D+L+W & 95.9\pmsdt{1.5} & 81.2\pmsdt{6.1} & 82.2\pmsdt{17.0} & \textbf{60.6}\pmsdt{0.3} & 57.8\pmsdt{0.6} & 50.0\pmsdt{2.7} & 51.9\pmsdt{9.7} & \textbf{41.4}\pmsdt{0.2} \\
M+L+W & 105.9\pmsdt{2.3} & 89.4\pmsdt{4.7} & 88.6\pmsdt{22.6} & \textbf{65.9}\pmsdt{0.3} & 62.8\pmsdt{0.4} & 53.0\pmsdt{1.2} & 57.2\pmsdt{18.5} & \textbf{43.0}\pmsdt{0.2} \\
\addlinespace[2pt]
R+D+M+L & 74.5\pmsdt{2.2} & 80.4\pmsdt{5.2} & 77.9\pmsdt{9.0} & \textbf{58.6}\pmsdt{1.5} & 45.8\pmsdt{0.8} & 49.0\pmsdt{1.9} & 51.9\pmsdt{7.8} & \textbf{39.1}\pmsdt{0.2} \\
R+D+M+W & 95.9\pmsdt{1.5} & 87.9\pmsdt{2.2} & 85.0\pmsdt{12.3} & \textbf{58.8}\pmsdt{1.3} & 57.2\pmsdt{1.3} & 52.1\pmsdt{2.1} & 52.0\pmsdt{6.5} & \textbf{39.5}\pmsdt{0.2} \\
R+D+L+W & 89.1\pmsdt{0.5} & 81.5\pmsdt{5.6} & 81.7\pmsdt{11.7} & \textbf{59.9}\pmsdt{1.1} & 54.2\pmsdt{1.2} & 49.7\pmsdt{2.5} & 52.3\pmsdt{8.1} & \textbf{40.2}\pmsdt{0.2} \\
R+M+L+W & 96.5\pmsdt{2.6} & 88.3\pmsdt{4.2} & 85.2\pmsdt{13.8} & \textbf{67.5}\pmsdt{3.6} & 57.4\pmsdt{0.9} & 51.5\pmsdt{1.6} & 52.7\pmsdt{7.3} & \textbf{41.5}\pmsdt{0.5} \\
D+M+L+W & 91.6\pmsdt{1.2} & 82.2\pmsdt{5.9} & 79.5\pmsdt{14.3} & \textbf{58.8}\pmsdt{0.5} & 55.4\pmsdt{0.5} & 50.3\pmsdt{2.5} & 50.9\pmsdt{8.2} & \textbf{40.2}\pmsdt{0.3} \\
\addlinespace[2pt]
R+D+M+L+W & 87.0\pmsdt{1.4} & 82.1\pmsdt{4.8} & 80.5\pmsdt{10.2} & \textbf{58.4}\pmsdt{1.4} & 52.9\pmsdt{0.9} & 49.3\pmsdt{1.9} & 51.9\pmsdt{7.3} & \textbf{39.1}\pmsdt{0.2} \\
\midrule
Avg.\,(31) & 98.4\pmsdt{1.2} & 97.3\pmsdt{2.9} & 100.5\pmsdt{26.6} & \textbf{72.7}\pmsdt{1.7} & 57.9\pmsdt{0.8} & 56.5\pmsdt{1.9} & 58.5\pmsdt{12.5} & \textbf{45.0}\pmsdt{0.2} \\
\bottomrule
\end{tabular}
\end{table*}

\begin{table*}[t]
\centering\scriptsize\renewcommand{\arraystretch}{1.0}
\setlength{\tabcolsep}{3pt}
\caption{\textbf{Humanoid-Everyday test, all 15 input subsets: \ours{} has the lowest nMSE and MAE with every subset.} Next-command error (MAE in rad), mean\,$\pm$\,SD over three seeds; the baselines are trained on complete inputs and read the subsets without retraining.}
\label{tab:sub_he}
\begin{tabular}{@{}l rrrr rrrr@{}}
\toprule
 & \multicolumn{4}{c}{nMSE\dn} & \multicolumn{4}{c@{}}{MAE\dn} \\
\cmidrule(lr){2-5}\cmidrule(l){6-9}
Input & Late fusion & MBT & Perceiver IO & \ours{} & Late fusion & MBT & Perceiver IO & \ours{} \\
\midrule
R & 1.603\pmsdt{.097} & 2.524\pmsdt{.287} & 1.598\pmsdt{.240} & \textbf{0.704}\pmsdt{.010} & 0.2451\pmsdt{.0044} & 0.3169\pmsdt{.0209} & 0.2451\pmsdt{.0169} & \textbf{0.1428}\pmsdt{.0031} \\
D & 1.839\pmsdt{.087} & 9.179\pmsdt{.849} & 1.443\pmsdt{.134} & \textbf{0.565}\pmsdt{.018} & 0.2407\pmsdt{.0026} & 0.5107\pmsdt{.0295} & 0.2369\pmsdt{.0169} & \textbf{0.1268}\pmsdt{.0034} \\
L & 1.129\pmsdt{.011} & 1.317\pmsdt{.070} & 1.754\pmsdt{.324} & \textbf{0.735}\pmsdt{.020} & 0.2137\pmsdt{.0025} & 0.2401\pmsdt{.0067} & 0.2590\pmsdt{.0206} & \textbf{0.1549}\pmsdt{.0028} \\
proprio & 0.527\pmsdt{.023} & 9.085\pmsdt{.381} & 0.581\pmsdt{.230} & \textbf{0.178}\pmsdt{.011} & 0.1266\pmsdt{.0018} & 0.4531\pmsdt{.0118} & 0.1109\pmsdt{.0100} & \textbf{0.0595}\pmsdt{.0013} \\
\addlinespace[2pt]
R + D & 1.400\pmsdt{.012} & 2.343\pmsdt{.079} & 1.535\pmsdt{.135} & \textbf{0.622}\pmsdt{.015} & 0.2160\pmsdt{.0016} & 0.2994\pmsdt{.0078} & 0.2405\pmsdt{.0100} & \textbf{0.1327}\pmsdt{.0031} \\
R + L & 1.216\pmsdt{.045} & 1.103\pmsdt{.022} & 1.530\pmsdt{.168} & \textbf{0.693}\pmsdt{.005} & 0.2153\pmsdt{.0031} & 0.2068\pmsdt{.0025} & 0.2406\pmsdt{.0134} & \textbf{0.1414}\pmsdt{.0020} \\
R + proprio & 0.290\pmsdt{.014} & 0.486\pmsdt{.039} & 0.250\pmsdt{.018} & \textbf{0.171}\pmsdt{.006} & 0.0901\pmsdt{.0014} & 0.1161\pmsdt{.0050} & 0.0730\pmsdt{.0044} & \textbf{0.0571}\pmsdt{.0009} \\
D + L & 1.347\pmsdt{.050} & 1.312\pmsdt{.063} & 1.398\pmsdt{.135} & \textbf{0.562}\pmsdt{.020} & 0.2139\pmsdt{.0017} & 0.2291\pmsdt{.0070} & 0.2311\pmsdt{.0134} & \textbf{0.1264}\pmsdt{.0036} \\
D + proprio & 0.247\pmsdt{.003} & 2.841\pmsdt{.254} & 0.398\pmsdt{.022} & \textbf{0.167}\pmsdt{.007} & 0.0743\pmsdt{.0003} & 0.2323\pmsdt{.0090} & 0.0957\pmsdt{.0038} & \textbf{0.0560}\pmsdt{.0006} \\
L + proprio & 0.288\pmsdt{.010} & 0.307\pmsdt{.013} & 0.383\pmsdt{.027} & \textbf{0.173}\pmsdt{.009} & 0.0897\pmsdt{.0010} & 0.0898\pmsdt{.0028} & 0.0944\pmsdt{.0028} & \textbf{0.0580}\pmsdt{.0009} \\
\addlinespace[2pt]
R + D + L & 1.213\pmsdt{.009} & 1.133\pmsdt{.013} & 1.505\pmsdt{.124} & \textbf{0.620}\pmsdt{.015} & 0.2054\pmsdt{.0012} & 0.2047\pmsdt{.0020} & 0.2383\pmsdt{.0092} & \textbf{0.1323}\pmsdt{.0029} \\
R + D + proprio & 0.200\pmsdt{.002} & 0.504\pmsdt{.006} & 0.212\pmsdt{.005} & \textbf{0.167}\pmsdt{.005} & 0.0649\pmsdt{.0004} & 0.1104\pmsdt{.0014} & 0.0641\pmsdt{.0008} & \textbf{0.0559}\pmsdt{.0007} \\
R + L + proprio & 0.212\pmsdt{.007} & 0.179\pmsdt{.006} & 0.243\pmsdt{.021} & \textbf{0.170}\pmsdt{.006} & 0.0735\pmsdt{.0010} & 0.0605\pmsdt{.0017} & 0.0713\pmsdt{.0052} & \textbf{0.0569}\pmsdt{.0010} \\
D + L + proprio & 0.201\pmsdt{.002} & 0.288\pmsdt{.014} & 0.385\pmsdt{.017} & \textbf{0.166}\pmsdt{.006} & 0.0649\pmsdt{.0001} & 0.0823\pmsdt{.0032} & 0.0943\pmsdt{.0031} & \textbf{0.0558}\pmsdt{.0006} \\
\addlinespace[2pt]
R + D + L + proprio & 0.180\pmsdt{.001} & 0.185\pmsdt{.004} & 0.210\pmsdt{.008} & \textbf{0.166}\pmsdt{.005} & 0.0605\pmsdt{.0002} & 0.0586\pmsdt{.0008} & 0.0639\pmsdt{.0015} & \textbf{0.0558}\pmsdt{.0007} \\
\midrule
Avg.\,(15) & 0.793\pmsdt{.003} & 2.186\pmsdt{.060} & 0.895\pmsdt{.062} & \textbf{0.390}\pmsdt{.002} & 0.1463\pmsdt{.0009} & 0.2141\pmsdt{.0026} & 0.1573\pmsdt{.0049} & \textbf{0.0942}\pmsdt{.0011} \\
\bottomrule
\end{tabular}
\end{table*}

\begin{table*}[t]
\centering\scriptsize\renewcommand{\arraystretch}{1.0}
\setlength{\tabcolsep}{3pt}
\caption{\textbf{MM-Fi unseen room (E04), all 31 input subsets: \ours{} has the lowest MPJPE on 28 and the lowest PA-MPJPE on 29.} Units, seeds, runs and abbreviations as in Table~\ref{tab:sub_mmfi_id}.}
\label{tab:sub_mmfi_ood}
\begin{tabular}{@{}l rrrr rrrr@{}}
\toprule
 & \multicolumn{4}{c}{MPJPE\dn} & \multicolumn{4}{c@{}}{PA-MPJPE\dn} \\
\cmidrule(lr){2-5}\cmidrule(l){6-9}
Input & Late fusion & X-Fi & COMPASS & \ours{} & Late fusion & X-Fi & COMPASS & \ours{} \\
\midrule
R & 257.5\pmsdt{26.6} & \textbf{182.4}\pmsdt{11.1} & 242.8\pmsdt{38.2} & 242.4\pmsdt{73.1} & 125.8\pmsdt{6.5} & 104.4\pmsdt{4.5} & 99.0\pmsdt{5.8} & \textbf{87.5}\pmsdt{6.2} \\
D & 117.8\pmsdt{18.2} & 129.1\pmsdt{9.8} & 193.8\pmsdt{97.5} & \textbf{77.4}\pmsdt{1.4} & 63.8\pmsdt{4.9} & 68.2\pmsdt{1.2} & 87.0\pmsdt{37.8} & \textbf{50.9}\pmsdt{0.1} \\
M & 183.7\pmsdt{2.1} & 214.2\pmsdt{15.7} & 218.5\pmsdt{56.0} & \textbf{168.6}\pmsdt{0.9} & 83.2\pmsdt{1.7} & 89.6\pmsdt{3.5} & 87.0\pmsdt{13.5} & \textbf{75.8}\pmsdt{0.8} \\
L & 118.3\pmsdt{17.0} & 131.2\pmsdt{9.2} & 164.0\pmsdt{67.2} & \textbf{116.2}\pmsdt{23.8} & \textbf{59.4}\pmsdt{3.8} & 64.1\pmsdt{2.5} & 65.1\pmsdt{14.0} & 68.2\pmsdt{22.4} \\
W & 360.6\pmsdt{20.8} & \textbf{337.0}\pmsdt{4.1} & 366.3\pmsdt{49.9} & 367.9\pmsdt{17.6} & 112.1\pmsdt{1.6} & 113.5\pmsdt{4.6} & 114.3\pmsdt{2.2} & \textbf{106.9}\pmsdt{1.9} \\
\addlinespace[2pt]
R+D & 171.6\pmsdt{4.1} & 128.3\pmsdt{7.3} & 183.1\pmsdt{28.2} & \textbf{82.8}\pmsdt{1.2} & 88.4\pmsdt{3.9} & 79.4\pmsdt{2.3} & 82.1\pmsdt{7.7} & \textbf{56.2}\pmsdt{0.6} \\
R+M & 197.0\pmsdt{15.0} & 163.7\pmsdt{12.3} & 210.1\pmsdt{3.6} & \textbf{136.7}\pmsdt{1.9} & 94.6\pmsdt{2.1} & 96.0\pmsdt{4.1} & 95.9\pmsdt{5.7} & \textbf{79.4}\pmsdt{3.2} \\
R+L & 173.2\pmsdt{15.2} & 140.1\pmsdt{18.1} & 173.5\pmsdt{9.6} & \textbf{116.0}\pmsdt{9.3} & 82.5\pmsdt{2.3} & 74.2\pmsdt{2.2} & 83.1\pmsdt{3.1} & \textbf{70.5}\pmsdt{2.6} \\
R+W & 296.6\pmsdt{22.4} & \textbf{192.7}\pmsdt{11.6} & 257.0\pmsdt{57.2} & 239.2\pmsdt{65.2} & 108.9\pmsdt{2.3} & 101.6\pmsdt{2.7} & 98.8\pmsdt{4.9} & \textbf{88.3}\pmsdt{6.2} \\
D+M & 133.4\pmsdt{6.9} & 127.3\pmsdt{15.2} & 152.3\pmsdt{36.5} & \textbf{76.3}\pmsdt{1.6} & 66.3\pmsdt{2.7} & 67.5\pmsdt{3.0} & 74.0\pmsdt{17.7} & \textbf{49.5}\pmsdt{0.2} \\
D+L & 107.1\pmsdt{11.6} & 116.6\pmsdt{15.4} & 139.3\pmsdt{31.8} & \textbf{75.8}\pmsdt{1.6} & 56.4\pmsdt{3.2} & 59.7\pmsdt{2.0} & 63.5\pmsdt{8.5} & \textbf{49.8}\pmsdt{0.5} \\
D+W & 222.2\pmsdt{6.6} & 162.1\pmsdt{20.2} & 174.5\pmsdt{53.4} & \textbf{79.3}\pmsdt{1.6} & 80.1\pmsdt{2.5} & 70.9\pmsdt{1.3} & 71.4\pmsdt{10.8} & \textbf{51.1}\pmsdt{0.1} \\
M+L & 134.9\pmsdt{7.2} & 135.0\pmsdt{7.9} & 144.1\pmsdt{38.9} & \textbf{109.4}\pmsdt{16.6} & 63.8\pmsdt{2.3} & 63.5\pmsdt{2.4} & 65.3\pmsdt{16.3} & \textbf{61.5}\pmsdt{13.8} \\
M+W & 245.3\pmsdt{12.8} & 241.6\pmsdt{24.8} & 239.5\pmsdt{44.5} & \textbf{196.2}\pmsdt{5.8} & 88.3\pmsdt{1.5} & 94.4\pmsdt{6.3} & 88.8\pmsdt{12.4} & \textbf{79.2}\pmsdt{1.8} \\
L+W & 222.0\pmsdt{18.2} & 154.7\pmsdt{14.8} & 157.4\pmsdt{45.9} & \textbf{115.0}\pmsdt{21.0} & 77.2\pmsdt{2.9} & 66.1\pmsdt{0.8} & \textbf{64.4}\pmsdt{12.5} & 68.5\pmsdt{21.6} \\
\addlinespace[2pt]
R+D+M & 159.8\pmsdt{4.1} & 126.7\pmsdt{4.7} & 169.4\pmsdt{16.3} & \textbf{81.0}\pmsdt{0.9} & 79.9\pmsdt{2.3} & 77.2\pmsdt{2.7} & 80.3\pmsdt{5.7} & \textbf{54.3}\pmsdt{0.5} \\
R+D+L & 145.1\pmsdt{8.1} & 127.0\pmsdt{14.9} & 151.0\pmsdt{11.7} & \textbf{81.1}\pmsdt{1.0} & 72.6\pmsdt{2.4} & 68.8\pmsdt{2.1} & 76.7\pmsdt{8.1} & \textbf{54.0}\pmsdt{0.6} \\
R+D+W & 225.2\pmsdt{10.3} & 146.6\pmsdt{1.6} & 188.4\pmsdt{29.5} & \textbf{83.5}\pmsdt{1.1} & 89.1\pmsdt{2.1} & 80.3\pmsdt{2.1} & 82.1\pmsdt{7.6} & \textbf{56.3}\pmsdt{0.4} \\
R+M+L & 161.1\pmsdt{11.2} & 140.8\pmsdt{19.6} & 161.7\pmsdt{4.6} & \textbf{111.2}\pmsdt{7.7} & 76.5\pmsdt{1.5} & 74.1\pmsdt{3.8} & 81.8\pmsdt{2.7} & \textbf{66.5}\pmsdt{2.3} \\
R+M+W & 239.9\pmsdt{17.0} & 177.1\pmsdt{20.5} & 224.2\pmsdt{21.2} & \textbf{139.0}\pmsdt{2.8} & 93.7\pmsdt{0.9} & 93.8\pmsdt{4.5} & 95.3\pmsdt{5.0} & \textbf{79.9}\pmsdt{3.5} \\
R+L+W & 225.7\pmsdt{17.7} & 152.7\pmsdt{15.8} & 184.6\pmsdt{20.6} & \textbf{116.7}\pmsdt{9.1} & 85.9\pmsdt{1.0} & 76.0\pmsdt{2.2} & 83.4\pmsdt{3.6} & \textbf{70.9}\pmsdt{3.1} \\
D+M+L & 119.3\pmsdt{6.8} & 119.1\pmsdt{11.8} & 125.1\pmsdt{14.5} & \textbf{74.8}\pmsdt{2.0} & 59.6\pmsdt{2.6} & 60.4\pmsdt{2.5} & 64.1\pmsdt{10.1} & \textbf{48.5}\pmsdt{0.2} \\
D+M+W & 191.8\pmsdt{4.4} & 145.0\pmsdt{7.7} & 162.3\pmsdt{43.8} & \textbf{77.9}\pmsdt{1.8} & 75.0\pmsdt{2.0} & 69.8\pmsdt{1.1} & 71.1\pmsdt{12.5} & \textbf{49.7}\pmsdt{0.2} \\
D+L+W & 177.1\pmsdt{11.8} & 129.9\pmsdt{15.9} & 136.8\pmsdt{20.2} & \textbf{76.4}\pmsdt{1.8} & 68.3\pmsdt{2.8} & 61.6\pmsdt{1.6} & 63.0\pmsdt{7.5} & \textbf{49.8}\pmsdt{0.4} \\
M+L+W & 192.2\pmsdt{12.3} & 156.7\pmsdt{20.3} & 153.2\pmsdt{46.4} & \textbf{108.7}\pmsdt{14.7} & 73.1\pmsdt{2.2} & 66.2\pmsdt{1.9} & 65.5\pmsdt{16.1} & \textbf{62.0}\pmsdt{13.5} \\
\addlinespace[2pt]
R+D+M+L & 142.8\pmsdt{6.2} & 127.4\pmsdt{14.5} & 141.9\pmsdt{5.4} & \textbf{79.7}\pmsdt{0.8} & 70.4\pmsdt{1.9} & 68.4\pmsdt{2.7} & 76.5\pmsdt{8.0} & \textbf{52.4}\pmsdt{0.8} \\
R+D+M+W & 200.9\pmsdt{8.9} & 141.4\pmsdt{10.8} & 175.8\pmsdt{19.4} & \textbf{81.8}\pmsdt{0.8} & 82.7\pmsdt{1.4} & 77.7\pmsdt{3.1} & 80.9\pmsdt{6.4} & \textbf{54.4}\pmsdt{0.5} \\
R+D+L+W & 190.5\pmsdt{11.4} & 138.7\pmsdt{14.5} & 157.6\pmsdt{15.2} & \textbf{81.5}\pmsdt{0.9} & 77.2\pmsdt{1.6} & 70.3\pmsdt{1.8} & 77.1\pmsdt{8.5} & \textbf{54.0}\pmsdt{0.7} \\
R+M+L+W & 201.5\pmsdt{14.3} & 158.0\pmsdt{22.7} & 172.9\pmsdt{12.0} & \textbf{111.9}\pmsdt{7.6} & 80.4\pmsdt{0.9} & 75.6\pmsdt{3.3} & 82.5\pmsdt{3.3} & \textbf{66.9}\pmsdt{2.8} \\
D+M+L+W & 166.0\pmsdt{8.2} & 138.6\pmsdt{14.1} & 135.2\pmsdt{24.7} & \textbf{75.4}\pmsdt{2.2} & 67.4\pmsdt{2.4} & 62.6\pmsdt{1.5} & 63.4\pmsdt{8.5} & \textbf{48.6}\pmsdt{0.2} \\
\addlinespace[2pt]
R+D+M+L+W & 178.4\pmsdt{9.5} & 133.7\pmsdt{18.4} & 149.2\pmsdt{9.4} & \textbf{80.2}\pmsdt{0.7} & 74.5\pmsdt{1.5} & 69.3\pmsdt{2.5} & 77.3\pmsdt{8.8} & \textbf{52.5}\pmsdt{0.8} \\
\midrule
Avg.\,(31) & 189.0\pmsdt{8.1} & 155.3\pmsdt{8.8} & 180.8\pmsdt{30.8} & \textbf{118.1}\pmsdt{3.7} & 79.9\pmsdt{1.7} & 76.3\pmsdt{1.8} & 79.4\pmsdt{8.6} & \textbf{63.4}\pmsdt{1.7} \\
\bottomrule
\end{tabular}
\end{table*}

\begin{table*}[t]
\centering\scriptsize\renewcommand{\arraystretch}{1.0}
\setlength{\tabcolsep}{4pt}
\caption{\textbf{V2X-Real test, all 8 node subsets: \ours{} has the highest AP50 and AP70 with every set of nodes.} AP in \%, mean\,$\pm$\,SD over three seeds; E: ego; V: the other vehicle; I1, I2: the two infrastructure nodes; 70\,m: \ours{} with the nodes within the baselines' range.}
\label{tab:sub_v2x}
\begin{tabular}{@{}l rrrrrr@{}}
\toprule
Input nodes & Late fusion & AttFuse & HEAL & Pyramid Fusion & \ours{} & \ours{}, 70\,m \\
\midrule
\multicolumn{7}{@{}l}{\textit{AP50\up}} \\
E & 25.2\pmsdt{0.9} & 26.1\pmsdt{0.6} & 30.5\pmsdt{0.9} & 26.6\pmsdt{0.7} & \textbf{39.2}\pmsdt{1.1} & \textbf{39.2}\pmsdt{1.2} \\
\addlinespace[1pt]
E + V & 27.0\pmsdt{1.3} & 29.6\pmsdt{1.3} & 35.4\pmsdt{0.7} & 34.7\pmsdt{0.4} & \textbf{43.1}\pmsdt{1.1} & 42.9\pmsdt{1.1} \\
E + I1 & 26.0\pmsdt{0.9} & 27.1\pmsdt{0.8} & 33.5\pmsdt{0.6} & 33.6\pmsdt{1.8} & \textbf{41.9}\pmsdt{1.4} & 41.8\pmsdt{1.4} \\
E + I2 & 28.6\pmsdt{1.4} & 27.0\pmsdt{1.1} & 34.9\pmsdt{1.0} & 33.8\pmsdt{1.2} & \textbf{41.4}\pmsdt{1.3} & 41.2\pmsdt{1.1} \\
\addlinespace[1pt]
E + V + I1 & 27.1\pmsdt{1.3} & 30.4\pmsdt{1.5} & 37.8\pmsdt{0.9} & 38.6\pmsdt{0.7} & \textbf{44.1}\pmsdt{1.2} & 43.9\pmsdt{1.3} \\
E + V + I2 & 29.2\pmsdt{1.5} & 30.1\pmsdt{1.6} & 38.7\pmsdt{0.5} & 38.7\pmsdt{0.6} & \textbf{43.8}\pmsdt{1.2} & 43.5\pmsdt{1.0} \\
E + I1 + I2 & 28.8\pmsdt{1.3} & 28.1\pmsdt{1.2} & 36.0\pmsdt{0.6} & 37.0\pmsdt{1.7} & \textbf{42.8}\pmsdt{1.4} & 42.6\pmsdt{1.4} \\
\addlinespace[1pt]
E + V + I1 + I2 & 29.3\pmsdt{1.2} & 30.8\pmsdt{1.7} & 39.4\pmsdt{0.2} & 40.7\pmsdt{1.1} & \textbf{44.4}\pmsdt{1.2} & 44.2\pmsdt{1.2} \\
\addlinespace[2pt]
Avg.\,(8) & 27.7\pmsdt{1.2} & 28.6\pmsdt{1.2} & 35.8\pmsdt{0.2} & 35.4\pmsdt{1.0} & \textbf{42.6}\pmsdt{1.2} & 42.4\pmsdt{1.2} \\
\midrule
\multicolumn{7}{@{}l}{\textit{AP70\up}} \\
E & 11.8\pmsdt{1.5} & 12.0\pmsdt{1.7} & 17.5\pmsdt{0.1} & 15.9\pmsdt{0.2} & \textbf{26.5}\pmsdt{0.9} & \textbf{26.5}\pmsdt{0.9} \\
\addlinespace[1pt]
E + V & 12.4\pmsdt{1.1} & 13.8\pmsdt{2.5} & 21.1\pmsdt{1.1} & 20.9\pmsdt{1.1} & \textbf{28.7}\pmsdt{0.7} & 28.6\pmsdt{0.7} \\
E + I1 & 12.1\pmsdt{1.4} & 12.5\pmsdt{1.8} & 19.1\pmsdt{1.7} & 19.6\pmsdt{2.2} & \textbf{27.4}\pmsdt{1.0} & \textbf{27.4}\pmsdt{1.0} \\
E + I2 & 12.5\pmsdt{1.0} & 12.4\pmsdt{1.9} & 19.7\pmsdt{0.8} & 19.5\pmsdt{1.1} & \textbf{27.4}\pmsdt{0.7} & 27.3\pmsdt{0.7} \\
\addlinespace[1pt]
E + V + I1 & 12.4\pmsdt{1.1} & 14.1\pmsdt{2.7} & 23.0\pmsdt{1.6} & 23.5\pmsdt{1.6} & \textbf{29.0}\pmsdt{0.7} & 28.9\pmsdt{0.7} \\
E + V + I2 & 12.7\pmsdt{0.7} & 14.0\pmsdt{2.6} & 22.6\pmsdt{2.1} & 23.3\pmsdt{1.0} & \textbf{29.0}\pmsdt{0.3} & 28.9\pmsdt{0.4} \\
E + I1 + I2 & 12.6\pmsdt{0.8} & 12.9\pmsdt{2.0} & 20.3\pmsdt{0.7} & 21.8\pmsdt{1.3} & \textbf{28.2}\pmsdt{0.8} & 28.0\pmsdt{0.8} \\
\addlinespace[1pt]
E + V + I1 + I2 & 12.8\pmsdt{0.7} & 14.4\pmsdt{2.8} & 23.0\pmsdt{1.7} & 24.8\pmsdt{1.3} & \textbf{29.3}\pmsdt{0.5} & 29.1\pmsdt{0.6} \\
\addlinespace[2pt]
Avg.\,(8) & 12.4\pmsdt{1.0} & 13.3\pmsdt{2.3} & 20.8\pmsdt{1.1} & 21.2\pmsdt{1.2} & \textbf{28.2}\pmsdt{0.7} & 28.1\pmsdt{0.7} \\
\bottomrule
\end{tabular}
\end{table*}

\subsection{Qualitative results on MM-Fi}\label{supp:qual_mmfi}
\begin{figure*}[t]
\centering
\includegraphics[width=0.93\textwidth]{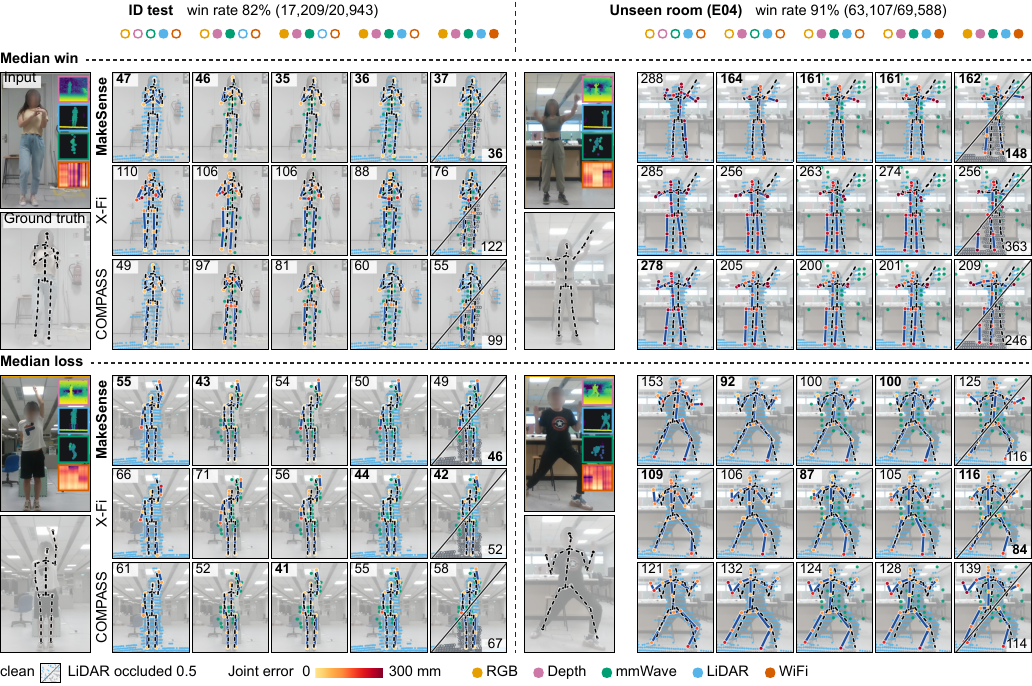}
\caption{\textbf{Median MM-Fi cases: \ours{} wins 82\,\% of ID and 91\,\% of unseen-room frames, with smaller loss than win margins.} Eligible frames, one seed (MPJPE in mm; columns as in main-paper Fig.~5).}
\label{fig:supp_qual_mmfi}
\end{figure*}

\noindent Each block shows a median case, not the best one: a frame is a win when \ours{} has the lowest five-sensor MPJPE of the three methods both clean and with LiDAR occluded, a loss when X-Fi or COMPASS is lower on clean input, and the drawn win or loss has the median margin (Fig.~\ref{fig:supp_qual_mmfi}). Fig.~\ref{fig:supp_sensor_chain} adds the sensors one at a time on the frame closest to the median for both \ours{} and X-Fi. Eligible frames have valid ground truth, the subject upright inside the photo and non-empty LiDAR and mmWave files; recordings drawn in other figures are left out.

\subsection{Qualitative results: other datasets}\label{supp:qual_other}
\begin{figure*}[t]
\centering
\includegraphics[width=0.93\textwidth]{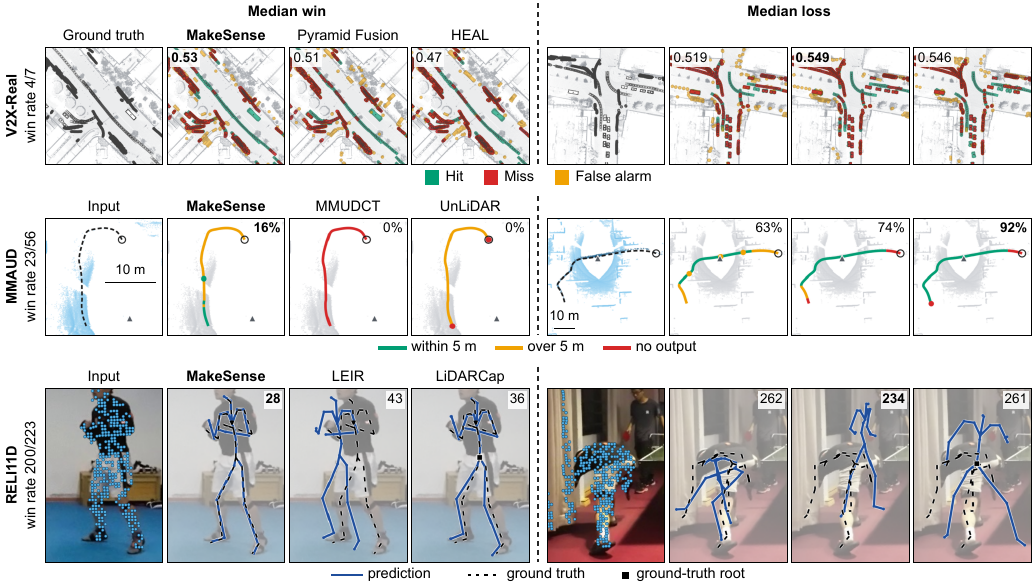}
\caption{\textbf{Median wins and losses: \ours{} wins 200 of 223 RELI11D windows, 4 of 7 V2X-Real clips and 23 of 56 MMAUD sequences.} One training seed; printed: clip F1 (V2X-Real), share of steps within 5\,m (MMAUD) and last-frame PA-MPJPE in mm (RELI11D).}
\label{fig:supp_qual_other}
\end{figure*}

\noindent Each row shows a median win and a median loss of \ours{} (Fig.~\ref{fig:supp_qual_other}). A win is the highest clip F1 of the six methods (V2X-Real), the highest share of steps within 5\,m of the three methods (MMAUD), or the lowest root-aligned MPJPE and PA-MPJPE over the window and on its last frame (RELI11D); cases drawn in the main paper are~left~out.

\end{document}